\documentclass[smallcondensed]{svjour3}

\pdfoutput=1

\usepackage{amsmath}
\usepackage{graphicx}
\usepackage{multirow}
\usepackage{amssymb}
\usepackage{subcaption}
\usepackage{siunitx}
\usepackage{url}
\usepackage[numbers]{natbib}
\usepackage{epstopdf}

\renewcommand{\vec}[1]{\mathbf{#1}}

\DeclareMathOperator*{\argmin}{arg\,min}
\DeclareMathOperator*{\argmax}{arg\,max}

\spnewtheorem{casestudy}{Mapping Scheme}{\bf}{\normalfont}

\def\makeheadbox{{%
\hbox to0pt{\vbox{\baselineskip=10dd\hrule\hbox
to\hsize{\vrule\kern3pt\vbox{\kern3pt
\hbox{Preprint submitted to Applied Intelligence.}
\kern3pt}\hfil\kern3pt\vrule}\hrule}%
\hss}}}

\begin{document}
	
        \title{Image-Based Benchmarking and Visualization for Large-Scale Global Optimization}

        \titlerunning{Image-Based Visualization for Large-Scale Global Optimization}        
        
        \author{Kyle Robert Harrison \and
                Azam Asilian Bidgoli \and
                Shahryar Rahnamayan \and
                Kalyanmoy Deb
        }
        
        \institute{K.R. Harrison \and A.A. Bidgoli \and S. Rahnamayan \at
                   Nature Inspired Computational Intelligence (NICI) Lab\\
                   Department of Electrical, Computer, and Software Engineering\\
                   University of Ontario Institute of Technology, Oshawa, Canada\\
                   \email{Shahryar.Rahnamayan@uoit.ca} \\
                   \emph{Present address of K.R. Harrison:} School of Engineering and Information Technology \\
                   University of New South Wales Canberra, Canberra, Australia 
                   \and
                   Kalyanmoy Deb \at
                   Computational Optimization and Innovation (COIN) Laboratory\\
                   Departments of Electrical and Computer Engineering, Computer Science and Engineering, and Mechanical Engineering\\
                   Michigan State University, East Lansing, USA
        }
        
        \date{\today}
        
        \maketitle
	
		\begin{abstract}
			In the context of optimization, visualization techniques can be useful for understanding the behaviour of optimization algorithms and can even provide a means to facilitate human interaction with an optimizer. Towards this goal, an image-based visualization framework, without dimension reduction, that visualizes the solutions to large-scale global optimization problems as images is proposed. In the proposed framework, the pixels visualize decision variables while the entire image represents the overall solution quality. This framework affords a number of benefits over existing visualization techniques including enhanced scalability (in terms of the number of decision variables), facilitation of standard image processing techniques, providing nearly infinite benchmark cases, and explicit alignment with human perception. Furthermore, image-based visualization can be used to visualize the optimization process in real-time, thereby allowing the user to ascertain characteristics of the search process as it is progressing. To the best of the authors' knowledge, this is the first realization of a dimension-preserving, scalable visualization framework that embeds the inherent relationship between decision space and objective space. The proposed framework is utilized with 10 different mapping schemes on an image-reconstruction problem that encompass continuous, discrete, binary, combinatorial, constrained, dynamic, and multi-objective optimization. The proposed framework is then demonstrated on arbitrary benchmark problems with known optima. Experimental results elucidate the flexibility and demonstrate how valuable information about the search process can be gathered via the proposed visualization framework.
		\end{abstract}


\section{Introduction}
	\label{sec:intro}
	
	Many real-world optimization problems involve a large number of problem dimensions. However, visualizing solutions with more than three dimensions is problematic and thus extracting useful information is complex. Specifically, high-dimensional data visualization often suffers from the ``curse of dimensionality" such that many existing visualization techniques do not scale well with respect to problem dimension. Nonetheless, visualization of high-dimensional data comes with a number of benefits. Firstly, visualization allows for the presentation of data in a much more concise and comprehensible manner. This can allow key patterns to be highlighted, which may otherwise be left unnoticed. Similarly, visualization provides a means for an interactive pipeline, whereby the decision maker can be made aware of underlying characteristics in the data, or the optimization process, and react accordingly. Furthermore, with regards to optimization, visualization of optimization results can give a clearer understanding of the effects of operational decisions and their results.
	
	Visualization techniques generally fall within one of two broad categories, namely those with dimensionality reduction and those without dimensionality reduction. Generally, the techniques that use dimension reduction provide a 2D or 3D projection of the data with the inherent goal of preserving spatial distance. While such techniques are scalable in terms of the number of problem dimensions, they suffer from the explicit drawback that they do not preserve dimensionality and, thus, the inferences that can be made from them are limited. Furthermore, the insights that can be made from the resulting projection are inherently tied to the quality of the projection; the projection may not convey complex patterns that are present in the data set. In contrast, visualization techniques that do not use dimension reduction are able to visualize problems with arbitrary dimension, in theory, without the use of projection. However, in practice, many of these techniques are limited in their scalability as the visualizations become infeasible when faced with problems that have a very large number of dimensions.
	
	\subsection{Dimensionality Reduction Techniques}

\begin{figure*}[ht!]
		\centering
		\begin{subfigure}{0.48\linewidth}
			\centering
			\frame{\includegraphics[width=\linewidth]{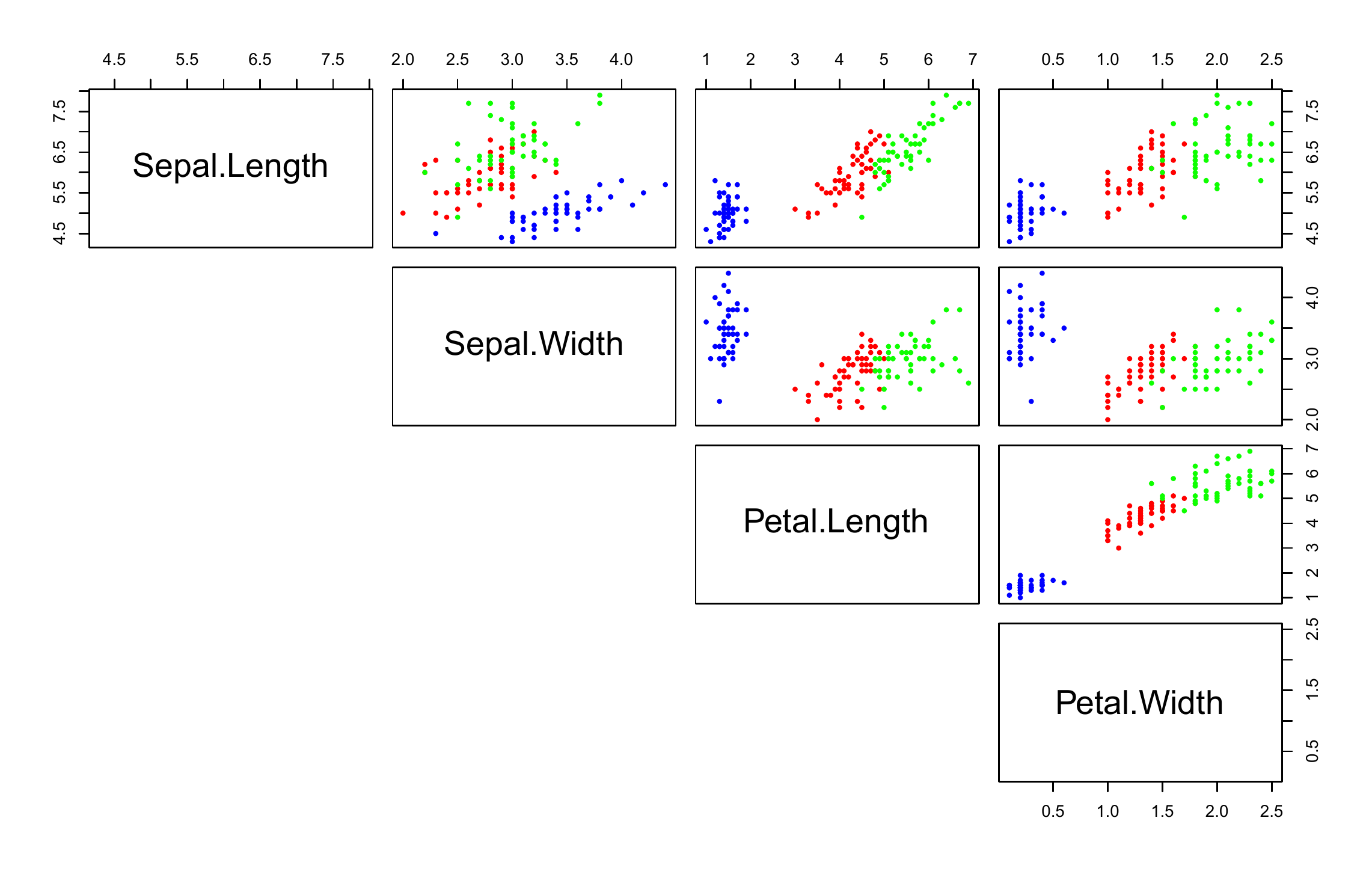}}
			\caption{Scatterplot Matrix}
			\label{fig:scatterplotMatrix}
		\end{subfigure}
		\begin{subfigure}{0.48\linewidth}
			\centering
			\frame{\includegraphics[width=\linewidth]{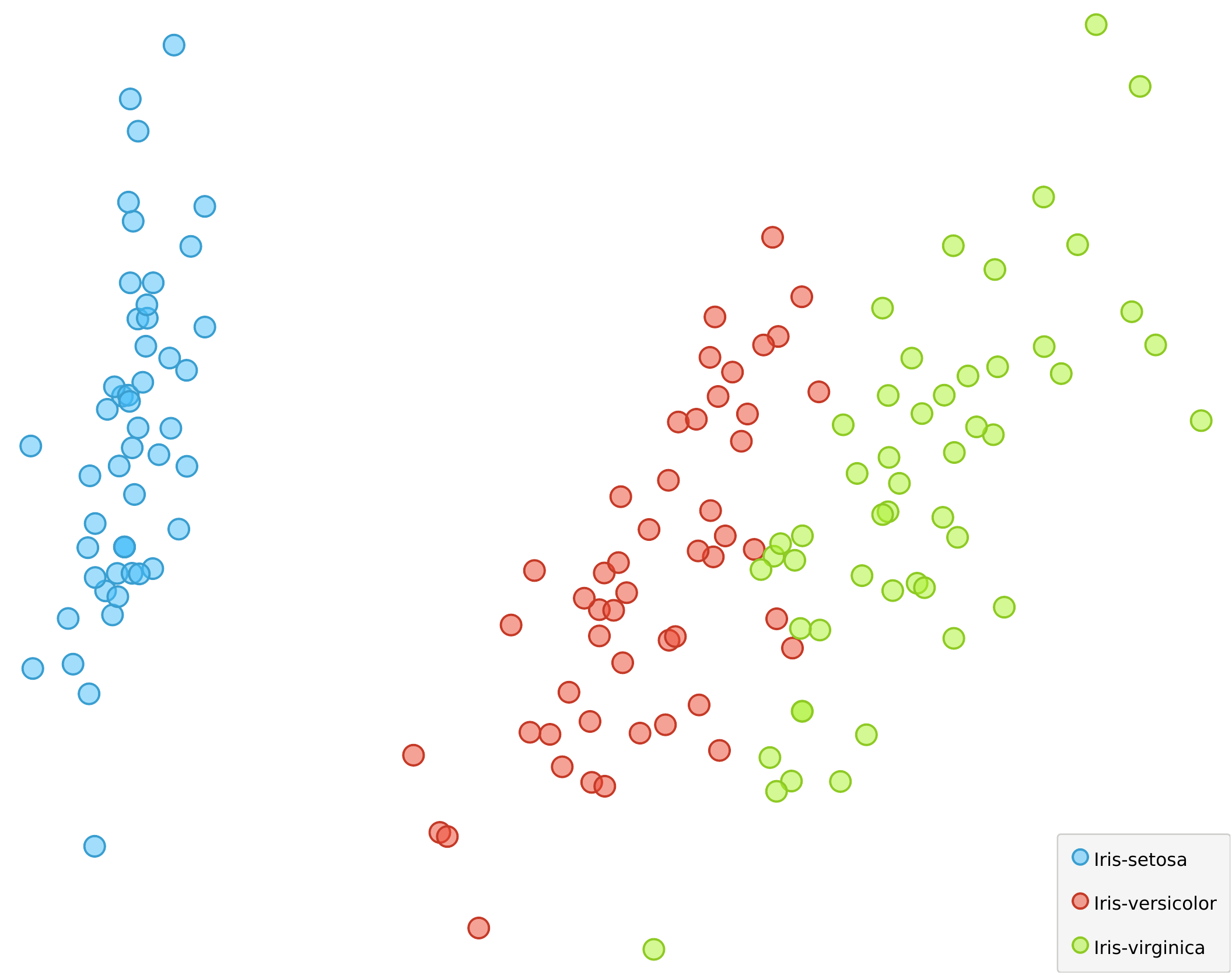}}
			\caption{Multidimensional scaling}
			\label{fig:mds}
		\end{subfigure}
		\begin{subfigure}{0.48\linewidth}
			\centering
			\frame{\includegraphics[width=\linewidth]{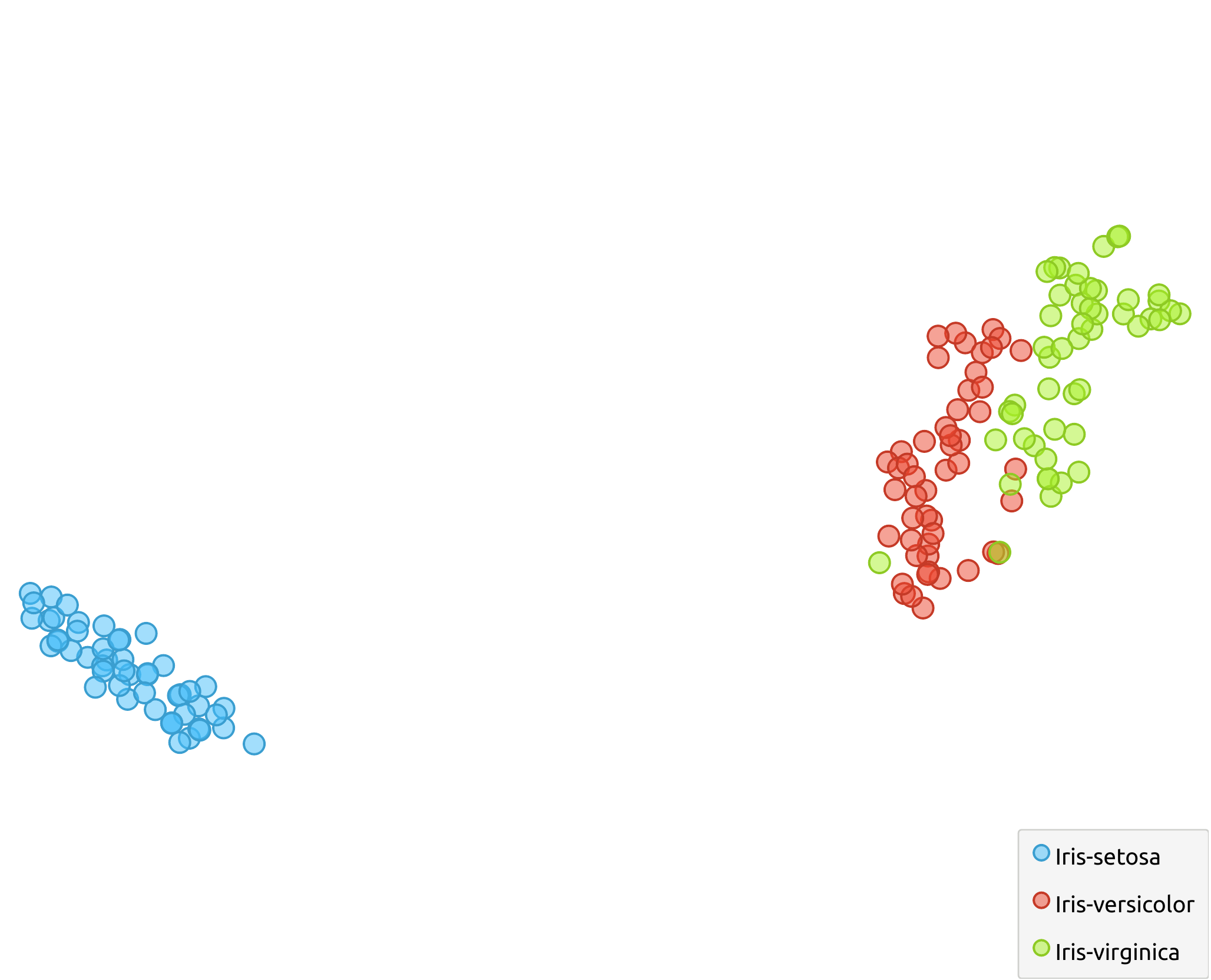}}
			\caption{t-SNE}
			\label{fig:tsne}
		\end{subfigure}
		\begin{subfigure}{0.48\linewidth}
			\centering
			\frame{\includegraphics[width=\linewidth]{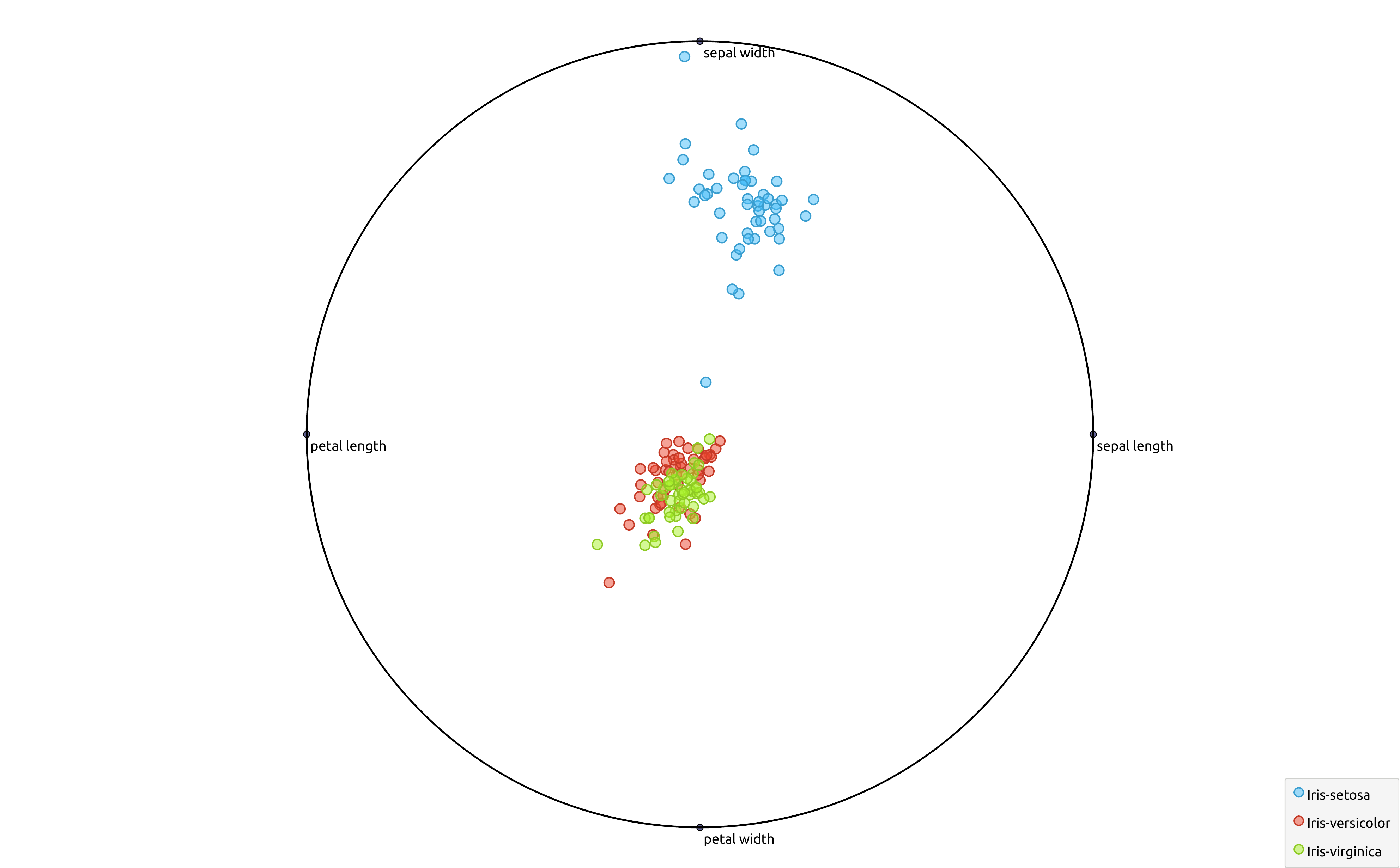}}
			\caption{RadVis}
			\label{fig:radvis}
		\end{subfigure}
		\caption{Sample dimension-reduction visualizations using the Iris dataset. The dataset contains 150 instances, 4 attributes, and 3 classes.}
		\label{fig:visualizations}
	\end{figure*}

	One of the simplest visualization techniques for high-dimensional data is a matrix of scatterplots, whereby each scatterplot independently visualizes each of the ${n \choose 2}$, or equivalently $\frac{n^2 - n}{2}$, distinct pairs of problem dimensions. However, this technique does not scale well in terms of problem dimensions as the space required to visualize all pairs exhibits quadratic growth. While it can be argued that a scatterplot matrix preserves dimensionality, as it provides a visual representation of all $n$ dimensions, it does so in a pairwise manner whereby only two dimensions are visualized simultaneously. Therefore, the scatterplot matrix is, in the context of this work, considered as a dimensionality-reducing technique. An example scatterplot matrix is shown in Figure \ref{fig:scatterplotMatrix}.
	
	A self-organizing map (SOM) \cite{Kohonen1990Self} is a type of unsupervised artificial neural network that provides a topology-preserving mapping from $n$-dimensional vectors to $m$-dimensional vectors, where $m < n$ and commonly $m = 2$. Typically, the output of a SOM can then be used to analyze clusters formed from high-dimensional data \cite{obayashi2003visualization}.
	
	Multidimensional scaling (MDS) \cite{torgerson1952multidimensional} refers to a family of data analysis methods for mapping $n$-dimensional data into $m$-dimensional points through non-linear, distance preserving ordination. In the case where $m < 3$, MDS can then be used to visualize the spatial relationship among the $n$-dimensional points. However, one limitation of MDS is that it does not capture global structure effectively \cite{maaten2008visualizing}. An example of MDS is shown in Figure \ref{fig:mds}.
	
	t-distributed stochastic neighbour embedding (t-SNE) \cite{maaten2008visualizing} provides a similar mapping capability as MDS, but with an enhanced ability to visualize global structure. t-SNE first defines a conditional probability distribution associated with each pair of $n$-dimensional points being selected as neighbours, according to their distance. A similar probability distribution is then formulated for the $m$-dimensional projection by minimizing the relative entropy between the two distributions. An example of t-SNE is shown in Figure \ref{fig:tsne}.

	Radial coordinates visualization (RadVis) \cite{hoffman1997dna} provides a nonlinear mapping of $n$-dimensional data to two dimensions using the physics of springs. The mapping for each data point is produced by attaching a ``spring" to each of the $n$ equidistant axes arranged on the perimeter of a circle such that, for each dimension, the corresponding spring constant is proportional to the data value for that dimension. The springs are then allowed to move freely (in two dimensions) until they have reached an equilibrium point. The resulting location provides the two-dimensional location of the original data point. This process is repeated for each data point. Similar to parallel coordinates, RadVis is implicitly limited in terms of scalability as it requires a circle to be divided into $n$ segments. An example of RadVis is shown in Figure \ref{fig:radvis}. Recently, RadVis was extended to facilitate mapping to a 3-dimensional space, whereby the third dimension can be used to visualize a quality metric \cite{ibrahim20163d}.
	
	While the aforementioned visualization techniques are commonly used, they all exhibit dimensionality reduction. In contrast, there are a number of techniques that are capable of visualizing $n$-dimensional data directly.
	
	\subsection{Dimensionality Preserving Techniques}
	
\begin{figure*}[ht!]
		\centering
		\begin{subfigure}{0.48\linewidth}
			\centering
			\frame{\includegraphics[width=\linewidth]{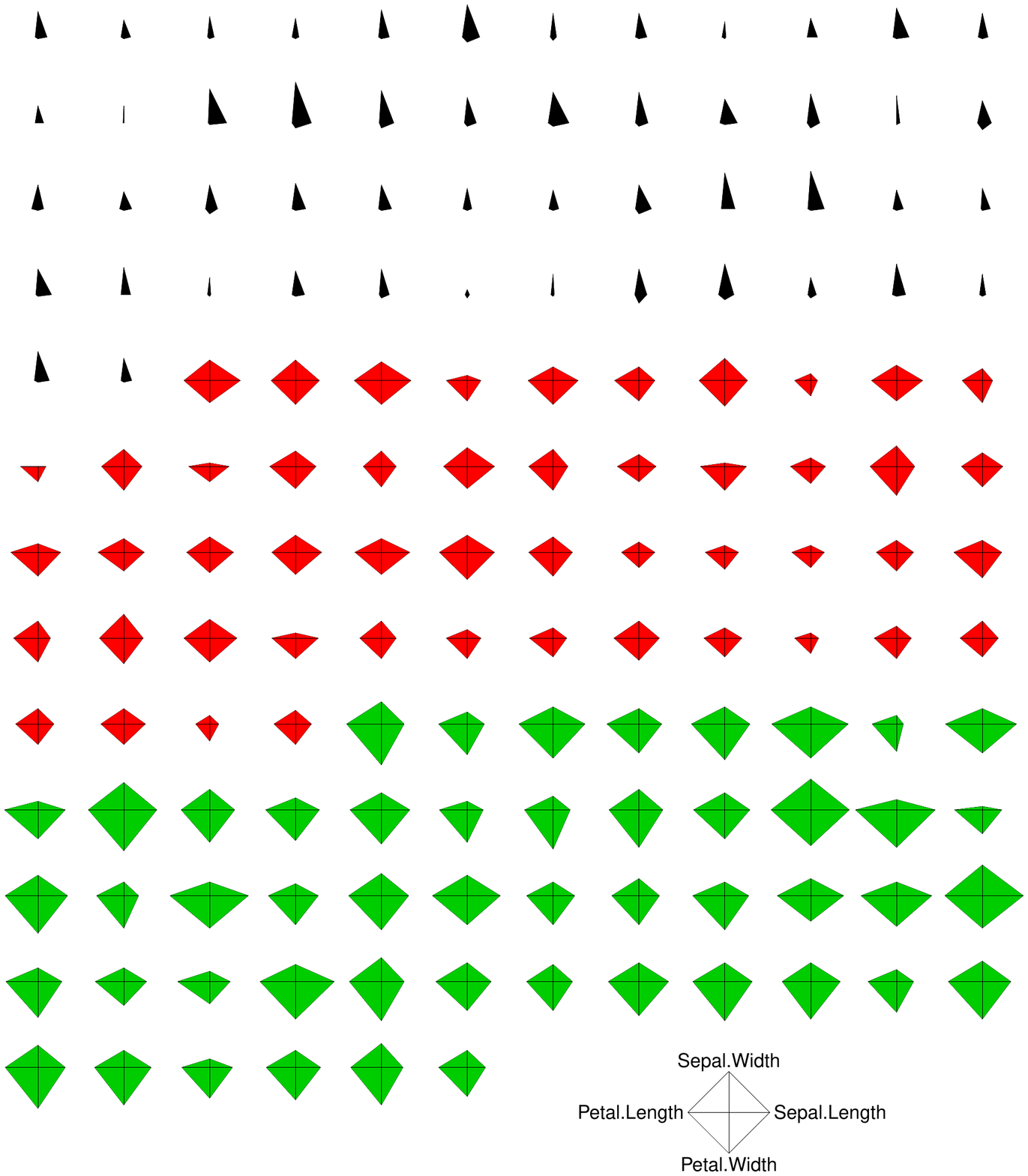}}
			\caption{Star plot}
			\label{fig:starplot}
		\end{subfigure}
		\begin{subfigure}{0.48\linewidth}
			\centering
			\frame{\includegraphics[width=\linewidth]{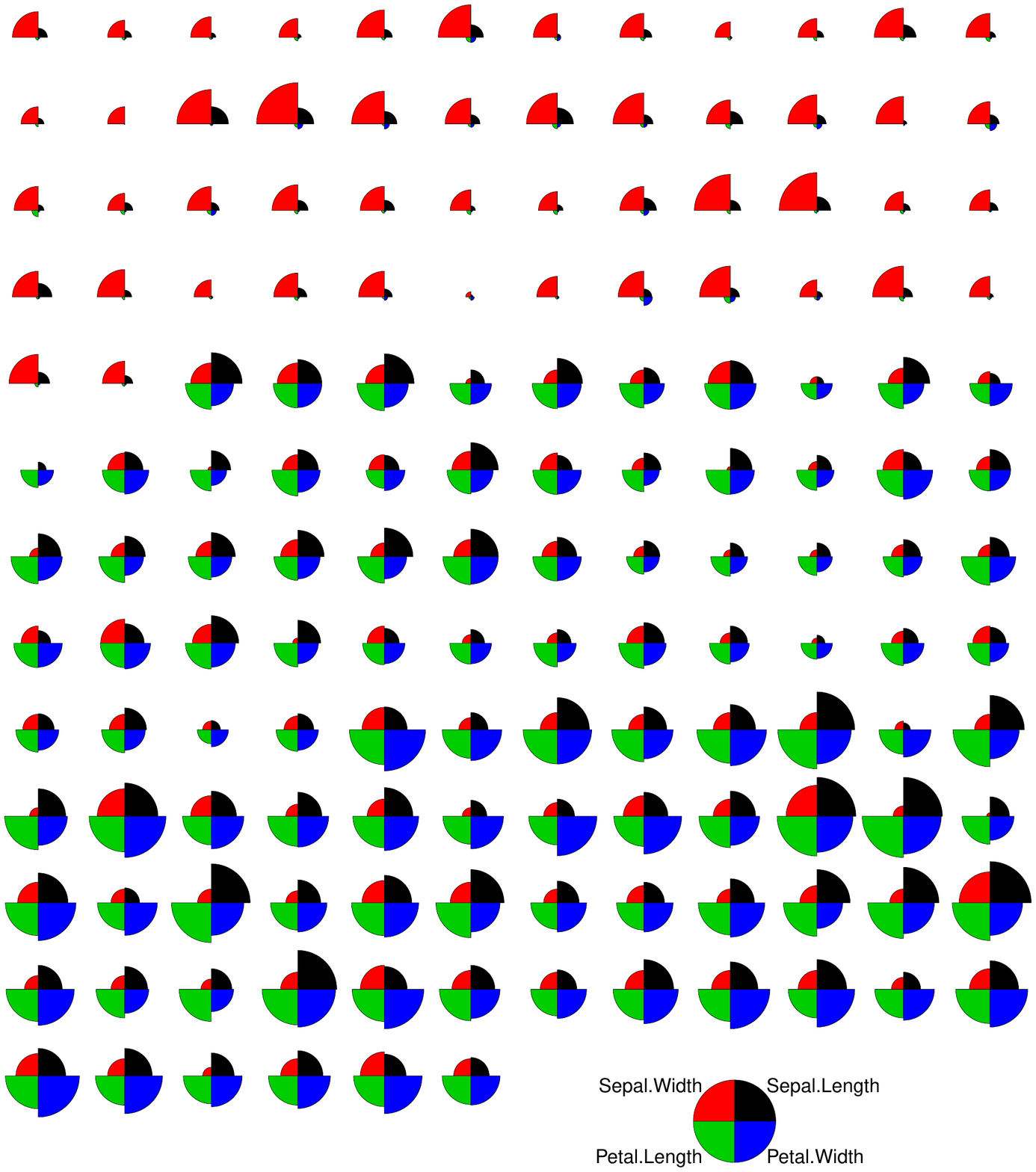}}
			\caption{Segment plot}
			\label{fig:segmentplot}
		\end{subfigure}
		\begin{subfigure}{0.48\linewidth}
			\centering
			\frame{\includegraphics[width=\linewidth]{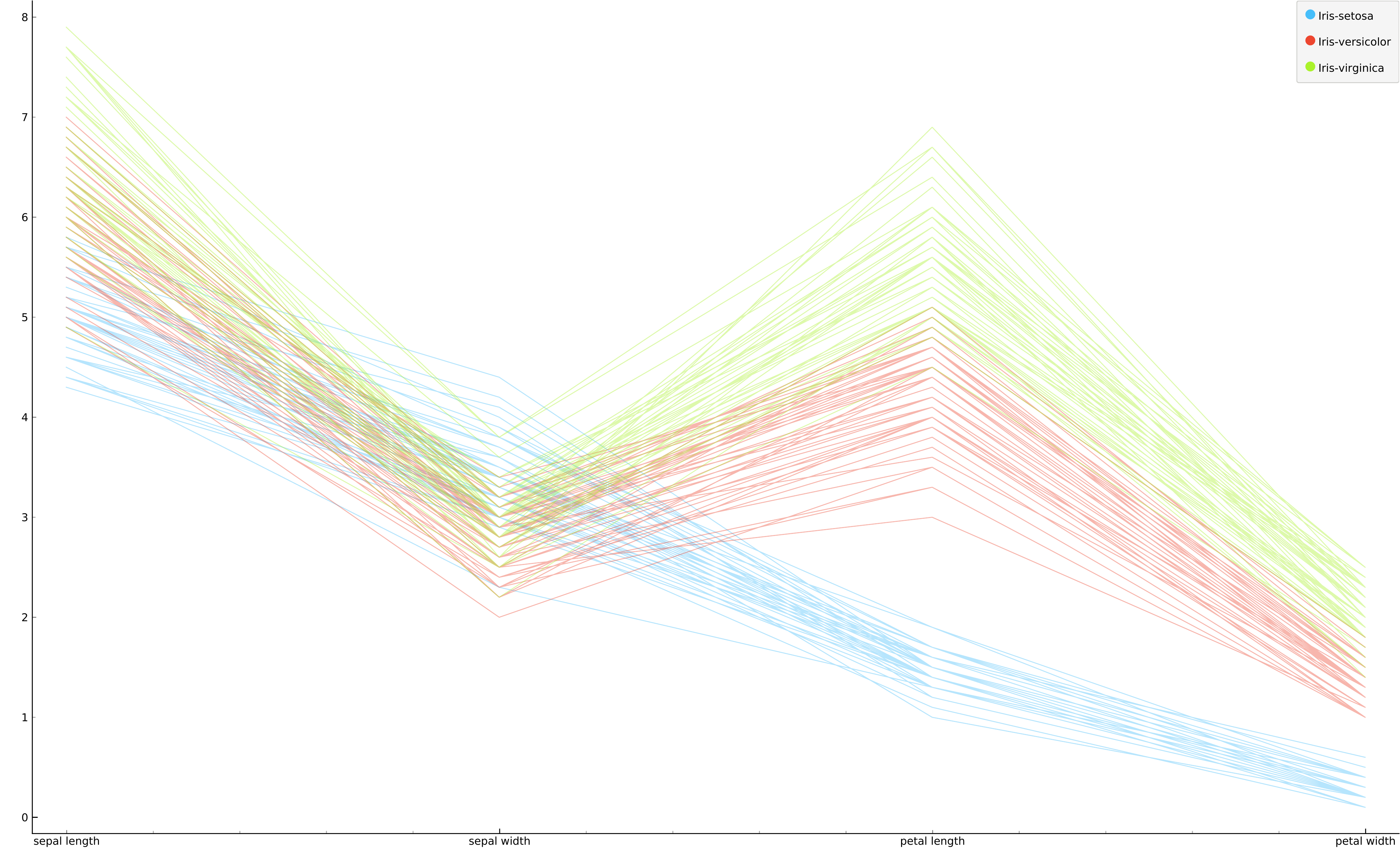}}
			\caption{Parallel coordinates}
			\label{fig:parallelCoords}
		\end{subfigure}
		\caption{Sample dimension-preserving visualizations using the Iris dataset. The dataset contains 150 instances, 4 attributes, and 3 classes.}
		\label{fig:visualizations2}
	\end{figure*}

	The simplest visualization technique that preserves dimensionality is arguably the star plot, which visualizes each dimension of a candidate solution along a radial axes. The length of the ``star" along each axis corresponds to the value in that particular dimension. While the star plot does not reduce the number of dimensions for visualizations, it becomes unwieldy as the number of problem dimensions increases. An example of star plots are given in Figure \ref{fig:starplot}.
	
	Similar to the star plot, the segment plot segments a circle into equally-spaced wedges, one for each problem dimension, with the relative size of each wedge corresponding to the value in that dimension\footnote{Not to be confused with the pie chart, where the relative width of the wedges indicates a percentage.}. However, the segment plot also suffers from the same limitation as the star plot -- it is inherently limited by the number of divisions of a circle that can be meaningfully perceived. Examples of segment plots are given in Figure \ref{fig:segmentplot}.
	
 	Another commonly used, dimension-preserving technique is parallel coordinates \cite{Inselberg1985Plane}, which visualizes $n$-dimensional data by aligning each of the coordinate axes in parallel, rather than orthogonal. This allows for arbitrary dimensions to be visualized as a series of line segments between successive axes. An example of parallel coordinates in given in Figure \ref{fig:parallelCoords}. While parallel coordinates are capable of visualizing high-dimensional data, they suffer from an implicit limit on their scalability due to their linear construction. Specifically, visualizing $n$-dimensional data requires $n$ parallel coordinates, which becomes infeasible as $n$ grows large. Furthermore, an additional consideration when employing parallel coordinates is the ordering of the axes, which can lead to drastically different observations being highlighted \cite{Johansson2016Evaluation}.
	
	A more recent visualization approach, designed specifically to visualize genotypes in an evolutionary algorithm, is the diversity and usage (DU) maps \cite{Medvet2018Unveiling}. DU maps visualize both the diversity and utility of each problem dimension as a single point in a 2D matrix, where the rows correspond to problem dimensions and the columns correspond to the search iteration. Two functions, which quantify the diversity and usage, must be supplied by the user and are used to select a colour for each dimension of the problem in each iteration of the search process. A diversity function, which produces a value in the range [0,1], quantifies the diversity of a particular problem dimension. The utility function, also with a range of [0,1], measures the degree to which a particular dimension contributes to the overall solution. Note that, for both functions, different implementations can be used for each problem dimension. While the DU map can be used to gather population-level insight about the performance of an evolutionary search algorithm, it does not produce visualizations of individual candidate solutions.
	
	Another visualization framework that is of particular relevance, but does not pioneer any particular visualization method, is VISPLORE \cite{Khemka2009Visplore}. VISPLORE is a toolkit for visualizing information about the search process of particle swarm optimization at various levels of granularity and includes both dimension-preserving and dimension-reducing visualizations. Most relevant to this study is the visualization of individual particles using density heatmaps, parallel coordinate plots, star plots, and history plots, which correspond to a hierarchical view of information flow between iterations. The particular strength of VISPLORE is that it packages many different visualizations into a coherent, usable toolkit.

	The visualization techniques that are most relevant to the context of this paper are pixel-based approaches, whereby decision variables are visualized as coloured pixels. However, the work in this domain is limited and often context-specific.  Furthermore, these approaches are pixel-oriented and do not necessarily generate a coherent image. Note that, implementations of these techniques are not readily available and, thus, examples are not provided for these techniques. 
	
	One example of a pixel-oriented approach is VisDB \cite{Keim1994VisDB}, which provides a pixel-based visualization that maps database items to pixels based on their respective similarity to the terms specified in a query. While VisDB was proposed as the result of a preliminary study on using pixel-oriented approaches to visualize high-dimensional data, it is inherently tied to visualizing query-related data and thus is not suitable for optimization-oriented visualization tasks.
	
	The recursive pattern visualization \cite{keim1995recursive} provides a generalized recursive process for arranging pixels in the context of visualizing high-dimensional sets of data. However, this framework does not provide a direct mapping from data to pixels itself. Rather, it provides a generalized process for handling the layout of the pixels such that implicit relationships can be visualized effectively. One inherent issue with this process is that it rearranges the data and thus, in an optimization context, would alter the order of the decision variables in the resulting visualization.
	
	The circle segments technique \cite{ankerst1996circle} partitions a circle into $n$ segments, each of which represent a single dimension. Within each segment, data attributes are visualized as a single, coloured pixel. However, this approach also suffers from a similar issue to that of RadVis, where the scalability is inherently related to the segmentation of a circle.
	
	\subsection{Summary of Existing Techniques}
	
	Table \ref{tbl:visualizations} provides a brief summary of the aforementioned visualization techniques, categorized into two groups, namely those with dimensionality reduction and those without dimensionality reduction. Table \ref{tbl:visualizations} also provides, for each visualization technique, the number of dimensions used for visualization and a relative ranking in terms of: (1) scalability in the number of data dimensions; (2) scalability in terms of the number of entities; and (3) their preservation of relationships in the data. Note that, the rankings in Table \ref{tbl:visualizations} are subjective and correspond to the authors' opinion on the scalability capabilities of each technique. 

	It is clear from Table \ref{tbl:visualizations} that both dimension-reducing and dimension-preserving visualizations suffer from their own drawbacks. Techniques with dimension reduction suffer from limitations in their visualization power due to the mapping onto lower-dimensional spaces, while dimensionality-preserving techniques suffer from implicit limitations in their scalability in terms of the number of dimensions. While the dimensionality-preserving, pixel-based approaches address the dimensional scalability limitations, they suffer from a more nuanced issue -- namely that their visualization power is inherently related to the layout of the pixels. Clearly, there is a lack of visualization techniques that strongly address both the dimensionality scaling and relationship preservation aspects of visualization.

	\begin{table}[htbp]
		\caption{Summary of existing visualization techniques. Various characteristics are (subjectively) ranked, where *** indicates the best ranking. Legend: DR -- dimensionality reduction, DP -- dimensionality preserving, Dim. -- dimensionality, Scal. (Dim.) -- scalability with respect to dimensionality, Scal. (Ent.) -- scalability with resepect to the number of entities, Rel. Pres. -- relationship preservation.}
		\begin{center}
			\scriptsize
			\begin{tabular}{|c|l|c|c|c|c|}
				\hline
	         \textbf{Type}           & \textbf{Technique}                             & \textbf{Dim.} & \textbf{Scal. (Dim.)} & \textbf{Scal. (Ent.)} & \textbf{Rel. Pres.} \\
				\hline
				\multirow{4}{*}{DR}  & Scatterplot Matrix                             &         $2$, n          &             **             &            ***             &                 *                  \\
	                                 & SOM \cite{Kohonen1990Self}                     &         $\leq3$         &            ***             &            ***             &                 **                 \\
	                                 & MDS \cite{torgerson1952multidimensional}       &         $\leq3$         &            ***             &            ***             &                 *                  \\
	                                 & t-SNE \cite{maaten2008visualizing}             &         $\leq3$         &            ***             &            ***             &                 **                 \\
	                                 & RadVis \cite{hoffman1997dna}                   &           $2$           &             **             &            ***             &                 **                 \\
	                                 & 3D RadVis \cite{ibrahim20163d}                 &           $3$           &             **             &            ***             &                 **                 \\
				\hline
				\multirow{3}{*}{DP} & Star Plot                                      &           $n$           &             **             &             *              &                ***                 \\
	                                 & Segment Plot                                   &           $n$           &             **             &             *              &                ***                 \\
	                                 & Parallel Coordinates \cite{Inselberg1985Plane} &           $n$           &             **             &             *              &                 **                 \\
	                                 & DU Maps \cite{Medvet2018Unveiling}             &           $n$           &             **             &             **             &                 *                  \\
	                                 & VisDB \cite{Keim1994VisDB}                     &           $n$           &             **             &             **             &                 **                 \\
	                                 & Recursive Pattern \cite{keim1995recursive}     &           $n$           &            N/A             &            N/A             &                N/A                 \\
	                                 & Circle Segments \cite{ankerst1996circle}       &           $n$           &             **             &            ***             &                 **                 \\
				\hline
	       Proposed         & Image-Based Visualization                      &           $n$           &            ***             &             *              &                ***                 \\
				\hline
			\end{tabular}
			\label{tbl:visualizations}
		\end{center}
	\end{table}
	
	To address the limitations imposed by existing visualization techniques, this paper proposes an image-based visualization framework for large-scale global optimization problems. The general goal of the framework is to facilitate the visualization of arbitrary solutions to arbitrary benchmark problems using images. For the purposes of this study, a concrete problem type, specifically image reproduction, is considered. The use of images as a visualization mechanism introduces a higher-level of visualization capabilities in comparison to pixel-based approaches and allows for semantic embedding of solution quality using standard image-processing techniques. For example, the image itself can represent the overall quality of the solution while the pixels can be used to visualize the optimization variables. To the best of the authors' knowledge, this is the first time that the inherent relationship between decision space and objective space is realized in a dimensionality-preserving, scalable visualization technique. However, it should be noted that scalability, in our proposed framework, refers to scalability in terms of the number of problem dimensions, not necessarily the number of entities that can be visualized.
	
	The remainder of this paper is structured as follows. Section \ref{sec:background} provides a brief description of the optimizers used in this study. Section \ref{sec:proposal} introduces the proposed image-based visualization framework and Section \ref{sec:experimental-design} describes the experimental design. Section \ref{sec:case-studies} then exemplifies the flexibility of image-based visualization via a number of different mapping schemes applied to large-scale global optimization. Section \ref{sec:known-optima} examines the use of image-based visualization on arbitrary benchmark problems with known optima. Finally, Section \ref{sec:conclusions} provides concluding remarks and avenues of future work.
	
\section{Background}
    \label{sec:background}
    
    This section provides a brief introduction to the optimization techniques used in this study. It should be noted that these algorithms are used as a proof-of-concept, rather than to provide an empirical comparison of their performance.
		
	\subsection{Particle Swarm Optimization (PSO)}
		\label{sec:pso}
	
		The PSO algorithm \cite{Kennedy1995Particle} consists of a collection of particles, which each represents a candidate solution to an optimization problem. Each particle retains three pieces of information, namely its current position, its velocity, and its (personal) best position found within the search space. Particle positions are updated each iteration by calculation and subsequent addition of a velocity vector to the particle's current position. 
		
		A particle's velocity is influenced by the attraction towards two promising locations in the search space, namely the best position found by the particle itself and the best position found by any particle within the particle's neighbourhood \cite{Kennedy2002Population}, in addition to a momentum term. The neighbourhood of a particle is defined as the other particles within the swarm from which it may take influence, most commonly the entire swarm, or the two immediate neighbours when the particles are arranged in a ring structure \cite{Kennedy2002Population}. 
		
		According to the inertia weight model \cite{Shi1998Modified}, the velocity is calculated for particle $i$ as
		\begin{equation}
			\label{eq:pso:velocity}
			\begin{split}
				v_{ij}(t+1)& = \omega v_{ij}(t) + c_1 r_{1ij}(t)(y_{ij}(t) - x_{ij}(t)) \\
				&+ c_2 r_{2ij}(t)(\hat{y}_{ij}(t) - x_{ij}(t)) ,
			\end{split}
		\end{equation}
		where $v_{ij}(t)$ and $x_{ij}(t)$ are the velocity and position in dimension $j$ at time $t$, respectively. The inertia weight is given by $\omega$, while $c_1$ and $c_2$ represent the cognitive and social acceleration coefficients, respectively. The stochastic component of the algorithm is provided by the random values $r_{1ij}(t), r_{2ij}(t) \sim U(0,1)$, which are independently sampled each iteration for all components of each particle's velocity. Finally, $y_{ij}(t)$ and $\hat{y}_{ij}(t)$ denote the personal and neighbourhood best positions in dimension $j$, respectively. Particle positions are then updated according to
		\begin{equation}
			\label{eq:pso:update}
			x_{ij}(t+1) = x_{ij}(t) + v_{ij}(t+1).
		\end{equation}
	
	\subsection{Differential Evolution (DE)}
		\label{sec:de}
	
		DE \cite{Storn1997Differential} is an evolutionary optimization algorithm that iteratively improves a population of candidate solutions, referred to as individuals. Individuals within the population, initially placed at random positions in the feasible search space, are updated using mutation, crossover, and selection operations. In the DE algorithm, a trial position is created for each individual through recombination; a trial position represents a potential new position for the individual. Creation of the trial position $\vec{t}$ for an individual $\vec{x}$ in dimension $i$ using the DE/rand/1/bin strategy \cite{Price199New} is given by
		\begin{equation}
			\label{eq:de:trial}
			t_i = 
			\begin{cases}
				a_i + F(b_i - c_i) & \text{if } rand() < CR \text{ or } i = randi(D) \\
				x_i & \text{otherwise}
			\end{cases},
		\end{equation}
		where $\vec{a}$, $\vec{b}$, and $\vec{c}$ are three randomly selected, distinct members of the population that are different from the current individual $\vec{x}$, $F \in [0,2]$ is the user-supplied differential weight, $rand() \sim U(0,1)$, $randi(D)$ selects a uniform random integer in the range $[1,D]$, $D$ is the problem dimensionality, and $CR \in [0,1]$ is the user-supplied crossover probability. If the generated trial position improves the individual's fitness, the trial position is accepted and the individuals position is updated accordingly. Otherwise, the trial position is discarded and the individual retains its current position.
	
	\subsection{Third Generalized Differential Evolution (GDE3)}

		As will be shown in Section \ref{sec:case-studies}, the proposed framework can also be applied in a multi-objective context. Therefore, a multi-objective variant of DE is employed to demonstrate this capability. The third version of Generalized Differential Evolution (GDE3) \cite{kukkonen2005gde3} extends the DE algorithm for multi-objective optimization problems with $M$ objectives and $K$ constraints. DE operators are applied using three randomly selected vectors to produce an offspring per parent in each generation. The general idea behind the selection mechanism is based on constraint-domination and the crowding distance measure. Before describing the selection strategy in GDE3, a number of concepts from multi-objective optimization must first be introduced.
		
		Given that the fitness of a multi-objective problem is a vector, an alternative way of comparing solutions is needed. In a multi-objective context, a decision vector $\vec{x}_1$ is said to dominate another decision vector $\vec{x}_2$, denoted by $\vec{x}_1 \prec \vec{x}_2$, if and only if
		
		\begin{subequations}
			\label{eq:dominance}
			\begin{equation}
				\forall o = 1,\ldots, n_o: f_o(\vec{x}_1) \leq f_o(\vec{x}_2)
			\end{equation}
			and
			\begin{equation}
				\exists o = 1, \ldots, n_o: f_o(\vec{x}_1) < f_o(\vec{x}_2).
			\end{equation}
		\end{subequations}
		In other words, a candidate solution $\vec{x}_1$ dominates $\vec{x}_2$ if $\vec{x}_1$ is at least as good as $\vec{x}_2$ in all objectives and is strictly better in at least one objective. The set of all non-dominated decision vectors is referred to as the Pareto set while the corresponding objective vectors formulate the Pareto front.
		
		Solutions can then be sorted, and assigned a rank, based on the number of other solutions that they dominate or, alternatively, by the number of other solutions that they are dominated by.
		
		Crowding distance is another metric used to compare solutions along with the dominance relation. Crowding distance is a measure that quantifies the diversity of the obtained solutions by calculating the distance between neighbouring solutions. Initially, the set of solutions with the same non-dominated rank are sorted according to each objective value in ascending order. In order to calculate crowding distance, the difference between the objective values of neighboring solutions are computed as \cite{deb2002fast}
		\begin{equation}
			\label{eqc}
			CD_{i}=\sum^{M}_{j=1} |f_{j}(i+1)-f_{j}(i-1)|.
		\end{equation}
		
		For each objective, the boundary solutions (with lowest and highest objective values) are assigned an infinite crowding distance values.
		
		The selection rules in GDE3 are as follows: when an original and a new trial vector are both infeasible solutions, the new trial vector is selected if it dominates the original vector in the constraint violation space, otherwise, the original vector is selected. In the case that one of them is a feasible vector, the feasible vector is always selected. If both vectors are feasible, the dominating vector is selected for the next generation. In the case where they are non-dominated with respect to each other, both vectors are selected. Therefore, the size of the next generation may be larger than the population of the current generation. In this case, the population is pruned using a selection strategy similar to that of NSGA-II \cite{deb2002fast}. First, the individuals in the population are sorted using non-dominated sorting, then according to the crowding distance measure. Similar to other population-based multi-objective algorithms, the selected individuals are passed to the next generation to continue the optimization process.

\section{Image-Based Visualization For Large-Scale Global Optimization}
	\label{sec:proposal}

    A primary motivation for image-based visualization is the inherent scalability with respect to problem dimensionality; it is not unreasonable to have an image with 1,000,000 pixels (i.e., 1 megapixel). In fact, an image of this size would be generally considered to have low resolution. However, many existing visualization techniques, which do not use dimensionality reduction, cannot reasonably visualize dimensions of this magnitude. For example, parallel coordinates would require 1,000,000 parallel axes while RadVis would require dividing the circumference of a circle into 1,000,000 segments. Clearly, even existing dimensionality preserving techniques degrade rapidly with increasing dimensionality. In contrast, visualization of 1,000,000 dimensions using image-based visualization requires only an image with 1,000,000 pixels (e.g., a size of 1000x1000 or 2000x500). Furthermore, the introduction of colour and other image features permits even greater scalability and visualization capabilities. Thus, the problematic ``curse of dimensionality", with respect to visualization, is effectively mitigated by this framework. However, it should be noted that this technique is best equipped to address problems with a composite dimensionality, i.e., those with dimension $D = w \times h, \enskip w, h > 1$. Nonetheless, by omitting some pixels from the target image (i.e., having an incomplete row/column in the image), this technique can be employed on problems with arbitrary dimensionality.
	
	In addition to the scalability, image-based visualization affords many other advantages over existing visualization techniques. Notably, one of the fundamental properties that the visualization function $F$ must exhibit is that it must align with human perception. That is, a non-expert practitioner should be able to reasonably perceive the visualization of a superior solution as being a higher-quality image. Moreover, the introduction of enhanced visualization capabilities facilitates interactive optimization, whereby a user can be alerted to hindrances in the optimization process in real-time, such as stagnation, and react accordingly. Similarly, image-based visualization can be trivially extended to produce visualizations of the optimization process over time by encoding the resulting images as frames in a video.

    \subsection{Formal Definition}
    
\begin{figure}[ht!]
    		\centering
    		\includegraphics[width=0.6\linewidth]{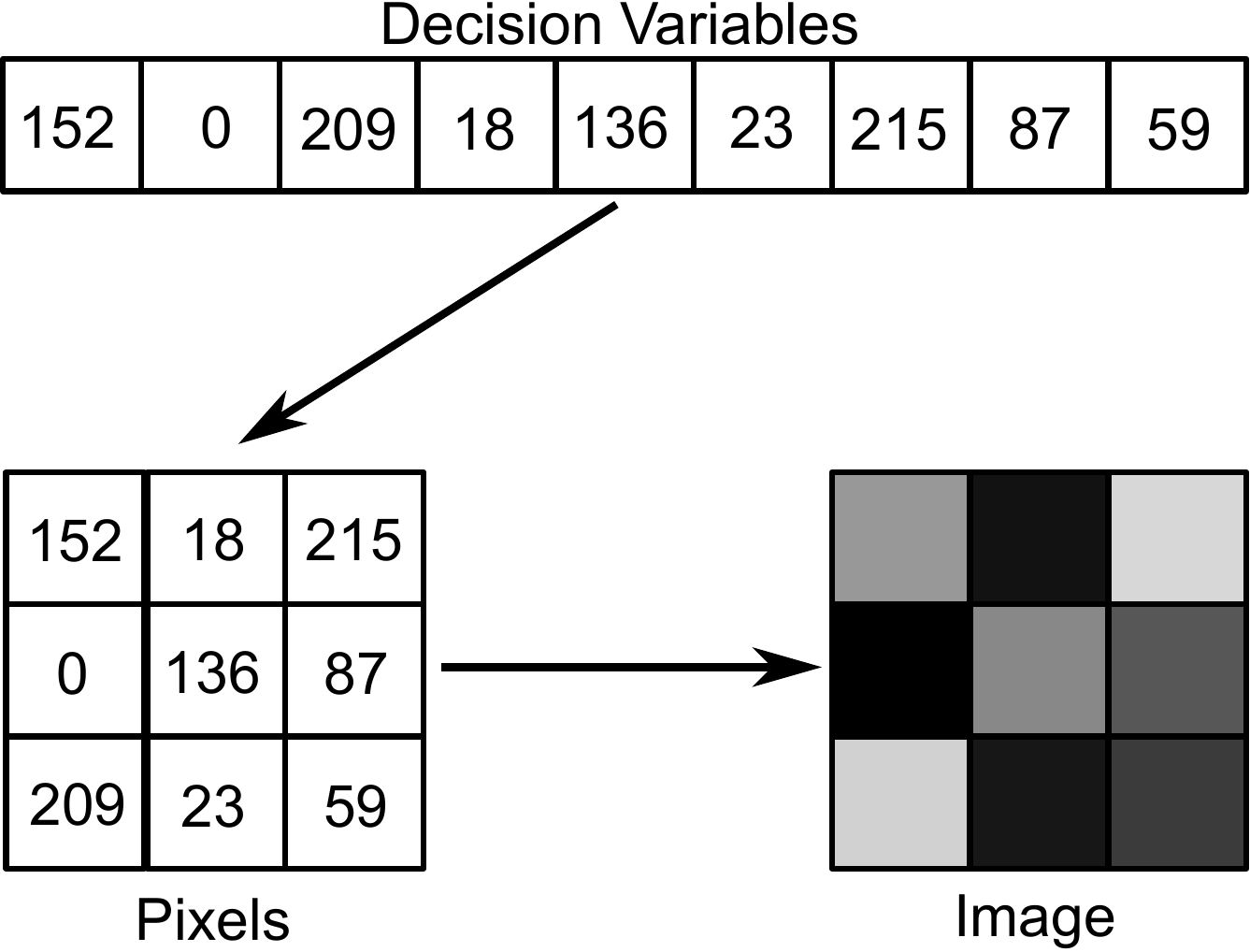}
    		\caption{Example mapping of a candidate solution to an image.}
    		\label{fig:mapping}
    	\end{figure}

    	Given an arbitrary optimization problem $\pi$ and candidate solution $\vec{s}$, the goal of image-based visualization is to devise a function, $F$, which produces an image, $I$, that is representative of the relative quality of $\vec{s}$ with respect to $\pi$. More concretely, the general goal is to devise a function given by
    	\begin{equation}
    		\label{eq:solution-unknown}
    		F(\pi,\vec{s}) \mapsto I.
    	\end{equation}
    	In the case where an optimal solution, $\vec{s}*$, is known \textit{a priori}, the objective can be further refined such that the goal is to devise a function given by
    	\begin{equation}
    		\label{eq:solution-known}
    		F(\pi,\vec{s},\vec{s}*) \mapsto I,
    	\end{equation} 
    	which produces an image that is representative of the absolute quality of $\vec{s}$ in comparison to the optimal solution $\vec{s}*$. Furthermore, $F$ also induces a mapping from decision-space to individual pixels, such that both decision space variables and the overall objective quality of $s$ is embedded within the resulting image, $I$. An example demonstrating how the decision variables are used to formulate an image is provided in Figure \ref{fig:mapping}. Effectively, a (linear) candidate solution is reformulated as a matrix and the decision variables are then mapped to the corresponding pixels. 
    	
    	Given that the core of image-based visualization lies at the intersection of optimization and image-processing, there are a number of additional benefits that arise from this framework. For example, image-based visualization facilitates the use of image-processing metrics as objective functions to construct new optimization problems. Similarly, it allows for the plethora of readily-available images to be used as benchmark cases. In addition, the flexibility of image processing capabilities facilitates the use of this framework on many different optimization problem types. Thus, the benefits of image-based visualization are numerous. 

	\subsection{Mapping Function for an Image Reconstruction Problem}
	    \label{sec:image-mapping}
	
		In this section, a simple mapping function is presented whereby the optimization objective was to replicate a target image according to some image similarity metric. Thus, the objective fitness function corresponds to the minimization of an image comparison metric that would quantify the difference between a candidate solution and a target image. This problem formulation was chosen as it permits a mapping function that is a direct encoding the candidate solution, $\vec{s}$, as the corresponding pixel values, formally defined by
    	\begin{equation}
    		\label{eq:mapping}
    		F(\pi, \vec{s}) \mapsto A \in \mathbb{R}^{h\times w}\enskip | \enskip A_{ij} = s[j\times h + i], 
    	\end{equation} 
    	where $A$ is a matrix of pixel greyscale intensity values with height $h$ and width $w$, taken as the dimensions of the target image such that $D = h \times w$. Thus, the pixel values directly visualize the decision variables of the problem while the resulting images visualize the overall solution quality. 
    	
    	While the mapping function given in Equation (\ref{eq:mapping}) is only relevant to a specific, direct application of the proposed visualization framework, more complex mapping functions that handle arbitrary optimization problems are certainly possible. For example, the next section describes a mapping function that is suitable for an arbitrary objective function, with a single known optimum.
	
    \subsection{Mapping Function for an Arbitrary Objective Function with a Known Optimum}
        \label{sec:function-mapping}
    
    	In this section, an example of a more complex mapping function, applicable to a much wider variety of optimization problems, is presented. This mapping function can be applied to any real-valued, boundary-constrained problem with a single, known optimum. The general concept is similar to that of Equation (\ref{eq:mapping}) in that each decision variable will correspond to a single pixel in the output image. However, the decision variable itself no longer directly encodes the pixel's colour. Rather, the output image will be constructed by normalizing and linearly scaling the error in each dimension of the candidate solution. Thus, decision variables that are less than the value of the optimal solution will be ``darkened" (i.e., have a greyscale value closer to 0), while variables that are greater than the value of the optimal solution will be ``lightened" (i.e., have a greyscale value closer to 1) as follows
    
    	\begin{equation}
    		\label{eq:linearMapping}
    		F(\pi, \vec{s}) \mapsto A \in \mathbb{R}^{h\times w}\enskip | \enskip A_{ij} = 
    		\begin{cases}
    			T_{ij} + \frac{e_{ij}}{u_{ij} - o_{ij}} (1 - T_{ij}) & \text{if } e_{ij} > 0 \\
    			T_{ij} - \frac{e_{ij}}{o_{ij} - l_{ij}} (T_{ij})     & \text{if } e_{ij} < 0 \\
    			T_{ij}                                               & \text{if } e_{ij} = 0
    		\end{cases}
    	\end{equation} 
    	where $T$ is the target image\footnote{Note that, the subscripts $ij$, corresponding to index $ij$ in the image matrix, are removed for brevity}, $s$ is the candidate solution, $o$ is the optimal solution, $e = s - o$ is the error between the candidate solution and the optimal solution, and $l$ and $u$ refer to the lower and upper bounds of the problem, respectively. As in Equation (\ref{eq:mapping}), the index $ij$ in the image matrix refers to dimension $j\times h + i$ in the candidate solution. 
    	
    	Effectively, this mapping scheme linearly scales the error in each dimension (i.e., pixel), based on the distance between the candidate solution and the optimal solution, using the greyscale value of the target pixel as the reference and the problem bounds as the extremes. Note that, the (positive) error term is maximized, for a particular dimension, when the candidate solution lies on the upper boundary of the problem in that dimension -- this corresponds to a white pixel. Conversely, the error term for a particular dimension is minimized when the candidate solution lies on the lower boundary of the problem in that dimension -- this corresponds to a black pixel. When the candidate solution equals the optimal solution in a given dimension, the corresponding pixel colour matches that of the target image. Any other value for a decision variable is encoded as a shade of grey, depending on its proximity to the optimal value. 
    	
\begin{figure}
    		\centering
    		\includegraphics[width=\linewidth]{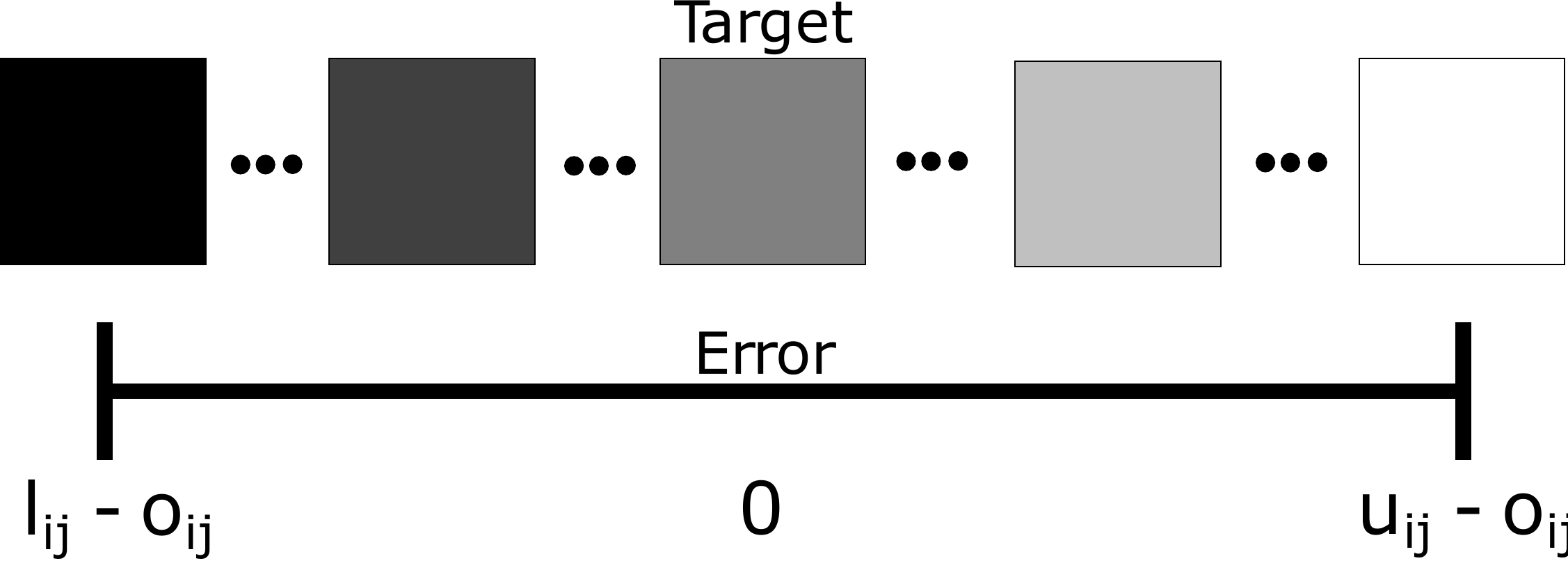}
    		\caption{Example visualization showing the linear mapping scheme for a single pixel.}
    		\label{fig:linearMapping}
    	\end{figure}

\begin{figure}
    		\centering
    		\begin{subfigure}{0.24\linewidth}
    			\centering
    			\frame{\includegraphics[width=\linewidth]{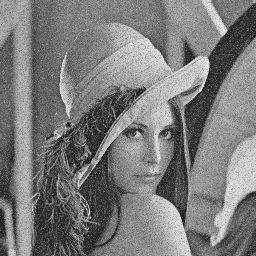}}
    			\caption{$\mathcal{N}(0, 0.5)$, fitness = 16455.3 }
    			\label{fig:good}
    		\end{subfigure}
    		\begin{subfigure}{0.24\linewidth}
    			\centering
    			\frame{\includegraphics[width=\linewidth]{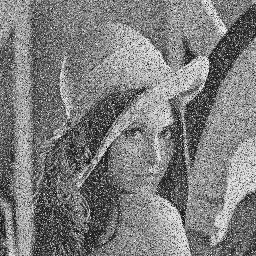}}
    			\caption{$\mathcal{N}(0, 1.5)$, fitness = 146781.4}
    			\label{fig:bad}
    		\end{subfigure}
    		\begin{subfigure}{0.24\linewidth}
    			\centering
    			\frame{\includegraphics[width=\linewidth]{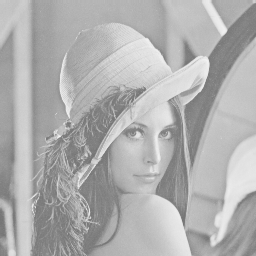}}
    			\caption{Vector of twos, fitness = 262144.0}
    			\label{fig:ones}
    		\end{subfigure}
    		\begin{subfigure}{0.24\linewidth}
    			\centering
    			\frame{\includegraphics[width=\linewidth]{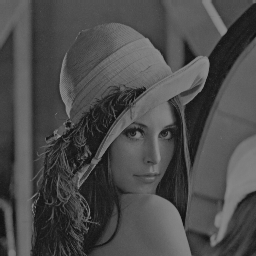}}
    			\caption{Vector of negative twos, fitness = 262144.0}
    			\label{fig:negones}
    		\end{subfigure}
    		\begin{subfigure}{0.24\linewidth}
    			\centering
    			\frame{\includegraphics[width=\linewidth]{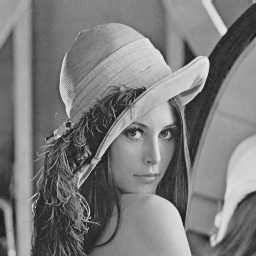}}
    			\caption{Optimal, fitness = 0.0}
    			\label{fig:optimal}
    		\end{subfigure}
    		\caption{Example visualizations produced by the linear mapping}
    		\label{fig:LinearMappingExamples}
    	\end{figure}

    	This mapping technique is visualized in Figure \ref{fig:linearMapping} and various examples, corresponding to the Spherical function (optimum at $\vec{0}$ with bounds of $[-5.12, 5.12]^{65,536}$), are provided in Figure \ref{fig:LinearMappingExamples}. To produce the visualizations shown in Figures \ref{fig:good} and \ref{fig:bad}, two solutions were generated using a normal distribution with a mean of 0 and standard deviations of 0.5 and 1.5, respectively. Clearly, the solution with the lower standard deviation (and hence closer proximity to the optimum), has less visual noise in the resulting image. Figures \ref{fig:ones} and \ref{fig:negones} depict solutions that are simply shifted from the optimum, thereby evidencing how values above the optimum produce lighter pixels (Figure \ref{fig:ones}) and values below the optimum produce darker pixels (Figure \ref{fig:negones}). Finally, Figure \ref{fig:optimal} demonstrates that applying this mapping to the optimum produces the target image. A key observation is that this mapping function preserves the relationship between solution quality and image quality, but is far more broadly applicable than the scheme presented in Section \ref{sec:case-studies}. However, there is an implicit assumption that solution quality can be quantified by proximity to the optimal solution, which may not necessarily be true for multi-modal problems. Nonetheless, this mapping function provides evidence that more complex, and broadly applicable, mapping functions can be devised in the proposed framework. 
    	
    	It should be explicitly noted that the mapping function provided by Equation (\ref{eq:linearMapping}) is only one particular example from a much larger study regarding the application of image-based visualization to arbitrary problems and, therefore, the remainder of this study does not consider it as comprehensively as the mapping function given in Equation (\ref{eq:mapping}). Rather, experimentation making use of Equation (\ref{eq:linearMapping}) is included largely as a proof-of-concept and to demonstrate the advanced capabilities of the proposed framework.

\section{Experimental Design}
	\label{sec:experimental-design}

    This section provides background information on the image evaluation metrics and optimizers used in this study.

	\subsection{Fitness Evaluation Metrics}
		\label{sec:fitness}
		
\begin{figure}
				\centering
				\begin{subfigure}{0.49\linewidth}
					\centering
					\includegraphics[width=\linewidth]{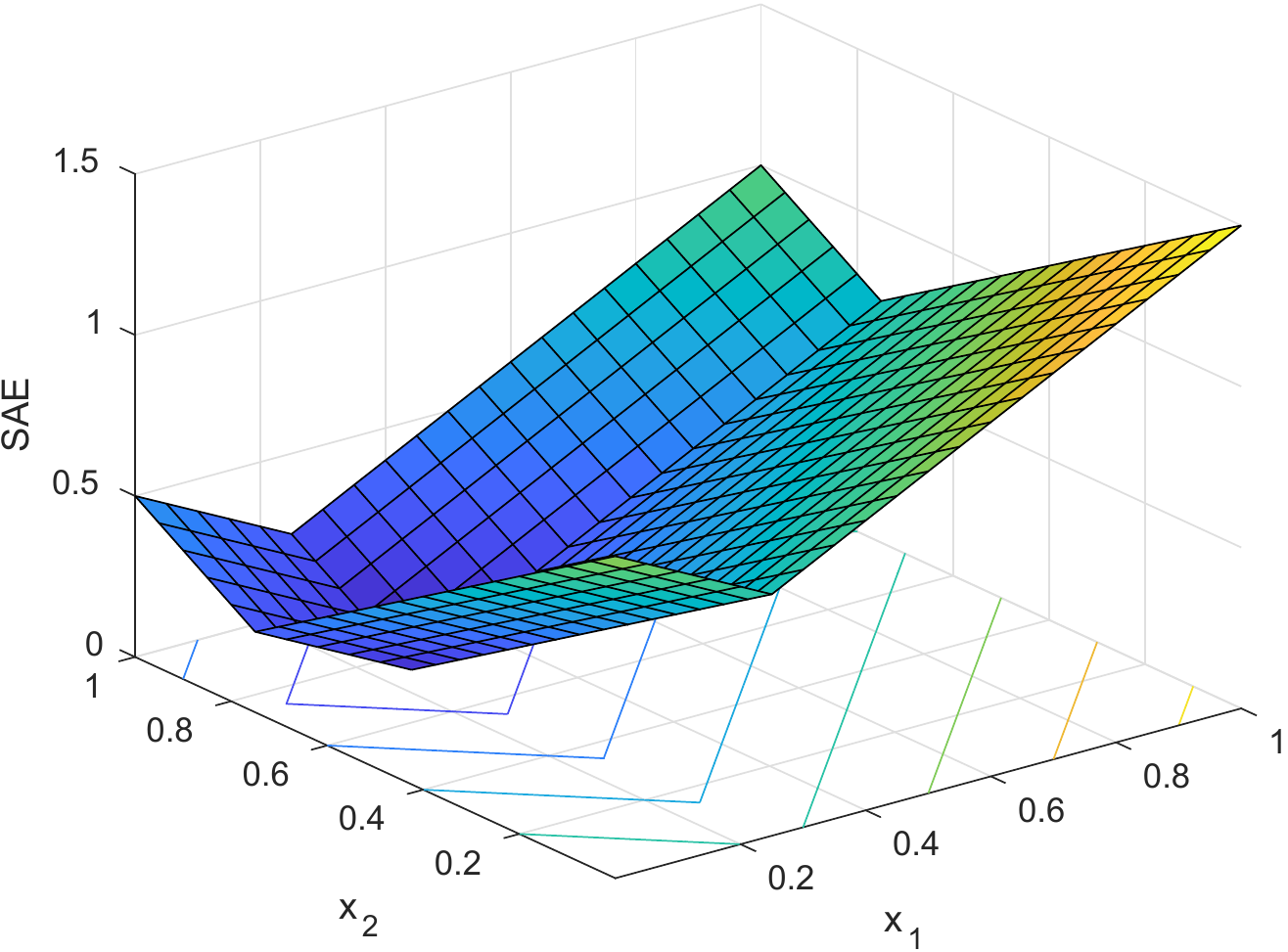}
					\caption{SAE}
					\label{fig:SAE}
				\end{subfigure}
				\begin{subfigure}{0.49\linewidth}
					\centering
					\includegraphics[width=\linewidth]{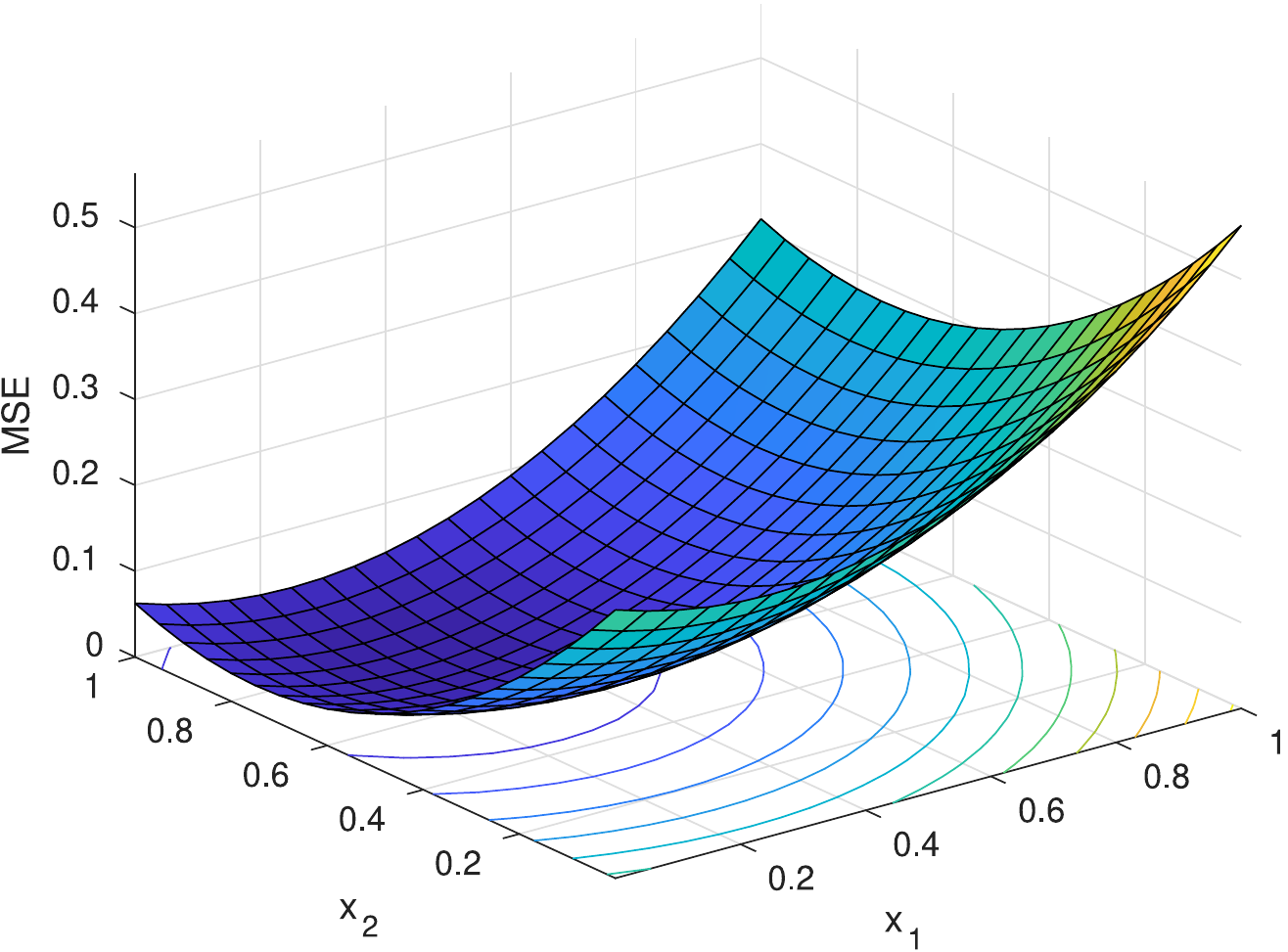}
					\caption{MSE}
					\label{fig:MSE}
				\end{subfigure}
				\begin{subfigure}{0.49\linewidth}
					\centering
					\includegraphics[width=\linewidth]{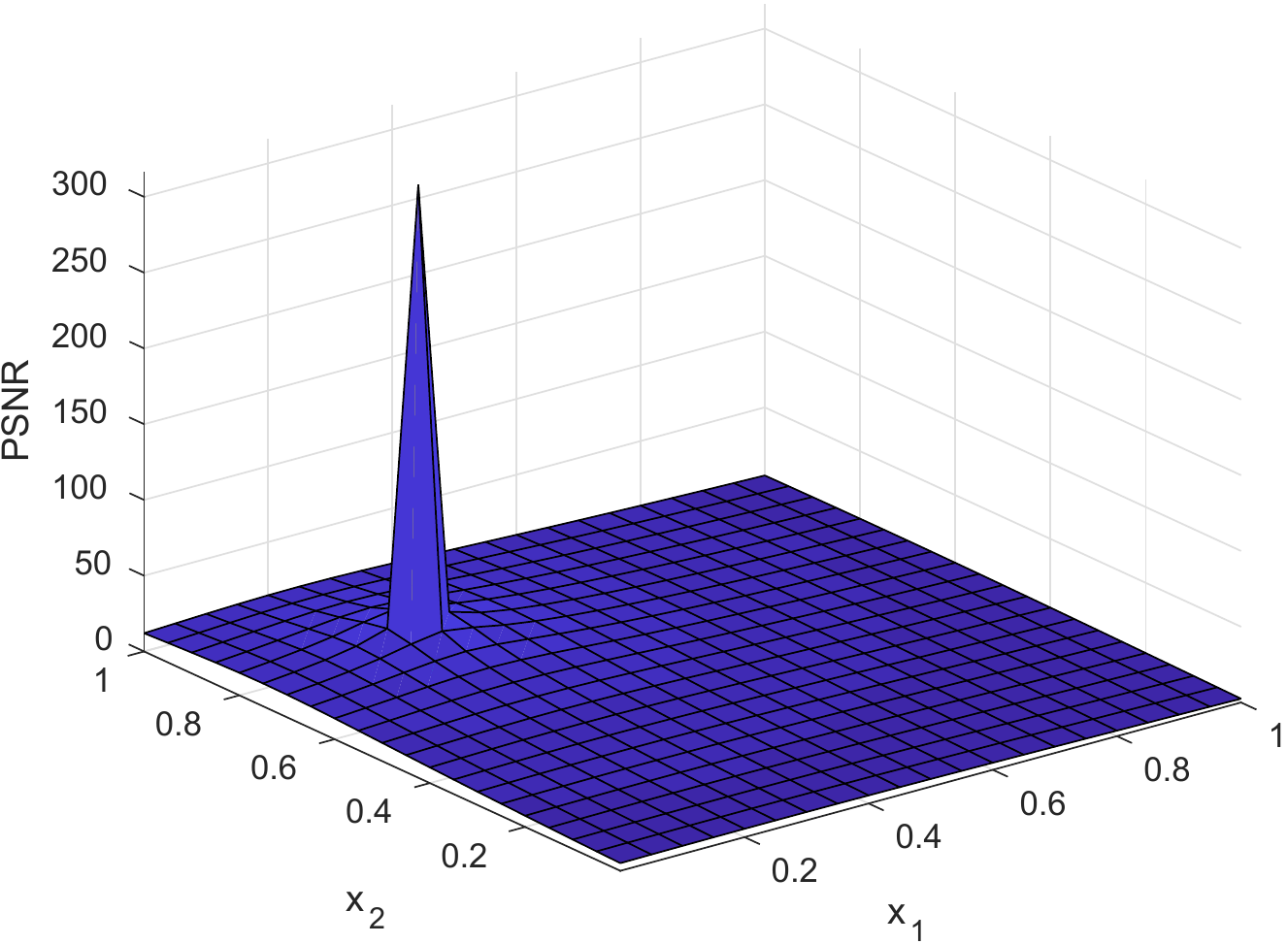}
					\caption{PSNR}
					\label{fig:PSNR}
				\end{subfigure}
				\begin{subfigure}{0.49\linewidth}
					\centering
					\includegraphics[width=\linewidth]{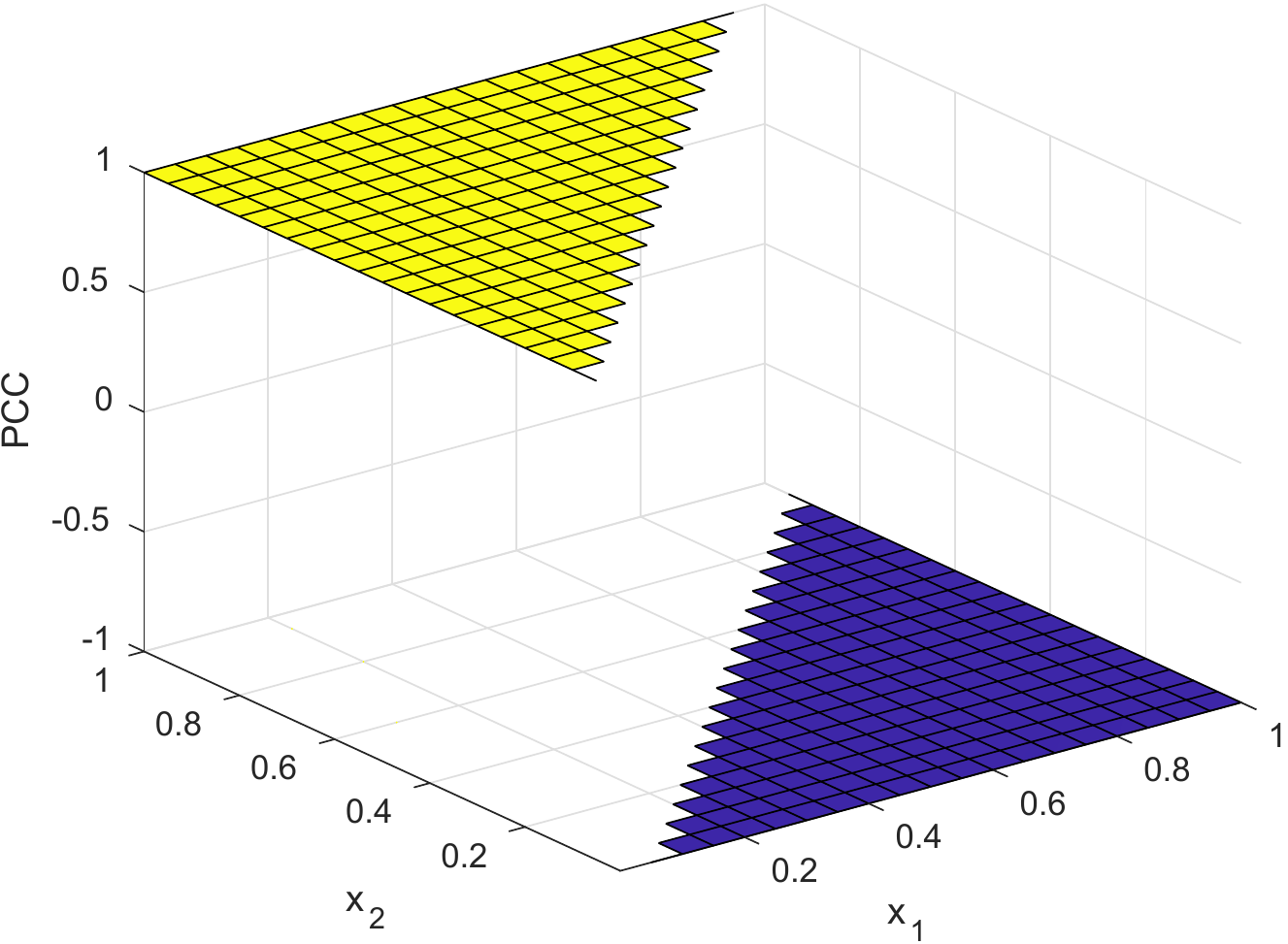}
					\caption{PCC}
					\label{fig:PCC}
				\end{subfigure}
				\caption{Visualization of the 2D landscape for each examined fitness function, with the exception of SSIM. The target solution, $B$, is located at $(0.25, 0.75)$.}
				\label{fig:measures}
			\end{figure}

		This section briefly describes the image comparison metrics that were used as objective fitness measures, assuming that the mapping function provided in Equation (\ref{eq:mapping}) was employed. For all examined fitness measures, it is assumed that $A$ and $B$ are $m \times n$ matrices representing two images, respectively, such that the entries $A_{ij}$ and $B_{ij}$ correspond to the pixels at location $(i,j)$ in images $A$ and $B$. Each metric imposes different characteristics on the landscape of the search process, as exemplified in Figures \ref{fig:measures} and \ref{fig:SSIM}, which presents the 2D fitness landscape for each measure with a target solution of $B = (0.25, 0.75)$. Note that, the landscape visualizations correspond to a target image with two pixels, specifically with greyscale values of 0.25 and 0.75, respectively and corresponds to a purely synthetic scenario constructed only to assist with visualizing the different landscapes induced by the various image comparison metrics. Each point in the landscape corresponds to some other two pixel image, while the height corresponds to the image comparison metric between that point (i.e., image) and the target image $B$. It should also be noted that the landscapes in Figures \ref{fig:measures} and \ref{fig:SSIM} are for the 2D formulations only, and that the landscape characteristics may change with an increase in dimensionality.
	
		\subsubsection{Sum of Absolute Error}
			\label{sec:sae}
		
			Given two images, the sum of absolute error (SAE) \cite{SAE} is calculated as
			
			\begin{equation}
				\label{eq:sae}
				\argmin_A SAE(A,B) = \sum_{i=1}^{m}\sum_{j=1}^{n} |A_{ij} - B_{ij}|,
			\end{equation}	
			whereby the SAE corresponds to sum of absolute differences between each pixel in the candidate solution and the corresponding pixel in the target image. A value of 0 for the SAE indicates an exact replica of the target image. The 2D landscape for the SAE metric is provided in Figure \ref{fig:SAE}. The SAE metric is both unimodal and fully separable. As the dimensionality increases, the basic characteristics of the landscape for the SAE metric will remain fixed.
			
		\subsubsection{Mean Squared Error}
			\label{sec:mse}
			
			The mean squared error (MSE) \cite{MSE} is the mean of the squared error of each pixel, given by
			\begin{equation}
				\label{eq:mse}
				\argmin_A MSE(A,B) = \frac{1}{mn}\sum_{i=1}^{m}\sum_{j=1}^{n} (A_{ij} - B_{ij})^2,
			\end{equation}
			where an MSE value of 0 indicates an exact replica of the target image. Compared to the SAE measure, MSE is more susceptible to outliers, and thus provides a heavier penalty for large errors, due to the square term in its calculation. The 2D landscape for the MSE metric is provided in Figure \ref{fig:MSE}. The MSE measure is also both fully separable and unimodal, and the characteristics will be similar as dimensionality increases.
			
		\subsubsection{Peak Signal-to-Noise Ratio}
			\label{sec:psnr}
			
			The peak signal-to-noise ratio (PSNR) \cite{PSNR} is used as a measure of quality between two images, typically an original and compressed image. This can be calculated according to
			\begin{equation}
				\label{eq:psnr}
				\argmax_A PSNR(A,B) = 10 \log_{10} \left(\frac{R^2}{MSE(A,B)}\right), 
			\end{equation}
			where $R$ is the fluctuation in the image data type. In the context of this work, $R = 1$ for continuous images and $R=255$ for discrete images. Alternatively, one can specify a value for $R$. Note that, in the case of continuous images, the PSNR is simply a logarithmic function of the inverse of the MSE, which provides a smoothing effect. For the PSNR metric, larger values indicate a higher degree of similarity. The 2D landscape for the PSNR metric is provided in Figure \ref{fig:PSNR} and shows a relatively neutral landscape, with a peak near the optimal solution. The PSNR metric is also fully separable and depicts unimodality in two dimensions. Based on the definition in Equation (\ref{eq:psnr}), these characteristics are expected to hold in higher dimensions.
			
		\subsubsection{2D correlation coefficient}
			\label{sec:correlation}
			
			The 2D (Pearson) correlation coefficient (PCC) \cite{PCC} measures the correlation between two matrices and can be used as an image similarity measure. The 2D correlation coefficient is given by
			
			\begin{subequations}
				\label{eq:pcc}
				\begin{equation}
					\argmax_A r(A,B)
				\end{equation}
				where
				\begin{equation}
					r(A,B) = \frac{\sum\limits_{i=1}^{m}\sum\limits_{j=1}^{n} (A_{ij} - \overline{A})(B_{ij} - \overline{B})}{\sqrt{\left(\sum\limits_{i=1}^{m}\sum\limits_{j=1}^{n} (A_{ij} - \overline{A})^2\right)\left(\sum\limits_{i=1}^{m}\sum\limits_{j=1}^{n} (B_{ij} - \overline{B})^2\right)}},
				\end{equation}
			\end{subequations}
			with a range of values between -1 and 1 where positive values indicate positive correlation and negative values indicate negative correlation. The 2D landscape for the PCC metric is provided in Figure \ref{fig:PCC}, which shows perfect positive correlation for solutions with $x_2 > x_1$ and perfect negative correlation for solutions with $x_1 > x_2$. In only two dimensions, there are no other correlations visible in the landscape, which would not be the case for higher dimensions. Therefore, the landscape characteristics are expected to change in higher dimensions. In contrast to the previous metrics, the PCC measure is non-separable.
			
		\subsubsection{Structural Similarity Index}
			\label{sec:ssim}
		
			The structural similarity index (SSIM) \cite{wang2004image} is a metric for comparing the similarity of two images. In contrast to other measures, the SSIM quantifies perceived changes rather than absolute changes. The SSIM can be calculated according to
			
			\begin{subequations}
				\label{eq:ssim-all}
				\begin{equation}
					\label{eq:ssim}
					\argmax_A SSIM(A,B) = [l(A,B)^\alpha \cdot c(A,B)^\beta \cdot s(A,B)^\gamma],
				\end{equation}
				where $l(A,B)$ is a measure of luminance, given by
				\begin{equation}
					\label{eq:ssim-lum}
					l(A,B) = \frac{2\mu_A\mu_B + c_1}{\mu_A^2 + \mu_B^2 + c_1},
				\end{equation}
					$c(A,B)$ is a measure of contrast, given by
				\begin{equation}
					\label{eq:ssim-con}
					c(A,B) = \frac{2\sigma_A\sigma_B + c_2}{\sigma_A^2 + \sigma_B^2 + c_2},
				\end{equation}
				and $s(A,B)$ is a measure of the structure given by
				\begin{equation}
					\label{eq:ssim-str}
					s(A,B) = \frac{\sigma_{AB} + c_3}{\sigma_A\sigma_B + c_3},
				\end{equation}
			\end{subequations}
			assuming the following definitions:
			\begin{itemize}
				\item $\mu_A$, $\mu_B$ are the averages of $A$ and $B$
				\item $\sigma_A$, $\sigma_B$ are the standard deviations of $A$ and $B$
				\item $\sigma_{AB}$ is the covariance of $A$ and $B$
				\item $L$ is the range of the pixel values
				\item $c_1 = (k_1L)^2$, with $k_1 = 0.01$ by default
				\item $c_2 = (k_2L)^2$, with $k_2 = 0.03$ by default
				\item $c_3 = \frac{c_2}{2}$
			\end{itemize}
			
			If the weights $\alpha$, $\beta$, and $\gamma$ are set to 1, as is the case with this study, the SSIM calculation simplifies to
			\begin{equation}
				\label{eq:ssim-simple}
				SSIM(A,B) = \frac{(2\mu_A\mu_B + c_1)(2\sigma_A\sigma_B + c_2)}{(\mu_A^2 + \mu_B^2 + c_1)(\sigma_A^2 + \sigma_B^2 + c_2)}.
			\end{equation}
		
\begin{figure}
				\centering
				\begin{subfigure}{0.49\linewidth}
					\centering
					\includegraphics[width=\linewidth]{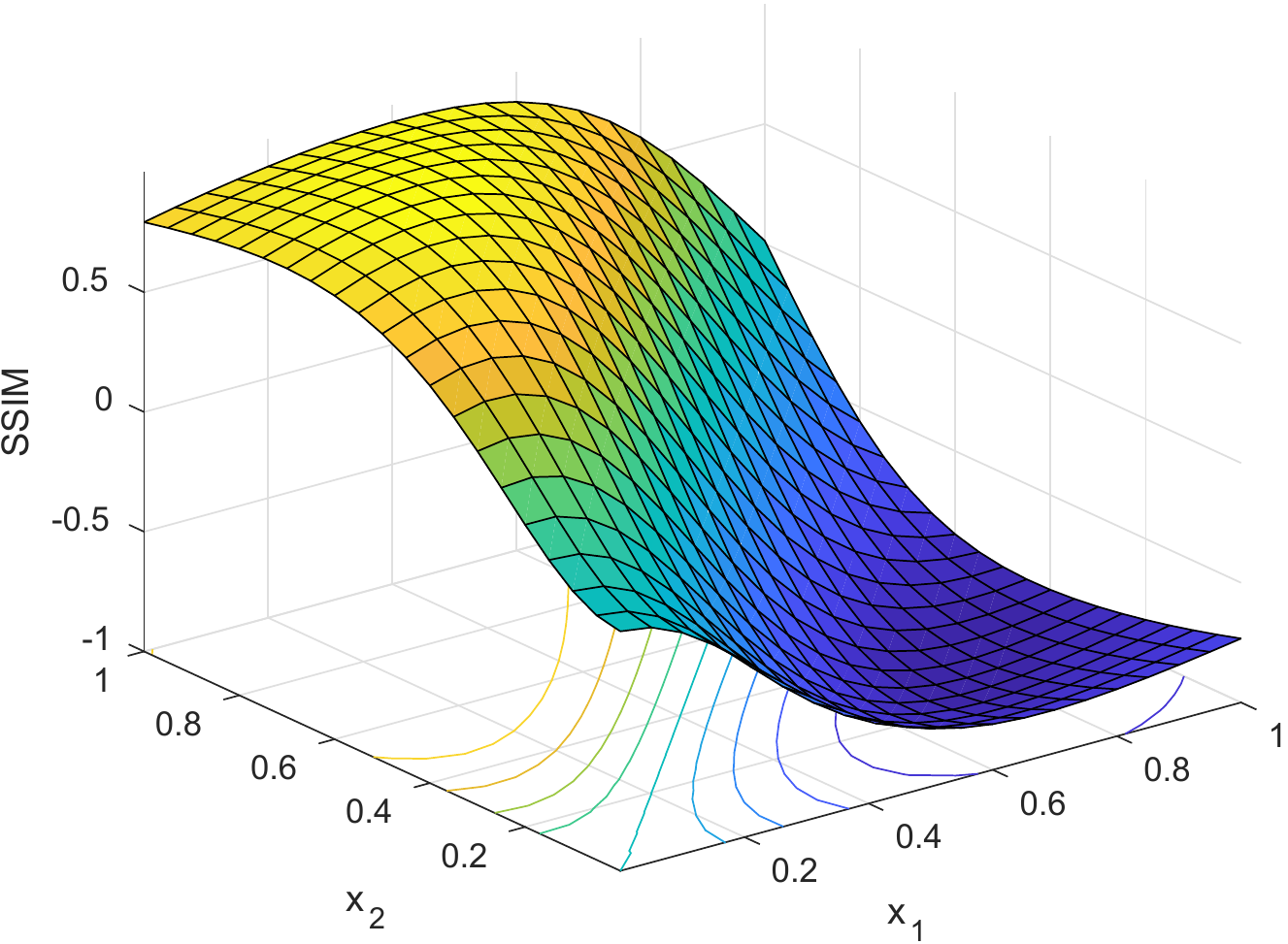}
					\caption{$B = (0.25,0.75)$}
					\label{fig:ssim-0.25-0.75}
				\end{subfigure}
				\begin{subfigure}{0.49\linewidth}
					\centering
					\includegraphics[width=\linewidth]{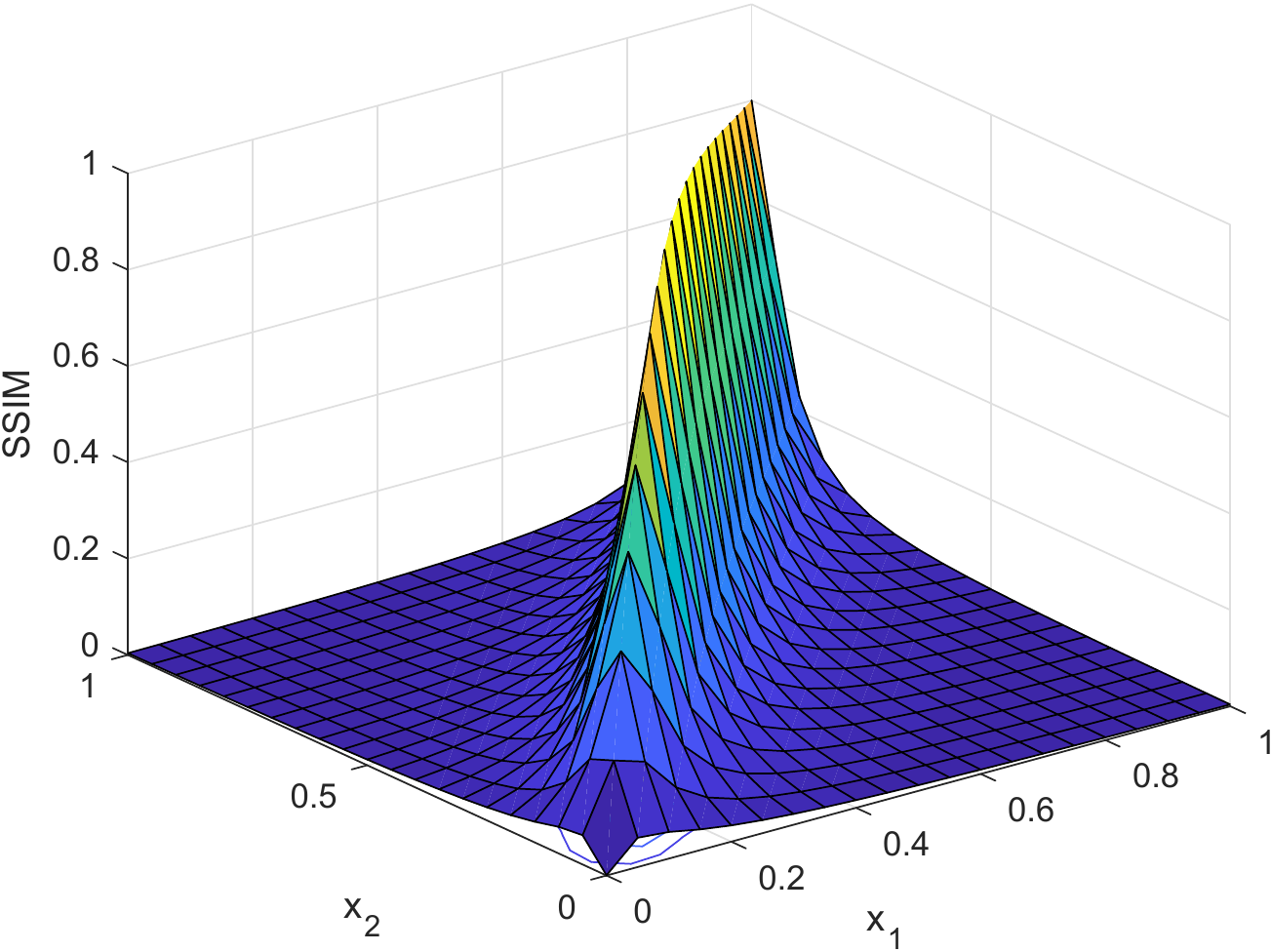}
					\caption{$B = (0.5,0.5)$}
					\label{fig:ssim-0.5-0.5}
				\end{subfigure}
				\begin{subfigure}{0.48\linewidth}
					\centering
					\includegraphics[width=\linewidth]{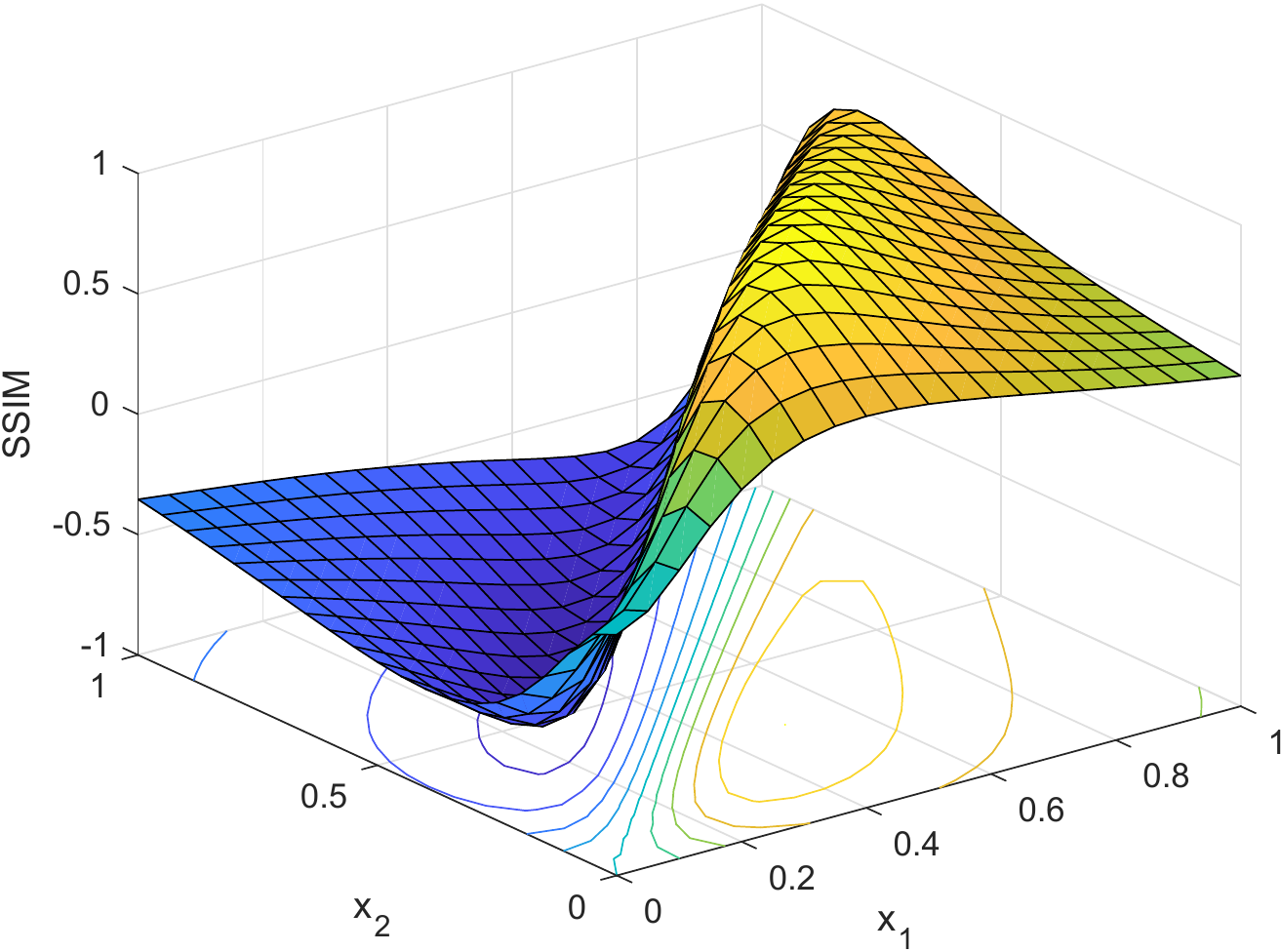}
					\caption{$B = (0.5,0.3)$}
					\label{fig:ssim-0.5-0.3}
				\end{subfigure}
				\caption{Visualization of the 2D landscape for SSIM with various target solutions.}
				\label{fig:SSIM}
			\end{figure}

			With regards to the SSIM measure, larger values indicate a higher degree of similarity. The 2D landscape for the SSIM metric is provided in Figure \ref{fig:SSIM}, which shows that the landscape characteristics vary with the target image. The SSIM metric is non-separable and the characteristics are expected to change with an increase in problem dimensionality.
			
	\section{Mapping Schemes for Image-Based Benchmarking}
		\label{sec:case-studies}
		
		In this section, a variety of mapping schemes which demonstrate the efficacy and flexibility of the proposed image-based visualization framework, in the context of image replication using the mapping function described in Section \ref{sec:image-mapping}, are presented.
		
		After each iteration in the optimization process, the mapping function given in Equation (\ref{eq:mapping}) was applied to the best solution in the population, thereby providing a visualization of the best fitness over time. This image is then displayed to the user, thereby giving an indication of the search characteristics. For example, an improvement in the overall image quality between two subsequent iterations indicated that the best fitness had improved while an image that does not change for a number of iterations would indicate that the global best has stagnated. Unless otherwise noted, the results presented are taken from a single execution of each algorithm. It should be explicitly noted that the purpose of these experiments was not to perform a comparative empirical study of the examined algorithms. Rather, the purpose of these experiments is to elucidate the efficacy and flexibility of the proposed image-based visualization framework. Therefore, the experimental results are meant to demonstrate that meaningful observations can be made regarding the respective optimization processes using image-based visualization.
		
		Unless otherwise stated, control parameter values were set as follows. For the PSO algorithm, a global-best (star) topology was employed with an asynchronous iteration strategy. Control parameter values were set as $\omega = 0.729844$, $c_1=c_2 = 1.49618$. Velocity clamping \cite{Kennedy1995Particle} was employed with $v_{max} = 0.1 * (x_{max} - x_{min})$ and $v_{min} = -v_{max}$. Velocities are initialized to $\vec{0}$ and, to prevent invalid attractors, a particle’s personal best position was only updated if a new position had a better objective function value and was within the feasible bounds of the search space. For the DE and GDE3 algorithms, the DE/rand/1/bin strategy was employed with $F = 0.5$ and $CR = 0.9$. To prevent infeasible solutions, a clamping boundary strategy was employed. All algorithms had a population size of 100 and were executed for $10,000D$ function evaluations, where $D$ is the problem dimensionality. Unless otherwise stated, all definitions and experiments assume that the objective was minimization.
		
		\begin{casestudy}[Continuous, unconstrained, single-objective]
			\label{cs:continuous}
			
			The most general form of optimization is unconstrained optimization, which can be defined as
			
			\begin{equation}
				\label{eq:unconstrained-optimization}
				\begin{array}{ll}
					\textit{Minimize:}   & f(\vec{x})         \\
					\textit{Subject to:} & \forall i: x_{i} \in dom(x_i), i = 1, \ldots, n
				\end{array}
			\end{equation}
			where $n$ is the number of decision variables and $dom(x_i)$ is the domain of each decision variable $x_i$, assuming the objective is to minimize $f$. In this mapping scheme, continuous, unconstrained, single-objective optimization problems are considered by defining $dom(x_i)$ as the continuous interval $[0,1]$, such that the decision variables represent the normalized greyscale values of the corresponding pixels.
			
			Figure \ref{fig:cont-pso-de-lena30} shows a comparison of two algorithms, namely PSO and DE using SAE, given in Equation (\ref{eq:sae}), as the fitness. Immediately, the positive correlation between the objective fitness and the visual quality was observed, which is a critical aspect of the image-based visualization that differentiates it from previous approaches. Interestingly, at iteration 7990, both PSO and DE had the same fitness value, yet produced two distinct images, indicating that they did not correspond to the same location in the search space. Without the visualization, one would not be able to differentiate two distinct solutions with equal fitness values. As the optimization progresses, both optimizers depicted very similar fitness values and both were able to replicate the target image very accurately.
			
			One drawback that can be seen from the image-based visualizations in Figure \ref{fig:cont-pso-de-lena30} is that once the solution quality reaches a certain threshold, the resulting images become too similar to discern much valuable information regarding the respective optimization processes. In order to address this concern, an alternative visualization, which highlights the dissimilarities between the resulting image and the target image, was also produced and is shown to the right of the corresponding image. In these visualizations, referred to as error heatmaps, each pixel was colourized according to the difference between its corresponding decision variable value and the value of the target decision variable. Note that, in the error heatmaps, an optimal solution would be visualized as a completely white image while sub-optimal decision variables are colourized according to the magnitude of their respective error.  Thus, the overall solution quality can be inferred by the amount of white space in the image. However, the intended purpose of the error heatmap is to highlight the dissimilarities, which is somewhat difficult to discern in the original images once the fitness is sufficiently close to the optimal, and therefore they serve as a complementary visualization technique. 
			
			To ascertain the effect of increasing the problem dimensionality, Figure \ref{fig:cont-pso-lenaVarious} shows the performance of PSO on a 900D (i.e., 30x30), 1600D (i.e., 40x40), and 2500D (i.e, 50x50) problem. These experiments also highlight the ability of the visualization technique to trivially scale with problem dimensionality. Figure \ref{fig:cont-pso-lenaVarious} clearly depicts the increased difficulty as dimensionality increased. After only 10000 iterations, the visual disparity between the images produced on each of the problems is readily apparent. The increased difficulty is also clearly visible in the error heatmaps -- after 50000 iterations, there was nearly no errors visible on the error heatmap for the 900D problem, while the error heatmap for the 2500D problem depicted a significant amount of error.

			\begin{figure}[!t]
				\centering
				\includegraphics[width=\linewidth]{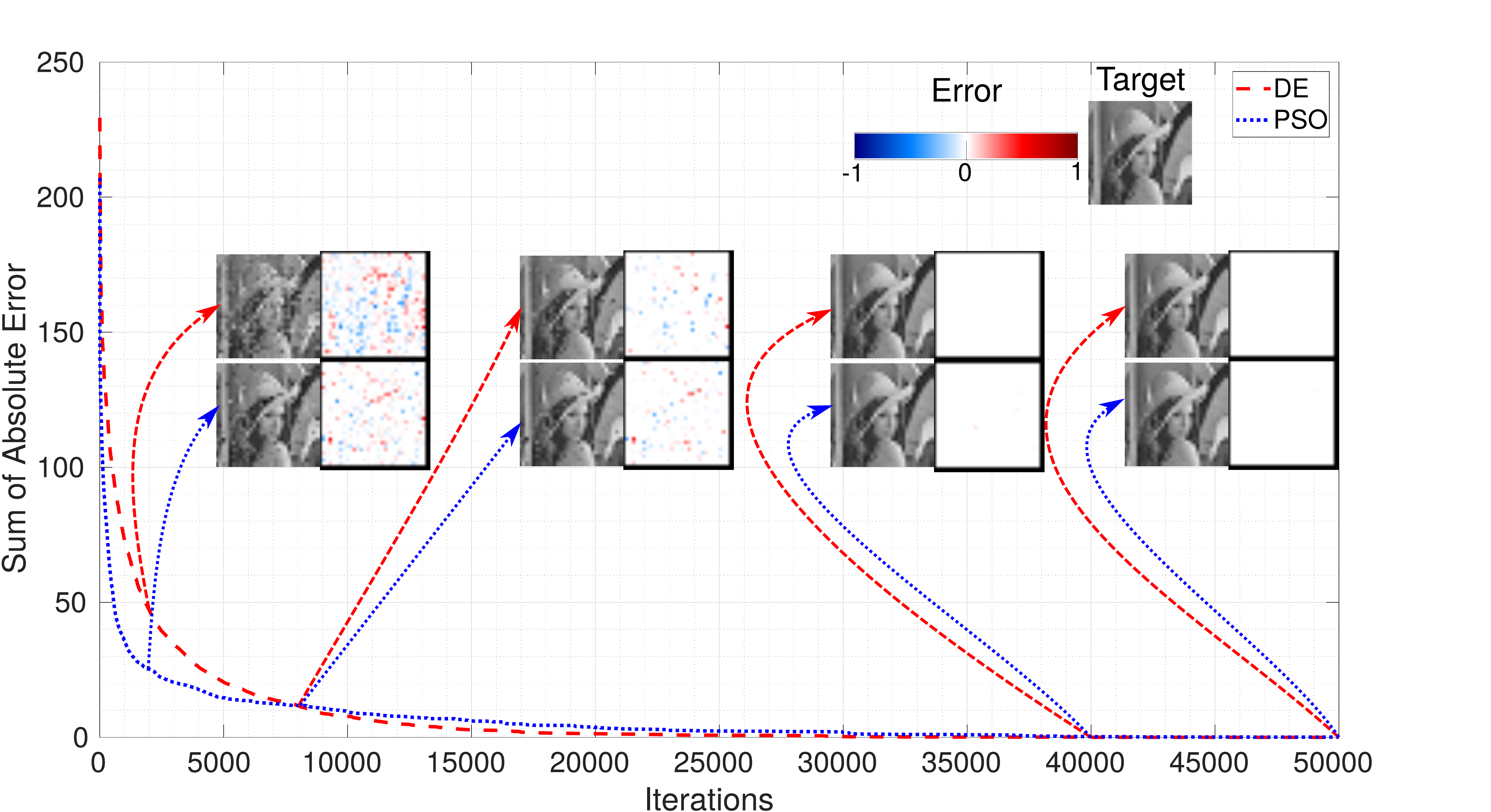}
				\caption{PSO and DE on 30x30 continuous Lena using SAE. Images scaled by 400\%.}
				\label{fig:cont-pso-de-lena30}
			\end{figure}
		
			\begin{figure}[!t]
				\centering
				\includegraphics[width=\linewidth]{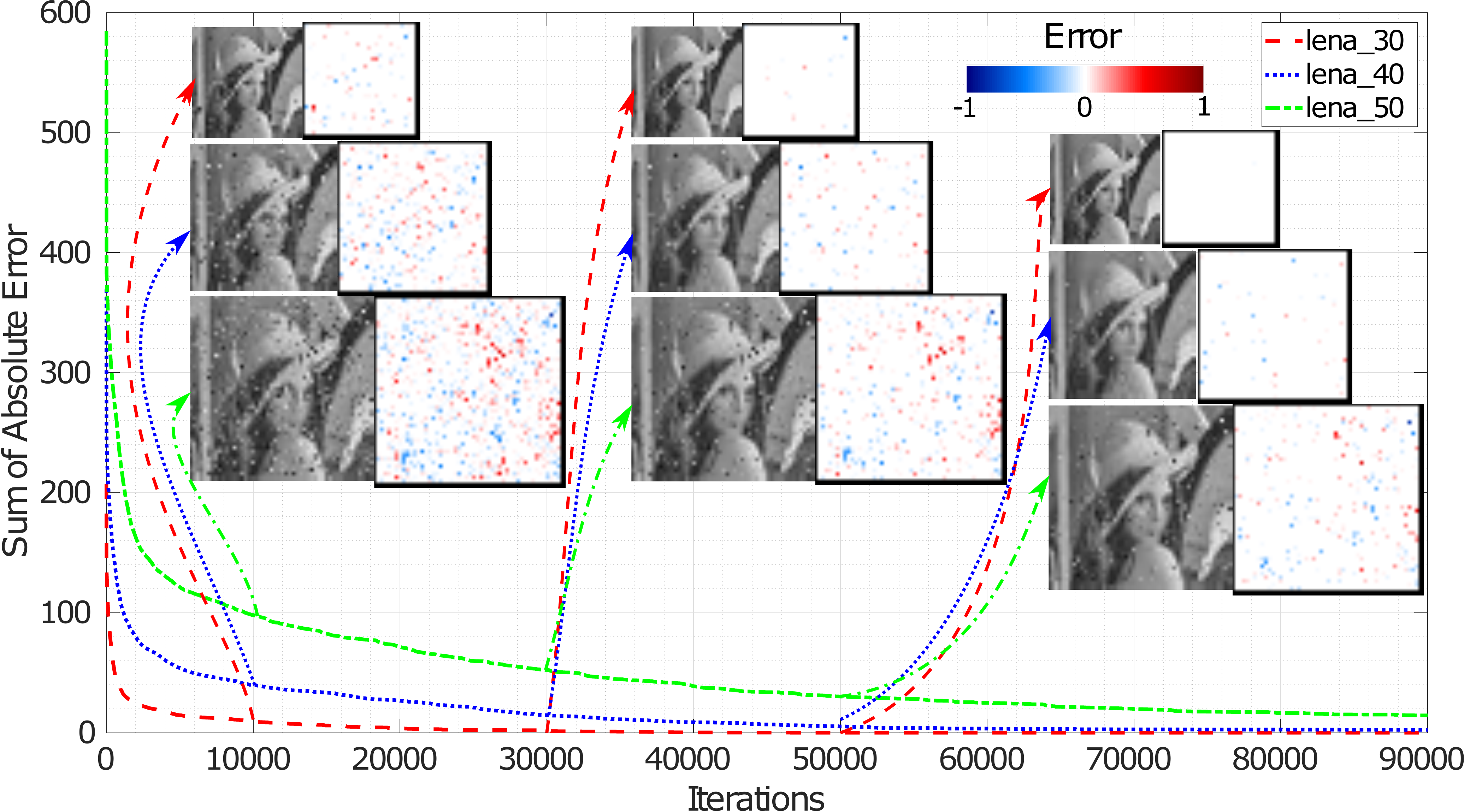}
				\caption{PSO on 30x30, 40x40, and 50x50 continuous Lena using SAE. Images scaled by 400\%.}
				\label{fig:cont-pso-lenaVarious}
			\end{figure}
		\end{casestudy}
		
		\begin{casestudy}[Alternative fitness functions for continuous, unconstrained, single-objective]
		
			In this scheme, alternative fitness functions for continuous, unconstrained, single-objective problems are investigated for the 30x30 Lena image using both the PSO and DE algorithm. The overall goal is to investigate various landscape characteristics and ascertain the visual similarity induced by alternative fitness functions. Note that, the results for the SAE measure from Mapping Scheme \ref{cs:continuous} are not repeated here.
			
			\subsection{Mean Squared Error}
				Figure \ref{fig:cont-pso-de-mse-lena30} presents the results when the MSE, as given in Equation (\ref{eq:mse}), was employed as the fitness function. As can be expected, the general observations made when MSE was used as a fitness function are largely the same as when SAE was used. Specifically, the MSE is an effective measure of similarity that appears to match well with human perception. However, the MSE produced images of similar quality quicker than those produced when SAE was employed; the image quality after 5000 iterations using MSE is roughly the same as the quality of images after 7990 iterations when SAE was used. This result is expected given that MSE more harshly penalizes large errors, thereby causing a more rapid initial improvement.
				
				\begin{figure}[!t]
					\centering
					\includegraphics[width=\linewidth]{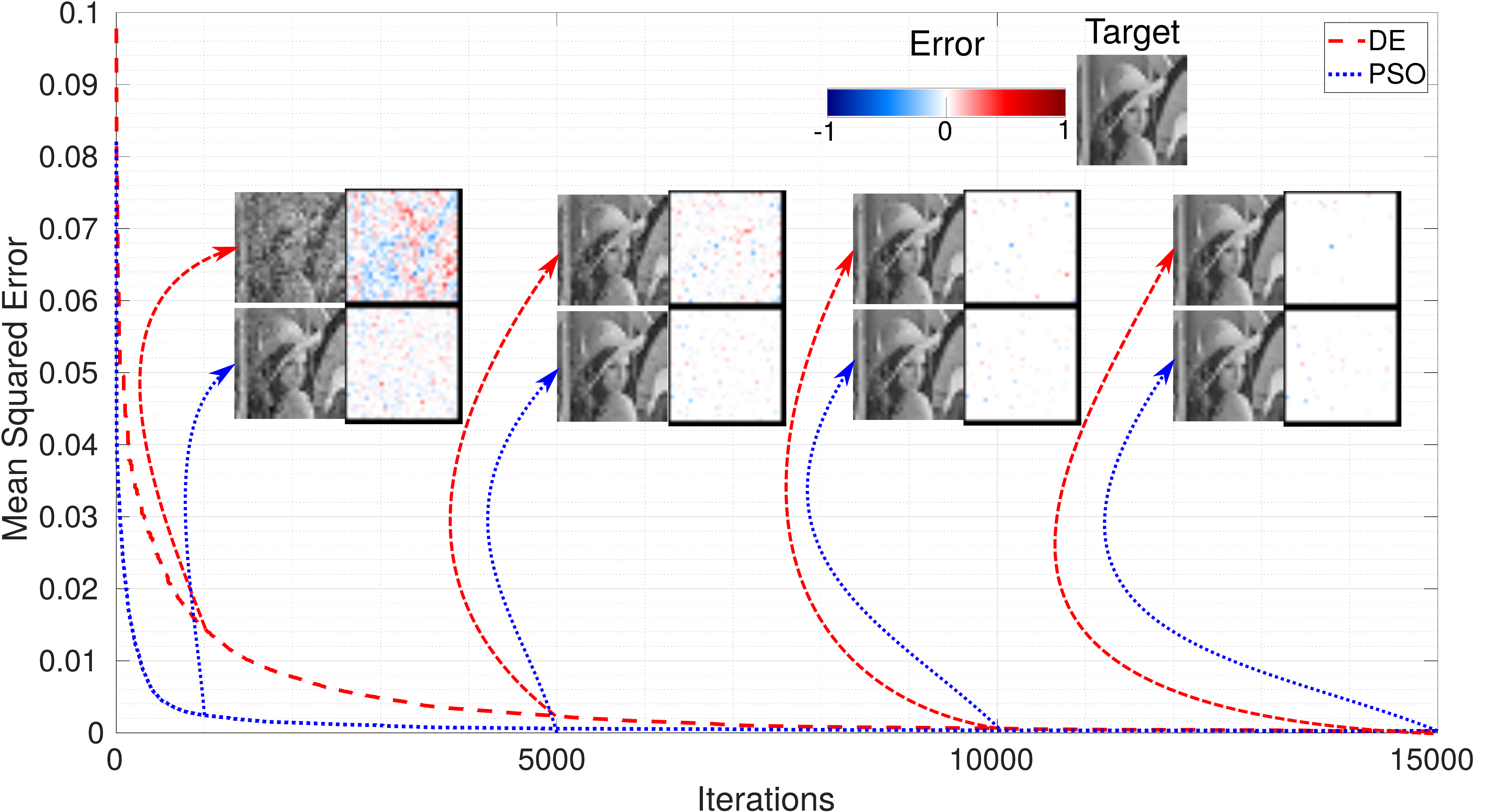}
					\caption{PSO and DE on 30x30 continuous Lena using MSE. Images scaled by 400\%.}
					\label{fig:cont-pso-de-mse-lena30}
				\end{figure}
			
			\subsection{2D correlation coefficient}
				Figure \ref{fig:cont-pso-de-pcc-lena30} presents the results when the 2D correlation coefficient, as given in Equation (\ref{eq:pcc}), was employed as the fitness function. To employ the PCC measure in a minimization context, objective values were negated. The images produced using PCC depict an inherent property of the PCC measure, namely that it measures correlation, rather than direct similarity. This correlation is most clearly depicted by the PSO results, whereby the resulting images are very highly correlated with the target image, which results in a near perfect fitness value but an image that is not identical to the target. Interestingly, this also leads to an error heatmap whereby the overall structure of the target image is visible, further highlighting the correlation between the candidate solution and the target image. In contrast, the DE algorithm produced a target image that had much greater visual similarity to the target, despite having a nearly identical fitness as the PSO algorithm.
			
				\begin{figure}[!t]
					\centering
					\includegraphics[width=\linewidth]{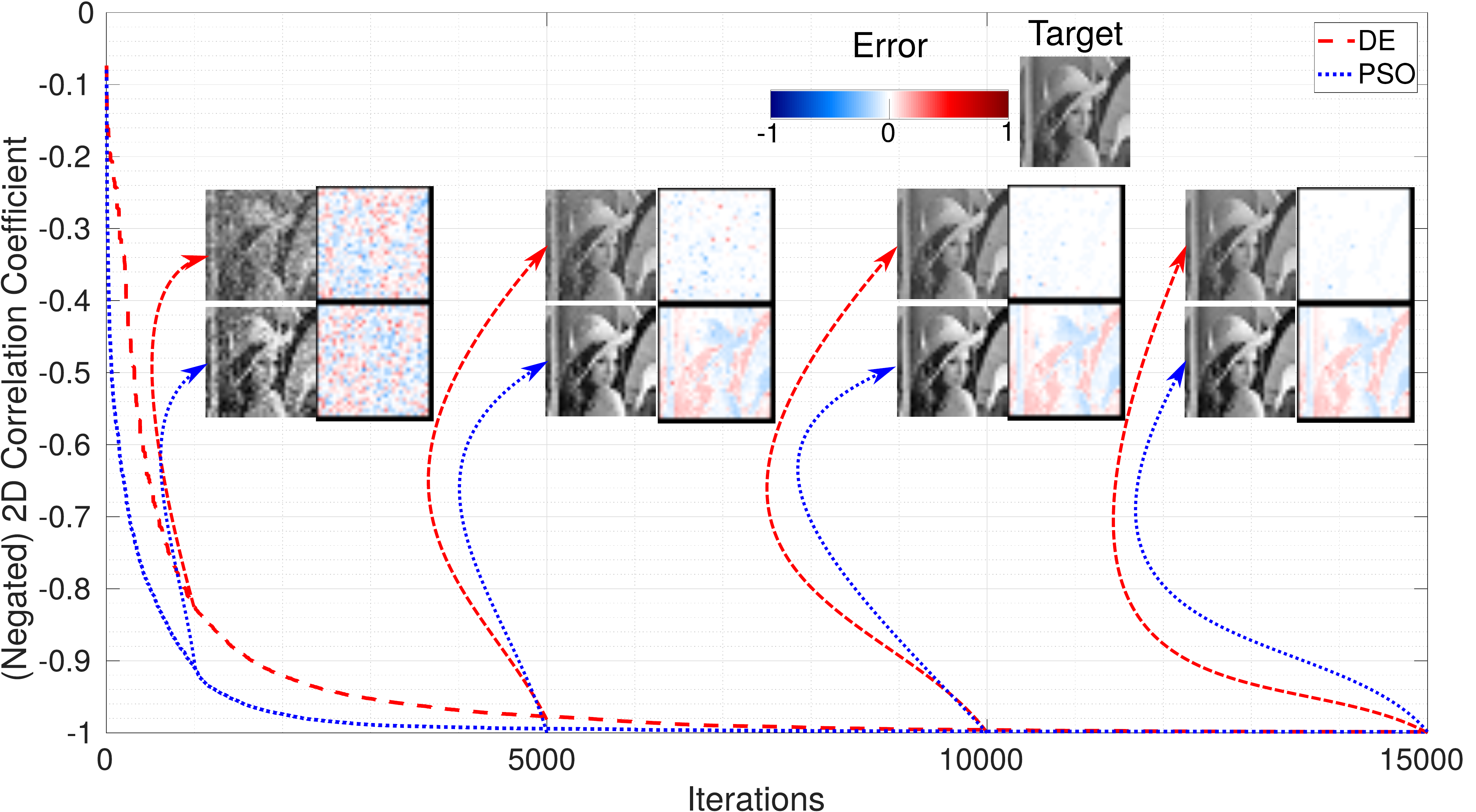}
					\caption{PSO and DE on 30x30 continuous Lena using PCC. Images scaled by 400\%.}
					\label{fig:cont-pso-de-pcc-lena30}
				\end{figure}
		
			\subsection{Peak Signal-to-Noise Ratio}
				Figure \ref{fig:cont-pso-de-psnr-lena30} presents the results when the PSNR measure, as given in Equation (\ref{eq:psnr}), was employed as the fitness function. To employ the PSNR measure in a minimization context, objective values were negated. Examining the results using PSNR, notably different fitness profiles were observed, attributed in part to the logarithmic nature of the PSNR measure. Specifically, the fitness plots depict numerous relatively flat regions, followed by sharp decreases in fitness. Nonetheless, the PSNR measure depicted enhanced visual quality as the fitness improved, thereby indicating it was a suitable metric for image-based visualization.
			
				\begin{figure}[!t]
					\centering
					\includegraphics[width=\linewidth]{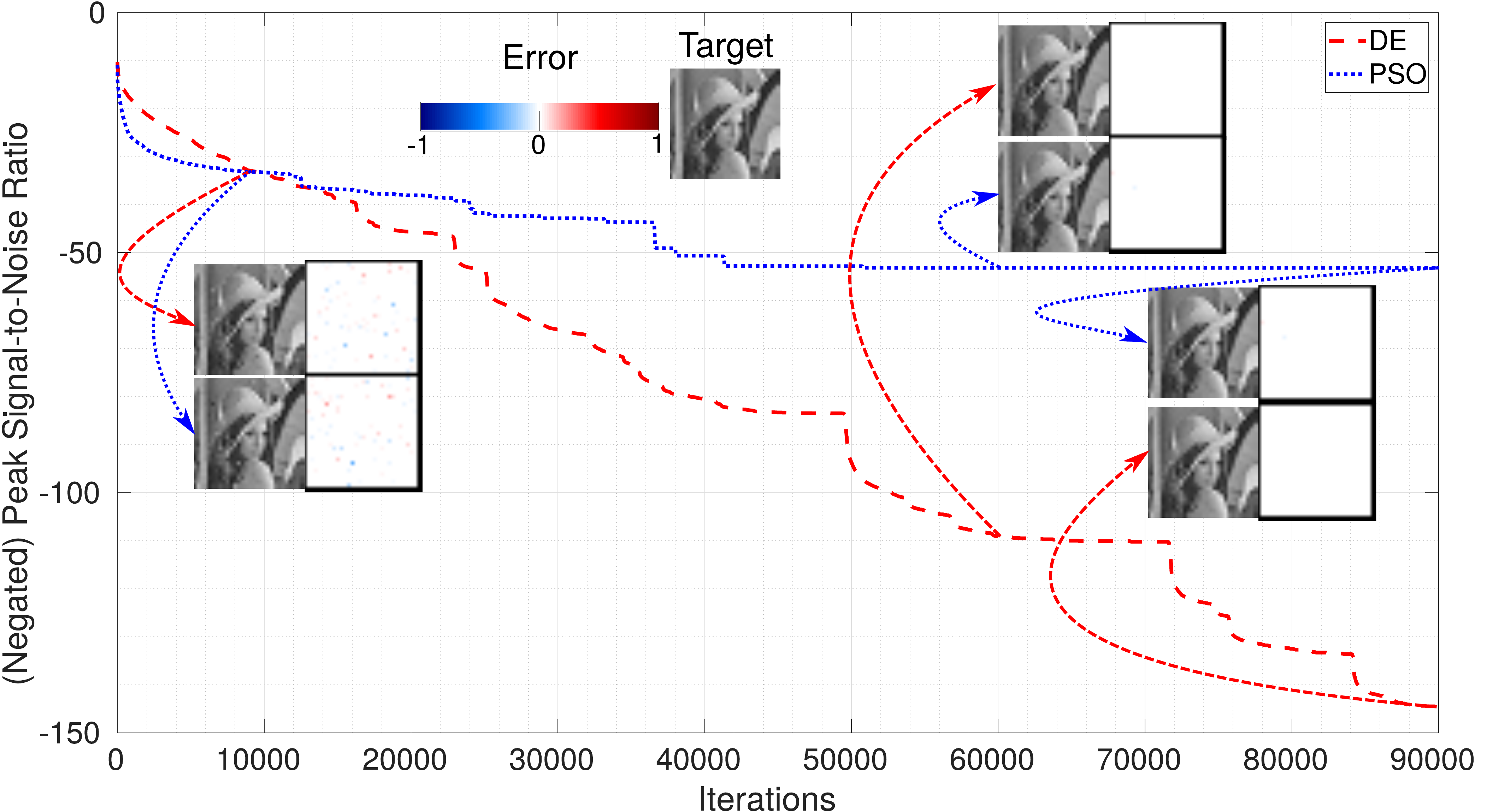}
					\caption{PSO and DE on 30x30 continuous Lena using PSNR. Images scaled by 400\%.}
					\label{fig:cont-pso-de-psnr-lena30}
				\end{figure}
		
			\subsection{Structural Similarity Index}
				The final measure employed in this study was the SSIM, as given by Equation (\ref{eq:ssim-all}). Again, the values of SSIM were negated to be employed in a minimization context. Note that, using SSIM as a fitness measure induces non-separability because Equation (\ref{eq:ssim-all}) makes use of properties dependent upon the entire image, such as averages and standard deviations, in contrast to using the fully-separable SAE measure, which is dependent only upon individual pixel values. Therefore, SSIM is expected to be a more challenging metric to optimize. As can be seen in Figure \ref{fig:cont-pso-de-ssim-lena30}, neither optimizer produced an image that was visually close to the target. An interesting observation was that the non-separability can be readily observed in the error heatmaps for both optimizers. This correlation is observed via the vertical segments in the error heatmaps, which alternate in colour across the horizontal axis. In contrast, the other, fully-separable measures, depicted scattered (i.e., independent) pixel-wise errors. Therefore, the difficulty introduced by the non-separability of SSIM was readily apparent in the results. Furthermore, the image-based visualization provided further information about the underlying fitness function.
			
				\begin{figure}[!t]
					\centering
					\includegraphics[width=\linewidth]{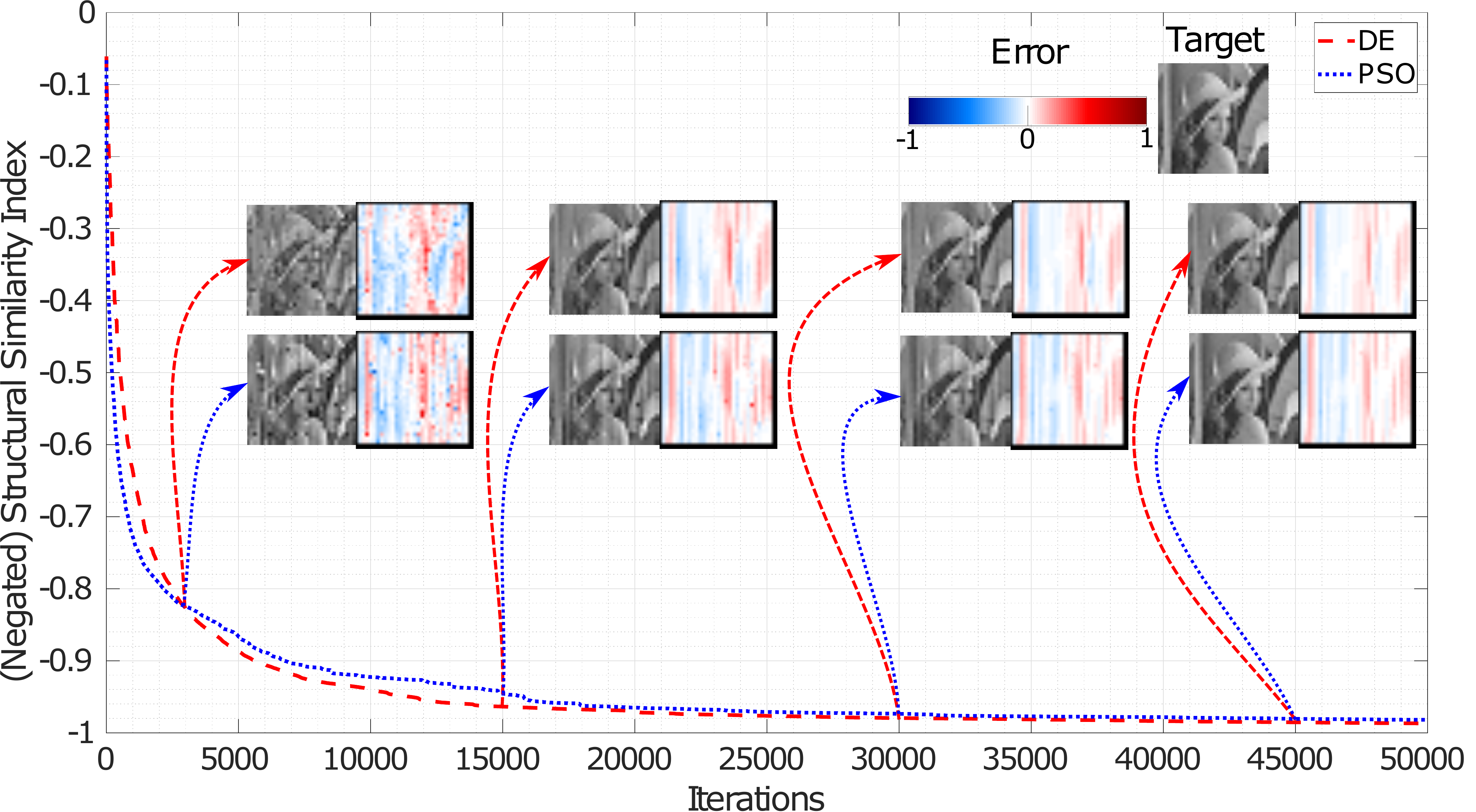}
					\caption{PSO and DE on 30x30 continuous Lena using SSIM. Images scaled by 400\%.}
					\label{fig:cont-pso-de-ssim-lena30}
				\end{figure}
	
			\subsection{Summary}
				To provide an overall summary of the results using various fitness functions, Figures \ref{fig:pso-fit-compare} and \ref{fig:de-fit-compare} present the resulting images at various iterations for PSO and DE, respectively. After 90000 iterations, the images produced when SAE, MSE, and PSNR were used as fitness functions were nearly identical to the target for both algorithms. In contrast, the worst visual quality was attained when SSIM was employed as the fitness function. The images produced when PCC was used as the fitness function were structurally similar to the target, yet the colour was primarily correlated, rather than identical. A further interesting observation is that the images produced by PSO after 1000 iterations had much higher visual quality than those produced by DE. This implies that the PSO demonstrated superior performance in the early stages of the search. However, after 10000 iterations, the images produced by DE were generally of a higher quality than those produced by PSO.
			
				\begin{figure}[!ht]
					\centering
					\includegraphics[width=\linewidth]{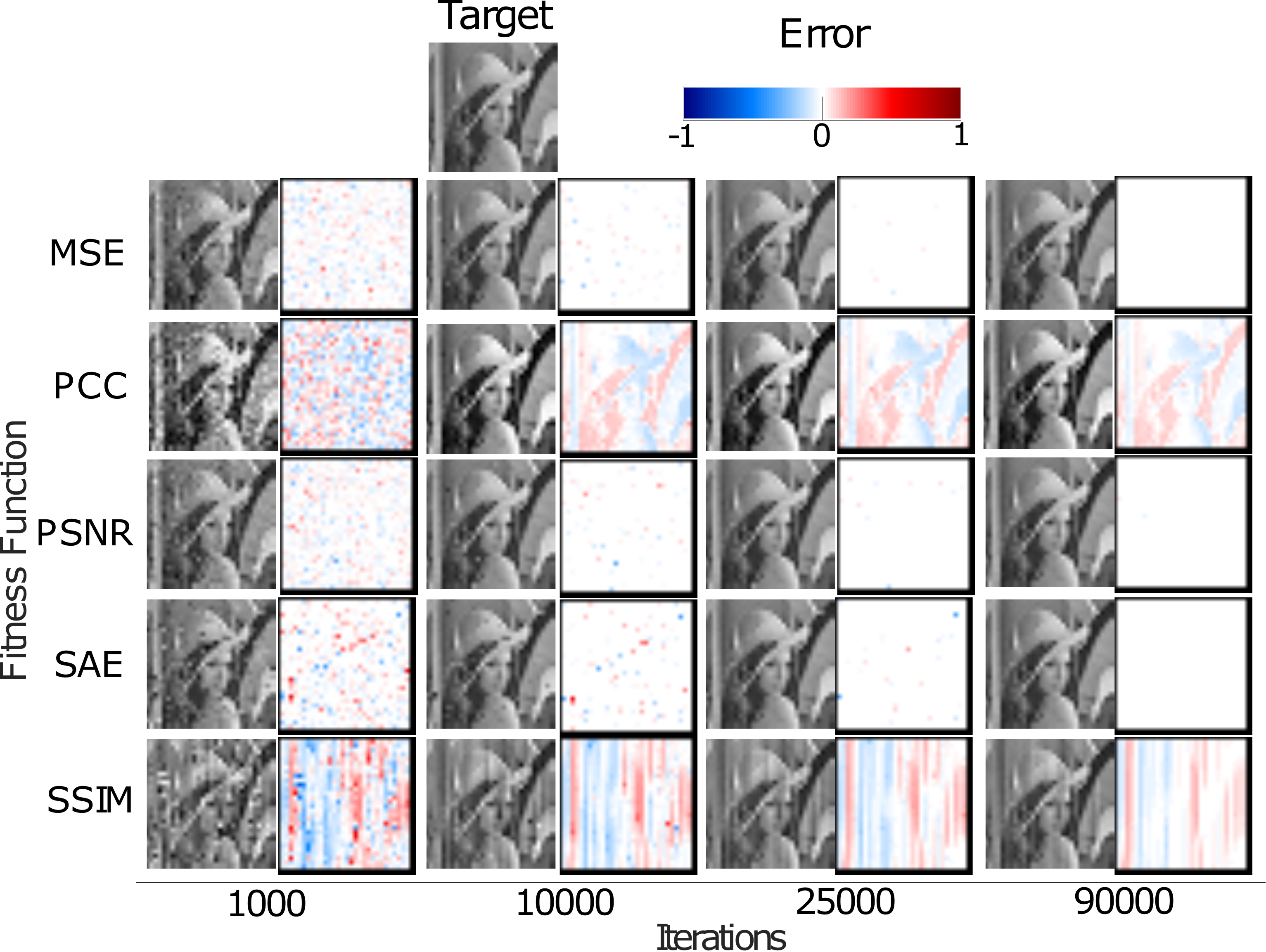}
					\caption{Comparison of various fitness functions using PSO on 30x30 continuous Lena. Images scaled by 400\%.}
					\label{fig:pso-fit-compare}
				\end{figure}
			
				\begin{figure}[!ht]
					\centering
					\includegraphics[width=\linewidth]{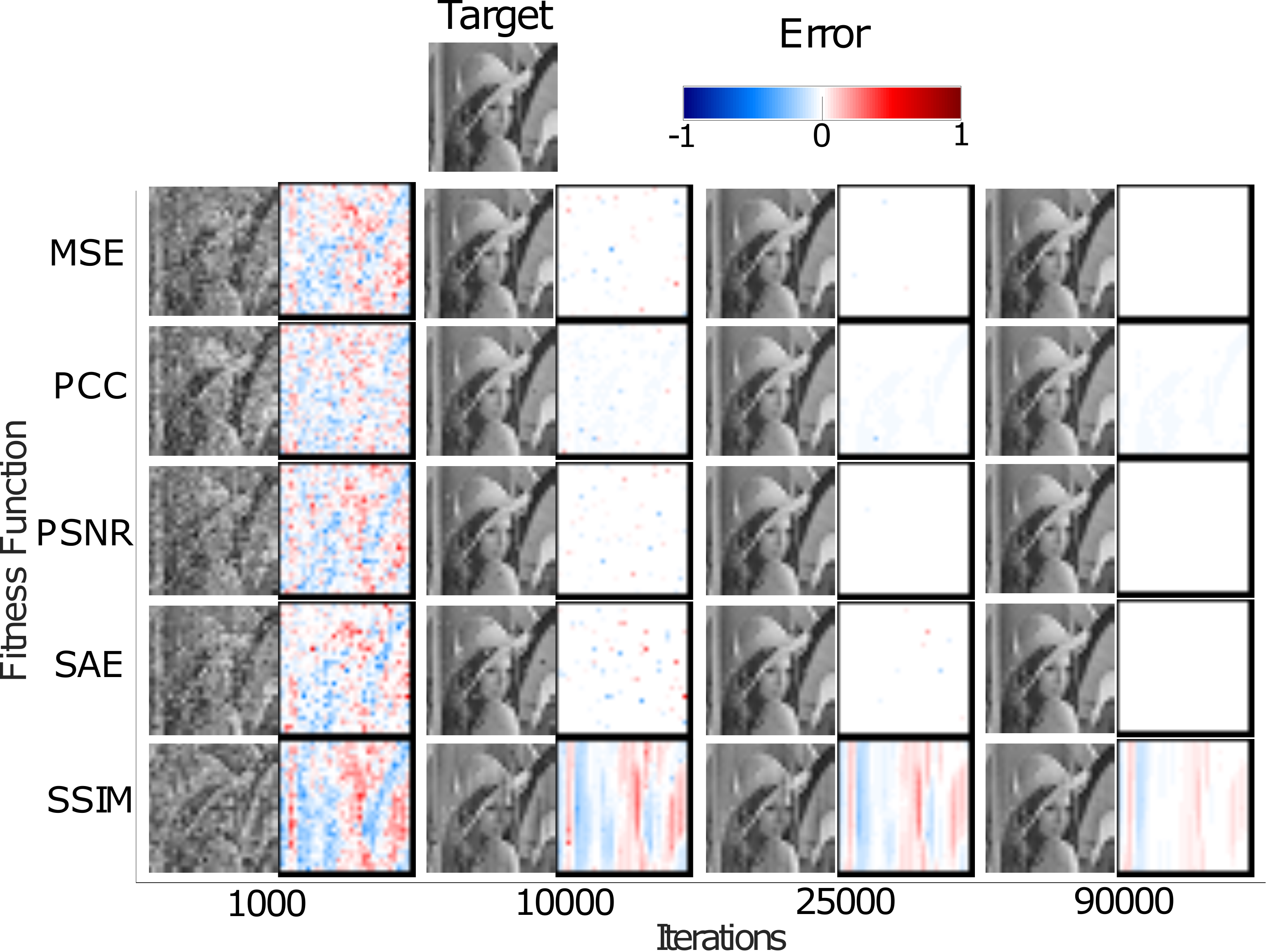}
					\caption{Comparison of various fitness functions using DE on 30x30 continuous Lena. Images scaled by 400\%.}
					\label{fig:de-fit-compare}
				\end{figure}
		\end{casestudy}
		
		\begin{casestudy}[Discrete, unconstrained, single-objective]
			
			In the image-based visualization framework, a discrete optimization problem can be easily defined by considering a target image with a discrete intensity value for each pixel, i.e., by defining $dom(x_i) = [0 .. 255]$ in Equation (\ref{eq:unconstrained-optimization}). As such, the pixel values are taken as (discrete) integers in range [0..255], which correspond to the 8-bit greyscale intensity. Given the larger domain, the corresponding fitness values are expected to be higher than the previous results using a continuous domain. Similar to the other case studies, both the DE and PSO algorithms were applied on the 30x30 Lena image. 
			
			Given that both algorithms are inherently continuous meta-heuristics, they must be adapted to handle discrete problems. A simple method to utilize continuous algorithms for discrete problems is to convert the variable values to discrete values before evaluating their objective fitness. In this case, the variable values (i.e., the pixel grey levels) are rounded to closest integer number in the range [0..255] immediately before the objective function evaluation. The remaining elements of the algorithms have not been modified; the update phase, operators, and the selection schemes are unaltered. 
			
			Figure \ref{fig:discrete-de-pso-lena30-SAE} presents the results on a discrete problem. The image resulting from employing DE does not depict good visual quality after 9,000 iterations, while PSO was able to attain an image with much better quality. Contrasting the quality of images produced by the continuous and discrete problems indicates that the image is improved at a much slower rate for discrete problems, which provides direct evidence of their increased difficulty. Continuous optimization problems tend to be easier to solve than discrete optimization problems; the smoothness of the landscape facilitates deduction of information about solutions in the neighbourhood. Furthermore, the many-to-one mapping imposed by the discretization process causes a reduction in exploration power for the optimizers. These observations are directly indicated through the image-based visualization framework. 
			
			\begin{figure}[!t]
				\centering
				\includegraphics[width=\linewidth]{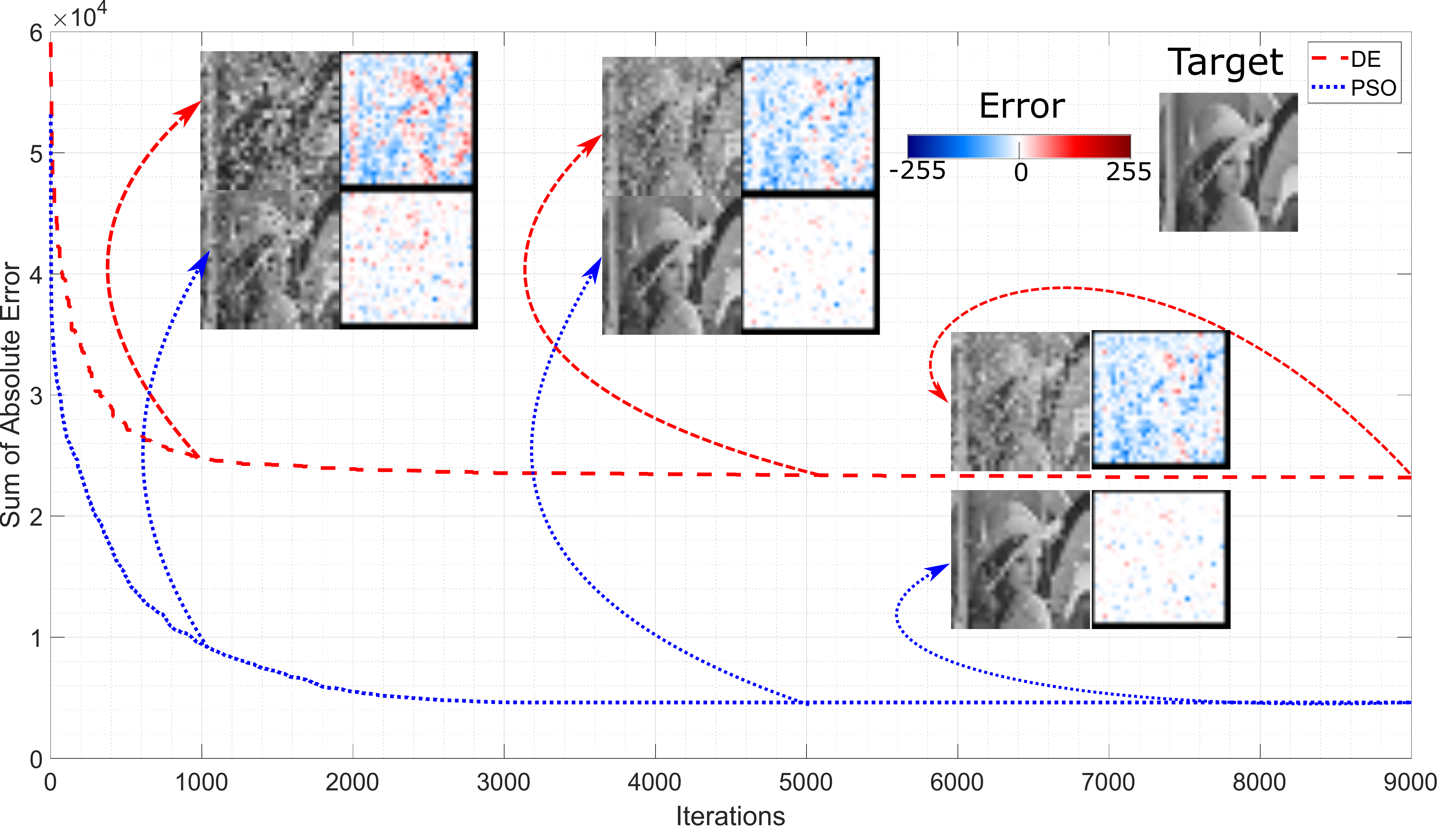}
				\caption{DE and PSO on discrete 30x30 Lena using SAE. Images scaled by 500\%.}
				\label{fig:discrete-de-pso-lena30-SAE}
			\end{figure}
		\end{casestudy}
		
		\begin{casestudy}[Binary, unconstrained, single-objective]
		
			To employ the image-based framework on a binary problem (i.e., a problem where the decision variables can take only the values \{0,1\}), a black and white image was considered. To formulate a binary problem, define $dom(x_i) = \{0, 1\}$ in Equation (\ref{eq:unconstrained-optimization}), such that the only possible values for the decision variables are the binary values 0 and 1, which represent black and white, respectively. 
			
			There are various optimization methods to handle binary problems. A binary optimization problem is a special case of discrete problems. Therefore, a simple approach to facilitate the use of existing continuous algorithms for binary problems is to convert the continuous decision vector to a binary representation using thresholding. Specifically, given a continuous decision vector with elements in the range [0,1], the values can be converted to binary by considering a threshold value of 0.5 such that decision variables with values less than 0.5 are converted to the binary value 0 (i.e., a black pixel) while decision variables values greater than 0.5 are converted to the binary value 1 (i.e., a white pixel) immediately before the objective function evaluation. This scheme was used for both the DE and PSO algorithms. The remaining elements of the algorithms have not been modified.
			
			Figure \ref{fig:bin-de-pso-lena30-SAE} presents the results on a binary problem. These results depict that the PSO algorithm only moderately decreased the error during the initial iterations, but after nearly 1000 iterations, the error did not significantly decrease. This observation can be readily made via the produced images; the quality of images did not improve, while the DE exhibited better results in terms of both the error value and the resulting images. Although the DE algorithm demonstrates only slight improvement after iteration 1,000, it was still able to produce an image with better visual quality when compared to the image produced by PSO, further reinforcing the positive correlation between visual quality and objective fitness.
			
			\begin{figure}[!t]
				\centering
				\includegraphics[width=\linewidth]{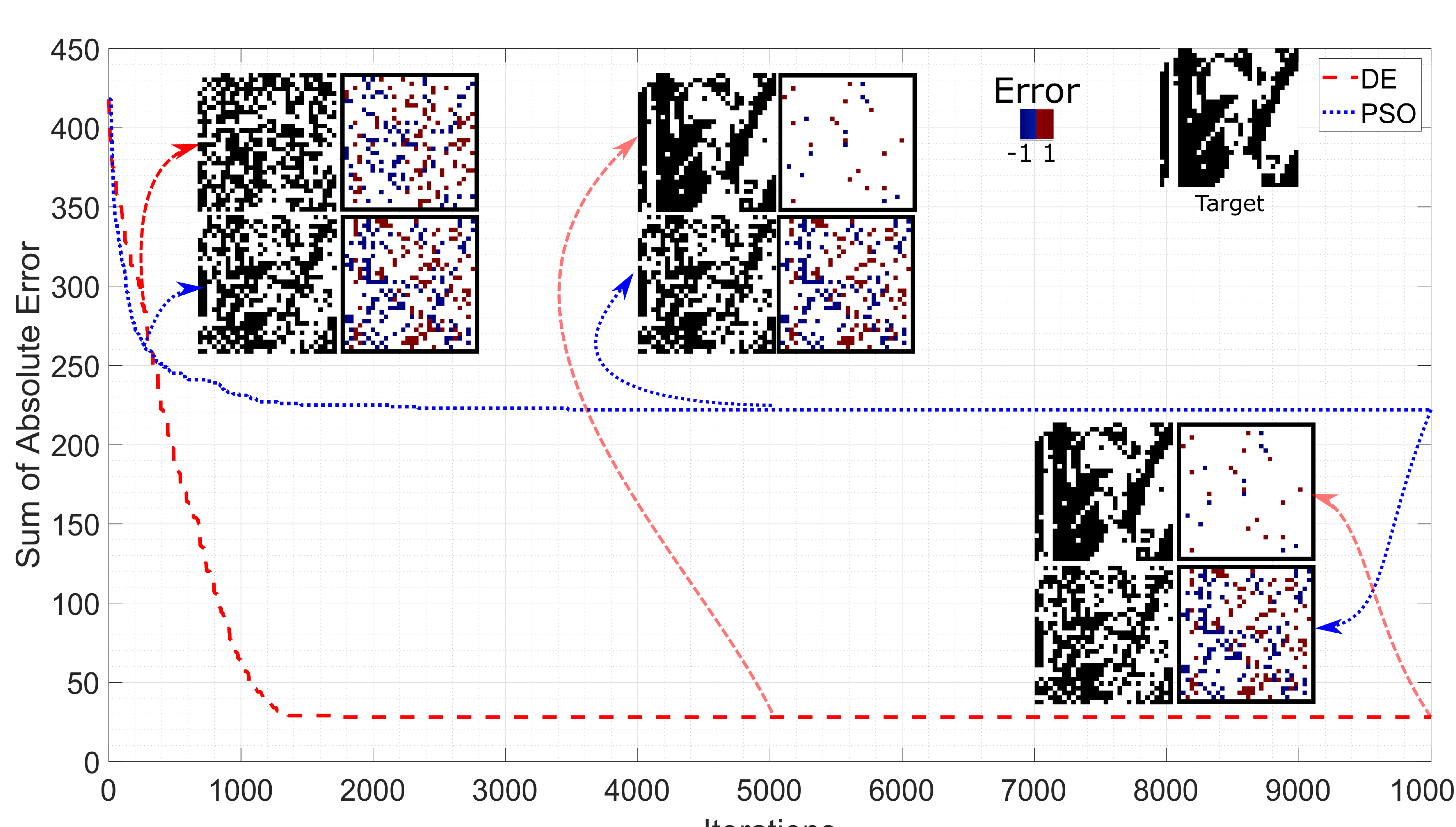}
				\caption{DE and PSO on binary 30x30 Lena using SAE. Images scaled by 500\%.}
				\label{fig:bin-de-pso-lena30-SAE}
			\end{figure}
		
		\end{casestudy}
		
		\begin{casestudy}[Combinatorial, single-objective]
			
			A combinatorial optimization problem can be defined as
			\begin{equation}
				\label{eq:combinatorial-optimization}
				\begin{array}{ll}
					\textit{Minimize:}   & f(\vec{x})                              \\
					\textit{Subject to:} &  \forall i : x_i \in C, \enskip i = 1, \ldots, n
				\end{array}
			\end{equation}
			where $C$ is a finite collection of elements. Note that, $C$ is a \textit{collection} and differs from a discrete \textit{set} of values as $C$ may contain duplicates. Moreover, combinatorial optimization problems are a special case of discrete optimization problems where the candidate solution $\vec{x}$ is composed of $n$ distinct elements from a collection of elements $C$, rather than $n$ values taken from a finite set $A$. That is, combinatorial problems are more restrictive than discrete problems as they impose additional constraints on candidate solutions. In the context of this study, $\vec{x}$ is an ordered permutation of the pixel intensity values. 
			
			Finding the best permutation in a high-dimensional space is a sophisticated problem given that it is not possible to evaluate all feasible permutations for a large-scale optimization problem -- for a problem with dimension $n$, the search space is $n!$. Therefore, an exhaustive search is infeasible and combinatorial optimization problems are generally considered to be harder problem than continuous or discrete optimization.

			In order to formulate a combinatorial problem in the context of this study, an image with discrete pixel values in range [0..255] was considered. The problem can be defined as finding the best permutation of the pixels to most accurately replicate the target image. The optimization was performed using combinatorial DE and PSO algorithms.  
			
			One combinatorial variant of DE employs a permutation matrix to generate permutations~\cite{price2006differential}. A permutation matrix $\pmb P$ is a matrix that maps a permutation vector to another permutation vector. The permutation matrix maps an integer permutation vector, $c_i$, to another integer permutation vector, $b_i$, by the relation $c_i=\pmb Pb_i$. In the DE method, this matrix is taken as the difference between $c_i$ and $b_i$. The analogous equation for the differential mutation in Equation~(\ref{eq:de:trial}) is then given by $c_i=\pmb P_Fa_i$, where $\pmb P_F$ is a modified permutation matrix, scaled by the parameter $F$, here with the meaning of a probability, in order to perform a fraction of the permutation represented by the original permutation matrix $\pmb P$. 
			
			In order to adapt the PSO algorithm for a combinatorial problem, Relative Position Indexing (RPI)~\cite{lichtblau2009relative} was employed. This approach maps an integer permutation vector to the floating-point interval [0,1] by dividing each element of the vector by the largest element, then applying the PSO over the transformed values in the continuous domain. To map the continuous vector back to the original domain, the smallest floating point value is mapped to the smallest integer value, then the next smallest floating point value is mapped to the next smallest integer value, and so on until all elements have been converted. For example, the permutation $\vec{x}(t) = [150, 10, 250, 40, 190]$ would be mapped to $\vec{x}'(t) = [0.6, 0.04, 1, 0.16, 0.76]$ and $\vec{x}'(t)$ would then be used as the position of the particle. The updated (continuous) particle position, such as $\vec{x}'(t+1) = [0.32, 0.8, 0.1, 0.51, 0.02]$ for instance, would then be mapped back to $\vec{x}(t+1) = [150, 250, 40, 190, 10]$.
			
			Figure \ref{fig:comb-ga-lena30-SAE} presents the results on a 30x30 combinatorial problem, which corresponds to a 900 dimensional problem, using the aforementioned DE and PSO algorithms. Thus, the algorithms were tasked with finding the best ordering of pixels among $900!$ ($\approx$ \num{6.75e2269}) distinct permutations, which is a very complex optimization task. As shown in Figure \ref{fig:comb-ga-lena30-SAE}, the inherent difficulty associated with combinatorial formulation is readily observed through the quality of images produced. Even after 50,000 iterations, the optimizers were unable produce high quality images. In the early phases of the search process, PSO resulted in lower error than DE, but the DE algorithm performed better over time. This is also evidenced by the error heatmap for DE depicting fewer errors in the later phases of the search.
			
			\begin{figure}[!t]
				\centering
				\includegraphics[width=\linewidth]{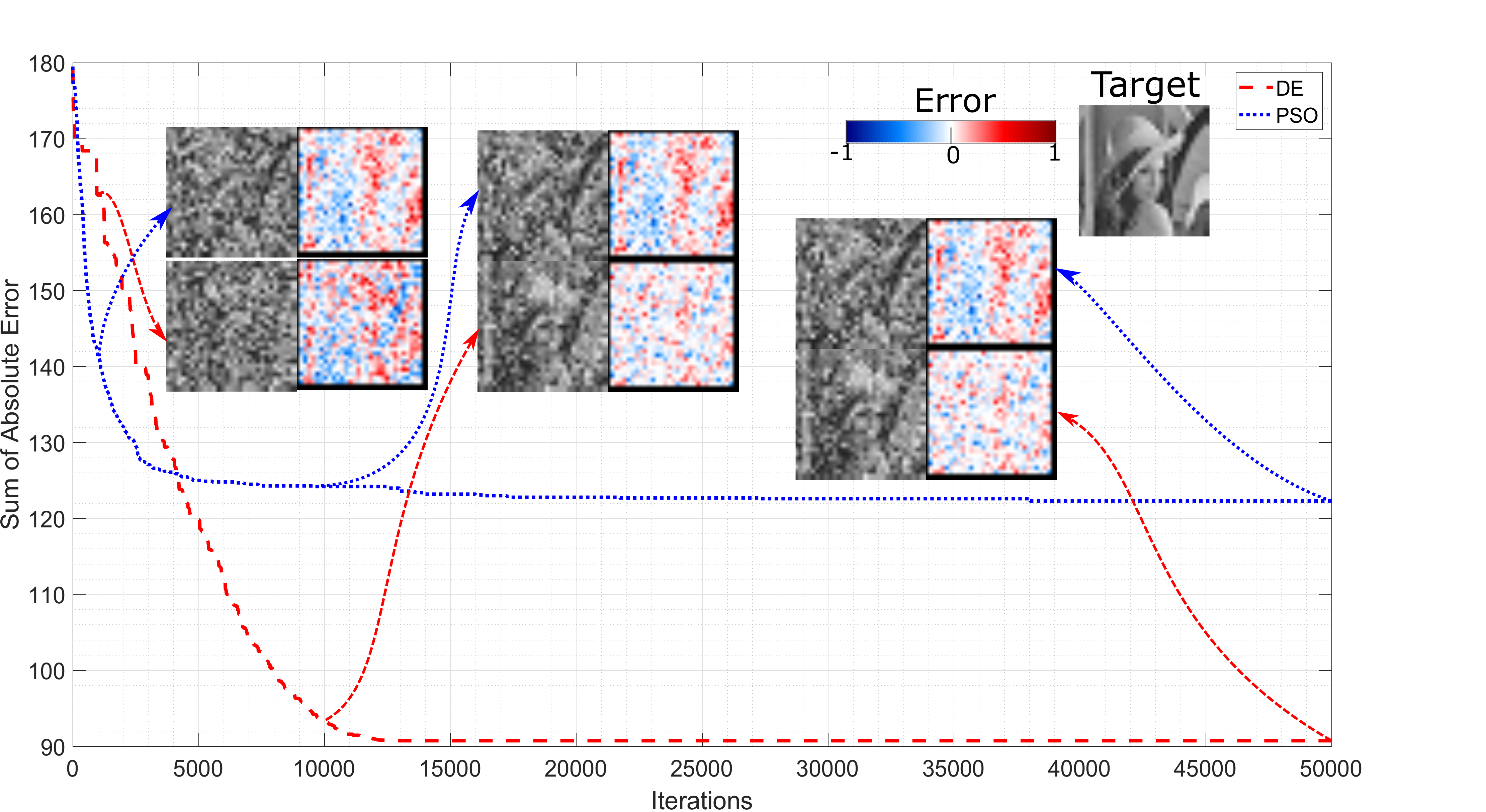}
				\caption{DE and PSO on combinatorial 30x30 Lena using SAE. Images scaled by 500\%.}
				\label{fig:comb-ga-lena30-SAE}
			\end{figure}
			
		\end{casestudy}
		
		\begin{casestudy}[Partially-separable, unconstrained, single-objective]
		
			Separability is often taken as measure of difficulty for an optimization problem. In general, separable problems are easier to solve than non-separable problems, because each variable of a separable problem is independent of the other variables~\cite{jamil2013literature}. A function of $p$ variables is referred to as separable if it can be written as a sum of $p$ independent functions of a single decision variable, such as
			\begin{equation}
				\underset{x_1,...,x_p}{\argmin} f(x_1,x_2,...x_p)=(\underset{x_1}{\argmin} f(x_1),...,\underset{x_p}{\argmin} f(x_p)).
			\end{equation}
			
			In terms of optimization problems, this implies that if each of the decision variables are independent, then the objective function can be decomposed into many sub-objective functions each of which involve only one decision variable. For example, since the SAE metric is defined as the sum of absolute error for each pixel, it is a fully separable function. Therefore, SAE can be optimized for each decision variable (i.e., pixel) independently, such that the independent results from minimizing the error on each pixel can be combined to produce an overall image whereby the SAE is also minimized. In order to produce a non-separable problem, a different fitness function must be employed. As discussed in Mapping Scheme 2, the SSIM is calculated using characteristics of the entire image, not independent pixels. Thus, the SSIM introduces a non-separable fitness function. To highlight the difficulty of non-separability in an optimization problem, both SAE and SSIM were evaluated on distinct regions (i.e., sub-images) of the target image. Therefore, a single image can be used to visually ascertain the effect of separability. Specifically, the target image was divided into two regions, where one region was evaluated using SAE and the other was evaluated using SSIM according to
			\begin{subequations}
				\begin{equation}
					\label{eq:partially-separable}
					f(\vec{x}, \vec{t}, p) = SAE(A_1, B_1) + \left|\frac{1}{SSIM(A_2,B_2)}\right|,
				\end{equation}
				\begin{equation}
					x\left[1:\lfloor pD \rfloor\right] \mapsto A_1
				\end{equation}
				\begin{equation}
					x[\lceil pD \rceil:D] \mapsto A_2
				\end{equation}
				\begin{equation}
					t\left[1:\lfloor pD \rfloor\right] \mapsto B_1
				\end{equation}
				\begin{equation}
					t[\lceil pD \rceil:D] \mapsto B_2
				\end{equation}
			\end{subequations}
			where $x$ is the candidate solution, $t$ is the (flattened) target image, and $p \in [0,1]$ is a control parameter that quantifies the level of separability. Using the parameter $p$, the level of separability can be explicitly controlled; setting $p = 1$ results in the SAE measure being used for the entire image (i.e., a fully separable problem), $p = 0$ results in the SSIM measure being used for the entire image (i.e., a non-separable problem), and $0 < p < 1$ results in exactly $(1 - p)\%$ of the image being evaluated using the fully-separable SAE metric. In this study, $p$ is set to 0.5, thus introducing a problem that is 50\% separable and 50\% non-separable.
			
			Figure \ref{fig:sep-de-pso-lena30-SAE-SSIM} presents the results on a partially-separable problem using DE and PSO algorithms, where the left half was evaluated using SAE as the fitness function and the right half of the image was evaluated using SSIM. The performance of two algorithms, DE and PSO were comparable and reached the same fitness value after approximately 4,000 iterations. Despite this, it was observed that the optimizers were able to produce images with reasonable visual quality on the half evaluated using SAE after only 10,000 iterations, while the SSIM half of the image had much lower quality. The quality of the images produced after 30,000 iterations was much better with regards to both metrics. Nonetheless, the region optimized with SAE still depicted much better quality. This result highlights that the increased difficulty associated with non-separable problems can be easily observed through the quality of images produced, thereby providing further evidence of the merits associated with image-based visualization.
			
			\begin{figure}[!t]
				\centering
				\includegraphics[width=\linewidth]{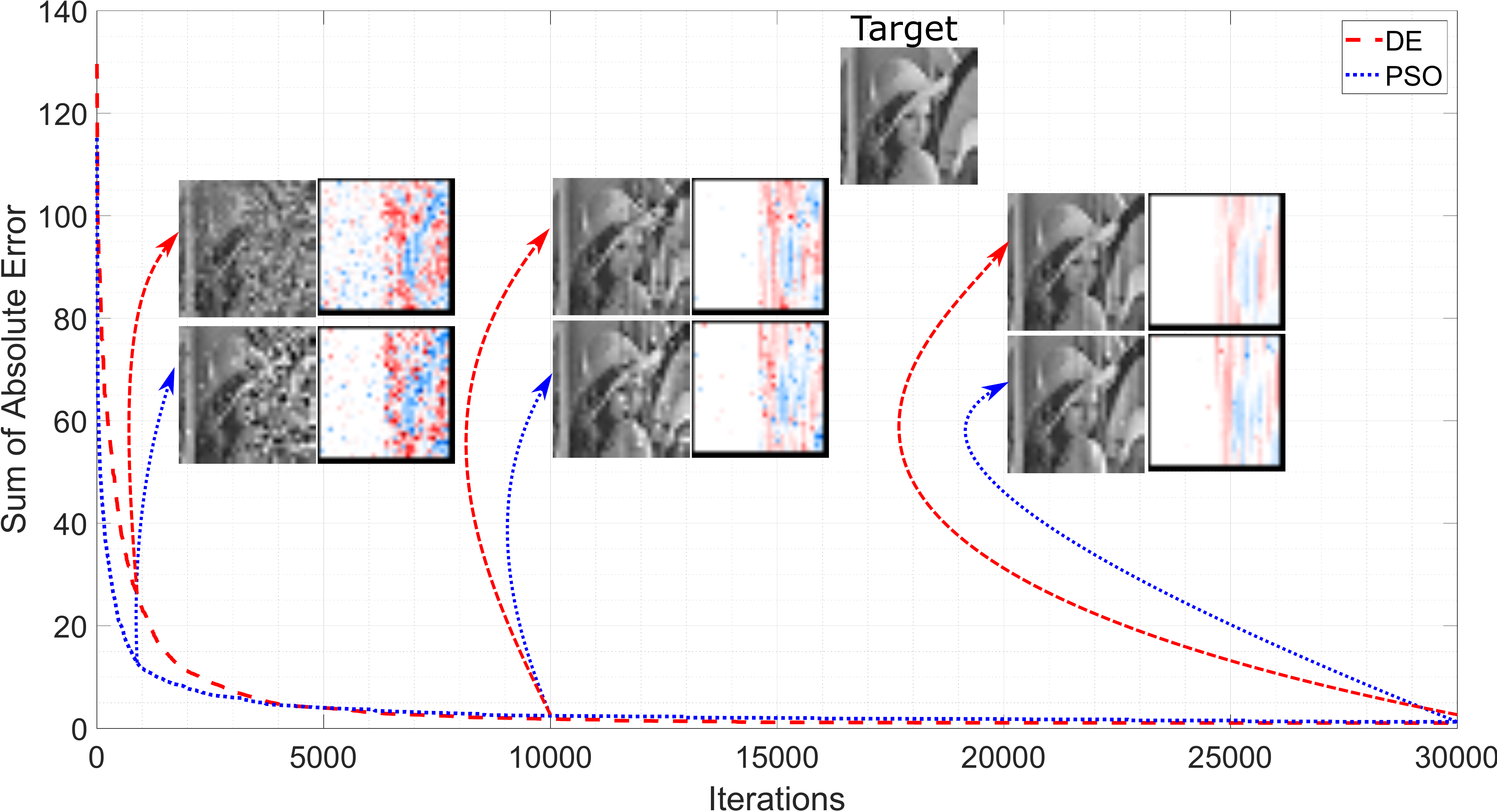}
				\caption{DE and PSO on continuous 30x30 Lena using 50\% SAE and 50\% SSIM. Images scaled by 500\%.}
				\label{fig:sep-de-pso-lena30-SAE-SSIM}
			\end{figure}
		
		\end{casestudy}
		
		\begin{casestudy}[Continuous, constrained, single-objective]
			
			The general definition of a constrained optimization problem is given as
			
			\begin{equation}
				\label{eq:constrained-optimization}
				\begin{array}{ll}
					\textit{Minimize:}   & f(\vec{x})                              \\
					\textit{Subject to:} & g_j(\vec{x}) \leq 0, j = 1, \ldots, n_g \\
					& h_k(\vec{x}) = 0, k = 1, \ldots, n_h    \\
					& \forall i: x_{i} \in dom(x_i), i = 1, \ldots, n
				\end{array}
			\end{equation}
			where $n_g$ and $n_h$ are the number of inequality and equality constraints, respectively, and $dom(x_i)$ is the domain of each variable, as defined in Equation (\ref{eq:unconstrained-optimization}). Note that, if there are no constraints present, i.e., $n_g = n_h = 0$, this definition is equivalent to the definition of an unconstrained optimization problem, as given in Equation (\ref{eq:unconstrained-optimization}).
			
			To produce a constrained continuous problem, the search domain, i.e., $dom(x_i)$, was widened to a range of $[-1,2]$, while the feasible domain remained at $[0,1]$. This introduces an inequality constraint that feasible candidate solutions must adhere to. Thus, decision variables that were outside of the feasible domain were considered to be constraint violations. Two variants of PSO were employed to handle the constraints. The PSO with no constraint handling (PSO-NCH) variant employed no constraint handling, i.e., was a vanilla PSO, and thus was purely focused on reducing the objective function. The PSO with scaled constraint handling, no penalty (PSO-SCHNP) variant employed Deb's constraint handling technique \cite{deb2000efficient}, whereby infeasible solutions were always inferior to feasible solutions and two infeasible solutions were compared according to the sum of the magnitudes of their violations. However, the fitness function did not account for the violations (i.e., there was no penalty term) and thus was simply the sum of absolute errors. Thus, the PSO-SCHNP variant first optimizes the constraints, then the fitness, which explains the increasing fitness value near the beginning of the optimization process in Figure \ref{fig:constraints-pso-de-sae-amount}.

			Figure \ref{fig:constraints-pso-de-sae-amount} presents the results of employing image-based visualization on a constrained optimization problem. In Figure \ref{fig:constraints-pso-de-sae-amount}, the individual decision variable violations are visualized via the pixel colour, while the colour of the border represents the absolute sum of violation amounts across the entire solution. This highlights another strength of image-based visualization, namely that additional features can also be visualized through the use of colour and, for example, adding borders. Interestingly, the raw fitness of PSO-NCH was superior to PSO-SCHNP. However, this is a result of its inherent focus on purely minimizing the fitness, without regard for the constraints, as can been seen by its inability to attain a feasible solution. In contrast, PSO-SCHNP was able to rapidly attain a feasible solution, after which the solution remained feasible, then slowly improved the fitness. Importantly, such observations could not be made without the use of this visualization framework, which inherently embeds important information about the search process in the visualization.
			
			\begin{figure}[!t]
				\centering
				\includegraphics[width=\linewidth]{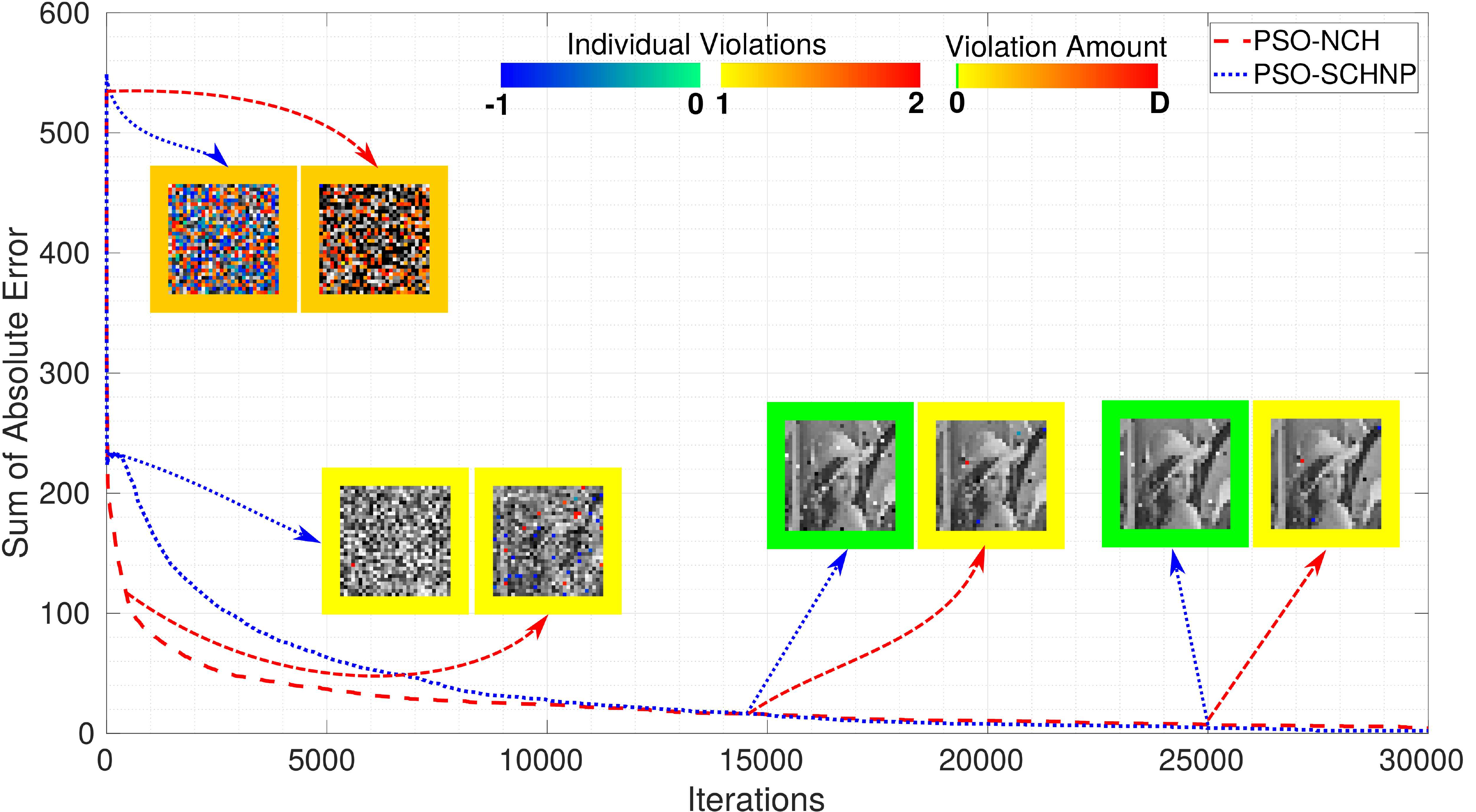}
				\caption{PSO and DE on 30x30 constrained continuous Lena using SAE. Borders visualize the absolute sum of violations. Images scaled by 400\%.}
				\label{fig:constraints-pso-de-sae-amount}
			\end{figure}
			
		\end{casestudy}
		
		\begin{casestudy}[Dynamic, continuous, unconstrained, single-objective]
			
			A dynamic environment is an environment where the optimal solution changes over time. Such dynamic environments can exhibit changes in the location of the optimum, the fitness of solutions, or both. This poses an additional challenge to an optimizer as it now must be responsible for not only finding the optimum, but also tracking the optimum throughout environmental changes. A dynamic optimization problem can be formally defined as
		
			\begin{equation}
				\begin{array}{ll}
					\textit{Minimize:}   & f(\vec{x}, t)                           \\
					\textit{Subject to:} & g_j(\vec{x}, t) \leq 0, j = 1, \ldots, n_g \\
					& h_k(\vec{x}, t) = 0, k = 1, \ldots, n_h    \\
					& \forall i : x_{i} \in dom(x_i), \enskip i = 1, \ldots, n
				\end{array}
			\end{equation}
			where the objective is to find the time-dependent optimal solution at time $t$, given by $$\vec{x}^*(t) = \argmin_\vec{x} f(\vec{x}, t)$$ subject to the time-dependent constraints $\vec{g}(\vec{x}, t)$ and $\vec{h}(\vec{x}, t)$. Note that, in a dynamic optimization problem, the fitness function and constraints may vary with time.
			
			To simulate a continuous dynamic environment in the context of this study, an animated GIF image was used as the target with $dom(x_i) = [0,1]$. The target GIF consisted of 21 frames, which were taken as successive environments for the optimizer. Specifically, the total number of iterations was divided equally into 21 environments, where each environment used the subsequent frame from the animation as the target image. Therefore, each environmental change introduced a new optimal solution and, as a byproduct, a different fitness value associated with each solution. In the context of this study, $n_g = n_h = 0$, thus producing an unconstrained dyanmic environment.
			
			To remove the influence of change detection mechanisms, the optimizers were aware of the frequency of environmental changes. A re-evaluating change response mechanism was employed whereby individuals re-evaluate their fitness (as well as the personal bests and global best for PSO) when an environmental change has occurred. The results presented in Figure \ref{fig:dynamic-pso-de-sae-reeval} depict the performance of PSO and DE on the dynamic problem. While PSO depicted superior performance on the initial environment, it is evident that the PSO algorithm struggled to effectively recover from the environmental changes, which is to be expected given that PSO is known to suffer from obsolete memory in dynamic environments \cite{blackwell2006multiswarms}. The visual quality of the subsequent images produced by the DE optimizer depicted better visual quality, thereby indicating superior overall performance by DE on this dynamic problem.
			
			\begin{figure}[!t]
				\centering
				\includegraphics[width=\linewidth]{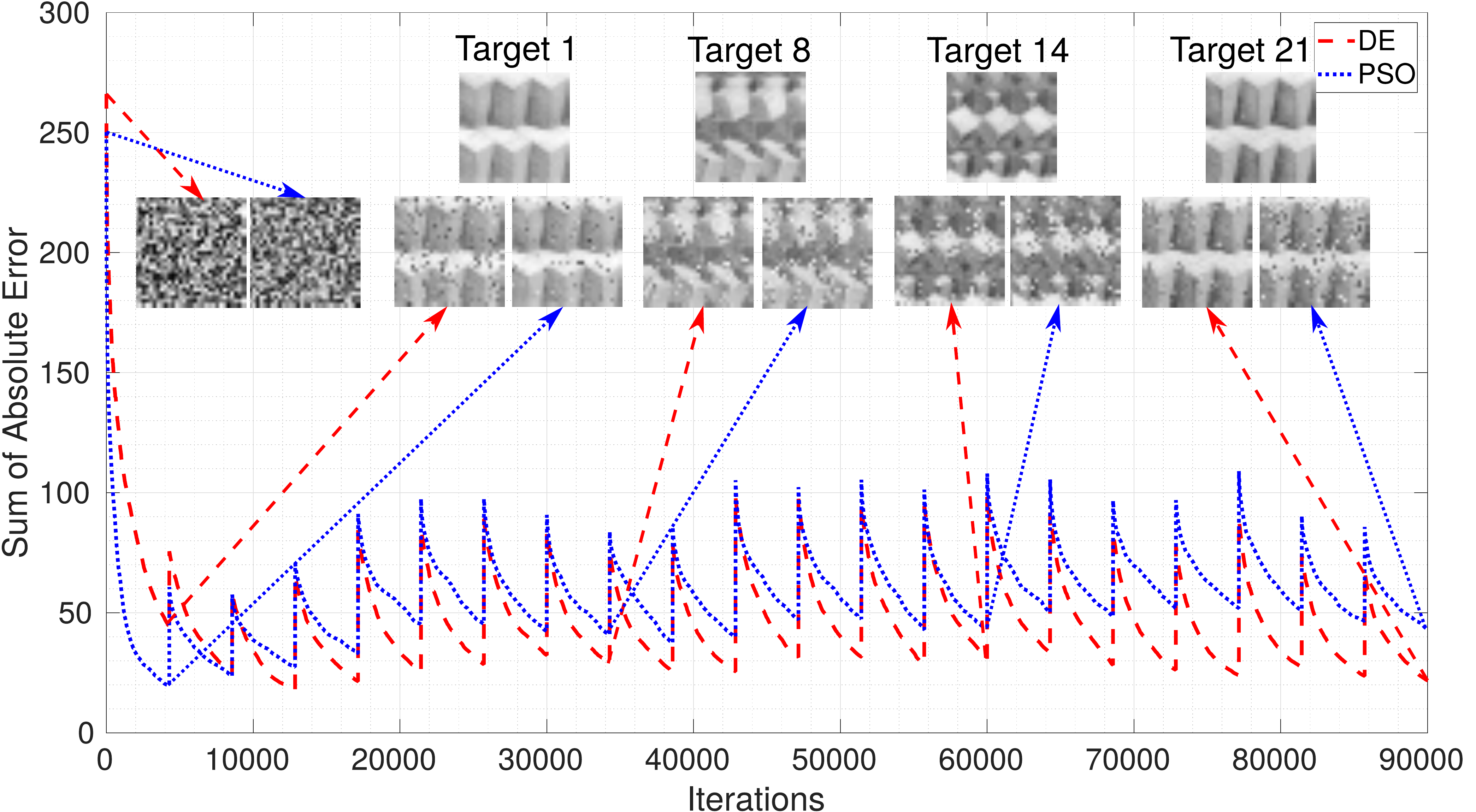}
				\caption{PSO and DE on 30x30 dynamic target image using SAE. Images scaled by 400\%.}
				\label{fig:dynamic-pso-de-sae-reeval}
			\end{figure}
		\end{casestudy}
		
		\begin{casestudy}[Multi-objective, unconstrained]
		
			A multi-objective optimization problem can be formally defined as
			
			\begin{equation}
				\label{eq:multi-objective-optimization}
				\begin{array}{ll}
					\textit{Minimize:}   & \vec{f}(\vec{x})                        \\
					\textit{Subject to:} & g_j(\vec{x}) \leq 0, j = 1, \ldots, n_g \\
					& h_k(\vec{x}) = 0, k = 1, \ldots, n_h    \\
					& \forall i : x_{i} \in dom(x_i), \enskip i = 1, \ldots, n
				\end{array}
			\end{equation}
			where $\vec{f}(\vec{x}) = (f_1(\vec{x}), ..., f_{n_o}(\vec{x}))$, referred to as an \textit{objective vector}, contains the values of the $n_o$ sub-objectives. In general, these sub-objectives are in conflict with each other such that they cannot be optimized simultaneously.  An optimizer must then make trade-offs between sub-objectives to produce a variety of non-dominated solutions (see Equation (\ref{eq:dominance})). Ideally, the optimizer will produce a set of non-dominated solutions that are well distributed along the Pareto front.
			
			In order to produce a multi-objective environment in the context of this study, two target images (i.e., sub-objectives) were considered simultaneously. To produce a bi-objective optimization problem, two target images were taken as an original target image and its inverse. The primary image was taken as a black and white image (corresponding to a binary optimization problem), and its inverse was created by inverting the pixel values (i.e., setting values of 0 to 1 and vice versa). As such, the second target image is in direct conflict with the original image. As with the dynamic case, $n_g = n_h = 0$, thus producing an unconstrained environment. GDE3 was utilized to optimize both images in a multi-objective context using PSNR as the evaluation metric. Formally, the problem can be defined by 
			\begin{equation}
				\label{eq:multi-objective-scheme}
				\begin{array}{ll}
				\textit{Minimize:}   & \vec{f}(\vec{x}) = (PSNR(\vec{x}, T), PSNR(\vec{x}, T'))\\
				\end{array}
			\end{equation}
			where $T$ is the target image and $T'$ is the inverse of $T$. 

			Figure \ref{fig:multi-gde3-house30-SAE} presents the results of the bi-objective optimization problem. It was observed that GDE3 was able to locate the extreme points on the Pareto front. Moving along the Pareto front, the quality of images were degraded because the optimizer is forced to make a trade-off between the two conflicting objectives. Thus, solutions towards the middle segment of the Pareto front don't necessarily correspond to meaningful images.
			
			\begin{figure}[!t]
				\centering
				\includegraphics[width=\linewidth]{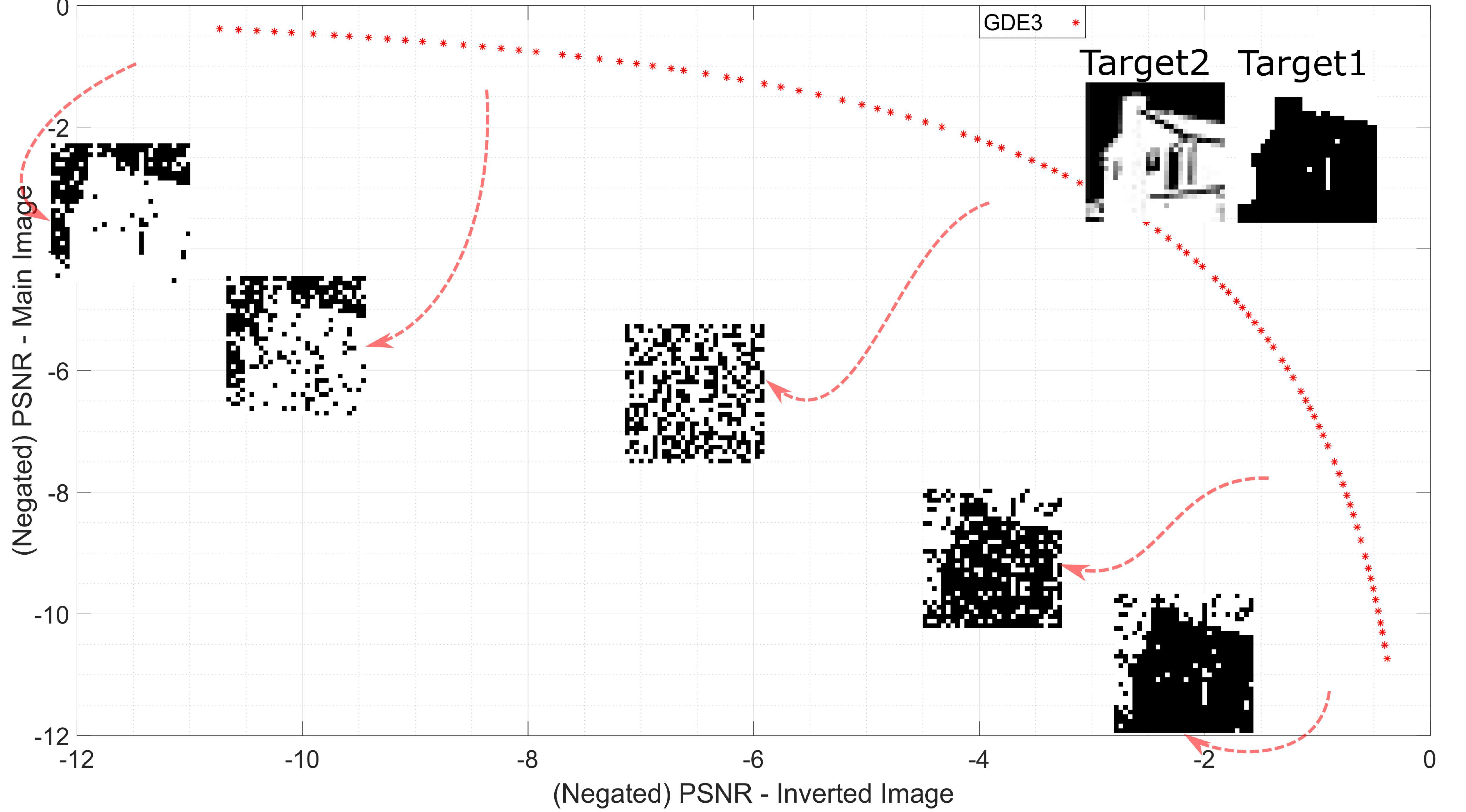}
				\caption{GDE3 on binary 30x30 House using PSNR on standard and inverted images. Images scaled by 500\%.}
				\label{fig:multi-gde3-house30-SAE}
			\end{figure}
		\end{casestudy}
		
		\begin{casestudy}[Visualization across multiple runs]
		
			Proper empirical evaluation of algorithms requires one to execute multiple independent runs of an algorithm. However, the preceding experiments only exemplify visualization of a single run. Thus, this scheme examines how image-based visualization can be used in the context of multiple runs. One possible strategy is to visualize the best images from each independent run. An alternative strategy is to visualize the average image produced at various iterations. However, visualization using the average image is somewhat problematic as the image presented doesn't correspond to any image produced by either optimizer. Rather, it is simply a visualization of the average best solution at various points.
			
			Figure \ref{fig:multiple-pso-de-sae} presents the individual images produced at various points along with the average images for both DE and PSO across five runs. A few interesting observations can be made from these images. Firstly, as with the individual runs, the superior early performance of PSO is readily apparent via the better images produced by all runs after 1000 iterations. Secondly, the optimizers appear to be stable, in that their performance at specific iterations was comparable for all runs. Finally, while the average image does not correspond to any particular solution, it does depict better visual quality than any of the images produced by the individual runs. This visually highlights the smoothing effect of averaging -- the average image provides a smoothing effect across all individual runs, thereby producing an image that more closely resembles the target.
			
			\begin{figure}[!t]
				\centering
				\includegraphics[width=\linewidth]{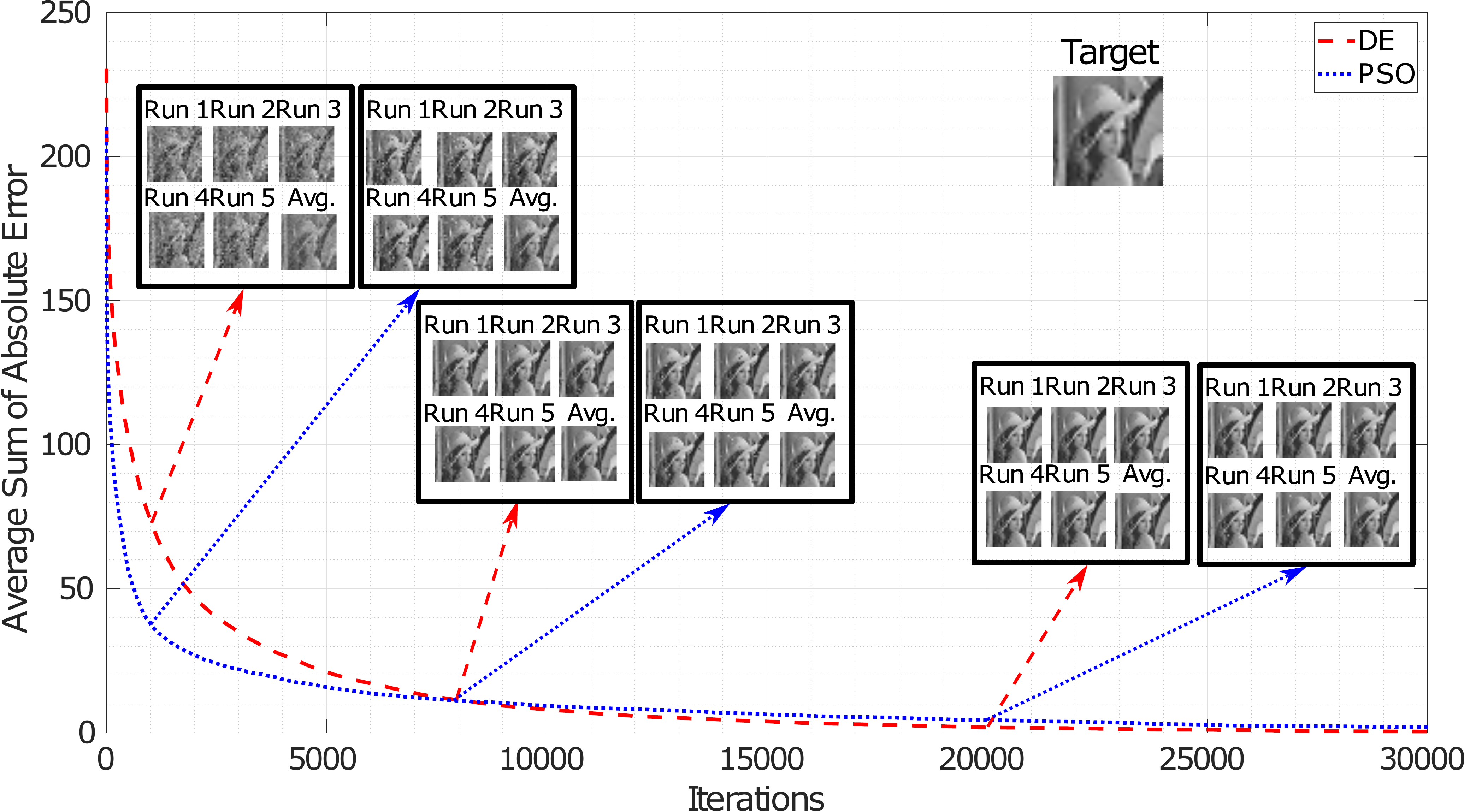}
				\caption{PSO and DE on 30x30 continuous Lena image using SAE. Images scaled by 200\%.}
				\label{fig:multiple-pso-de-sae}
			\end{figure}
			
		\end{casestudy}
		
    \section{Image-Based Visualization for Arbitrary Single-Objective Benchmark Functions with Known Optima}
    \label{sec:known-optima}
    
        This section examines the image-based visualization framework applied to arbitrary single-objective (minimization) benchmark problems with known optima using the mapping function given in Equation \eqref{eq:linearMapping}. All problems were optimized in 900 dimensions, corresponding to a 30x30 target image and were optimized with both the DE and PSO optimizers. The experimental procedures, specifically the algorithmic control parameters and image generation process, were the same as in Section \ref{sec:case-studies}, with the exception of Equation \eqref{eq:linearMapping} used as the mapping function. Table \ref{tbl:problems} gives a brief overview of the benchmark functions examined in this study, presented in alphabetical order. These functions were chosen to represent a wide variety of characteristics, such as modality, separability, and location of the optimum. 
    
        \begin{table}[htbp]
    		\caption{Benchmark Functions}
    		\begin{center}
    			\begin{tabular}{|l|l|c|}
    				\hline
    				\textbf{Problem} &    \textbf{Domain} & \textbf{Optimum}   \\ \hline
    				Qing   &    $[-500, 500]^D$   &  $\vec{0}$ \\
    				Rastrigin    &     $[-5.12, 5.12]^D$   & $\vec{0}$ \\
    				Rosenbrock      &    $[-30, 30]^D$ & $\vec{1}$   \\
    				Salomon        &    $[-100, 100]^D$  &  $\vec{0}$\\
    				Spherical        &    $[-5.12, 5.12]^D$ &  $\vec{0}$ \\
    				Styblinski-Tang   &   $[-5, 5]^D$ &  $\approx \vec{-2.90354}$\\
    				Wavy      &     $[-\pi, \pi]^D$    & $\vec{0}$ \\\hline
    		
    			\end{tabular}
    			\label{tbl:problems}
    		\end{center}
    	\end{table}
    	
\begin{figure}[!t]
    		\centering
    		\includegraphics[width=\linewidth]{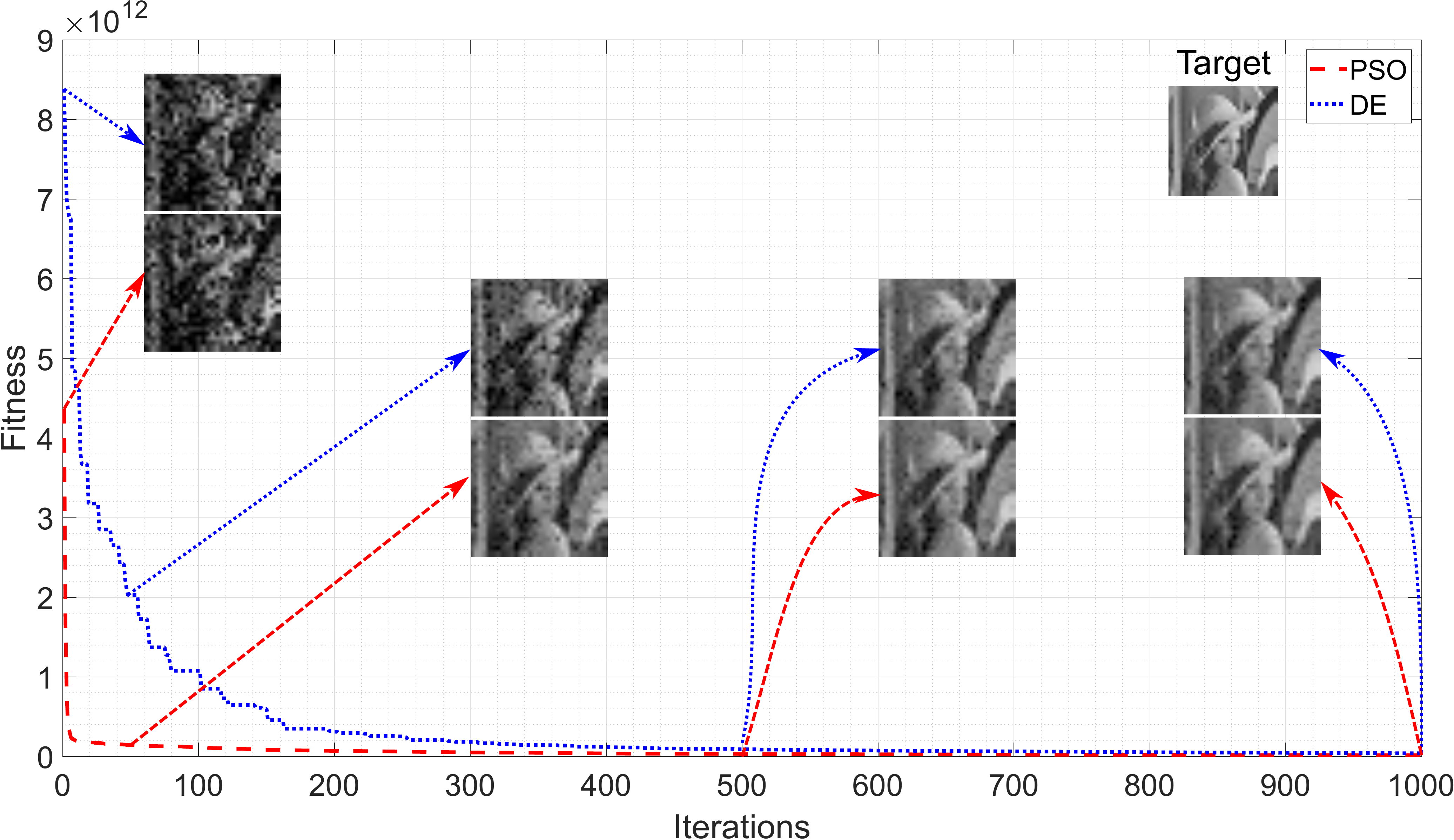}
    		\caption{PSO and DE on the 900D Qing function using 30x30 continuous Lena image. Images scaled by 500\%.}
    		\label{fig:known-pso-de-qing}
    	\end{figure}

    	Figure \ref{fig:known-pso-de-qing} presents a comparison of PSO and DE on the Qing function. It is observed that relative to the initial solutions, the fitness of PSO more rapidly improves. This improvement in fitness is directly correlated with an improvement in the image quality. However, after 1000 iterations, both optimizers arrive at solutions with similar fitness, thereby producing images that are of similar visual quality. Nonetheless, the pixel-level differences of the resulting images indicate that despite the optimizers attaining a similar fitness value, the solutions correspond to different areas of the search space. Similarly, the relatively high-quality images indicate that the fitness associated with both optimizers are near-optimal.
    
\begin{figure}[!t]
    		\centering
    		\includegraphics[width=\linewidth]{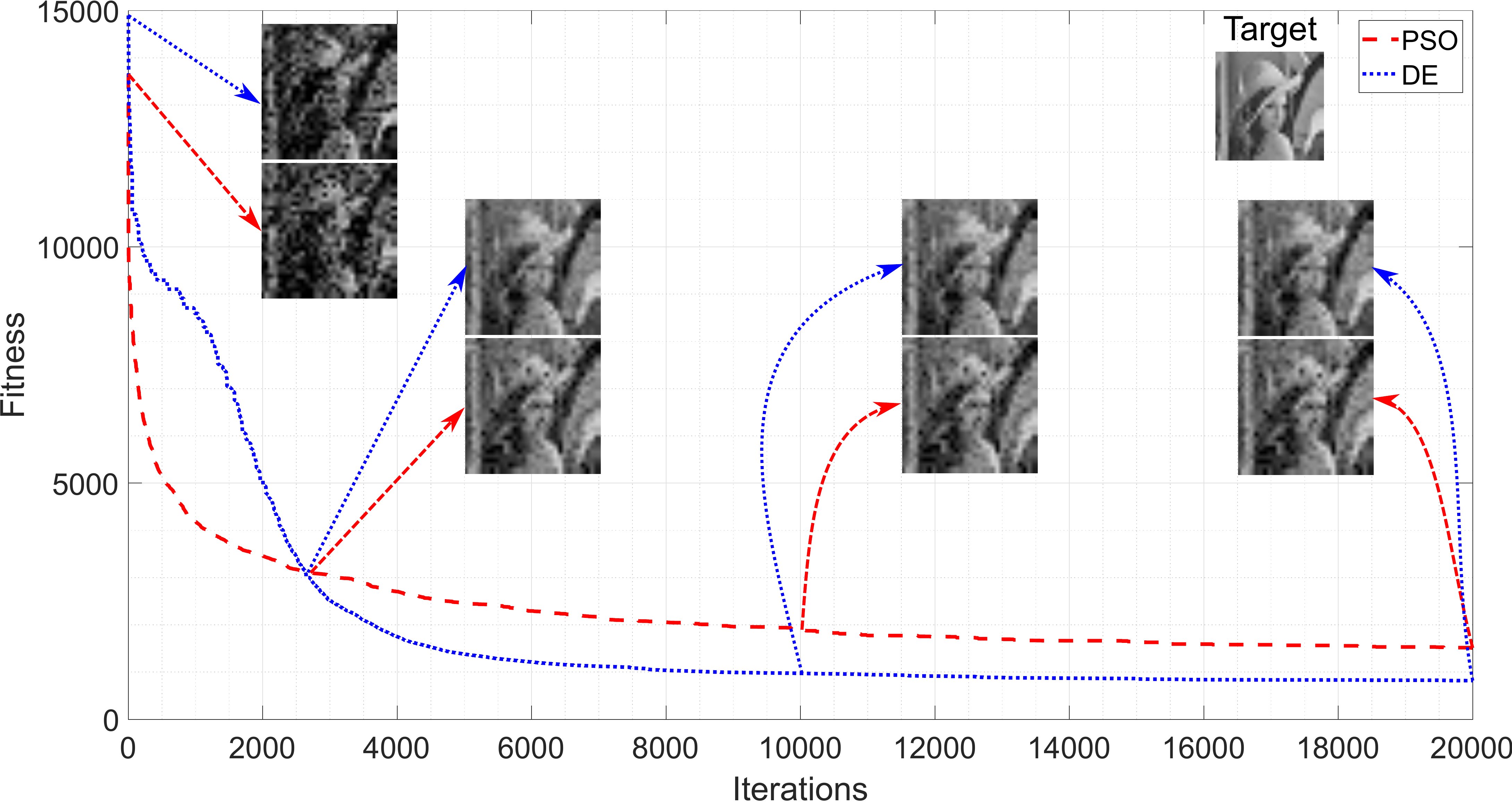}
    		\caption{PSO and DE on the 900D Rastrigin function using 30x30 continuous Lena image. Images scaled by 500\%.}
    		\label{fig:known-pso-de-rastrigin}
    	\end{figure}

    	Figure \ref{fig:known-pso-de-rastrigin} presents a comparison of PSO and DE on the Rastrigin function. As with the Qing function, it was observed that PSO lead to a more rapid improvement in fitness. However, the performance of the optimizers was nearly identical at 2625 iterations, at which point the differences in the images clearly indicates that different solutions were attained. From iteration 2625 onwards, the performance of DE, and hence the resulting image quality, was better than PSO. Despite the improved performance, it is evident from the quality of image produced that even after 20000 iterations, the solution produced by DE is still not sufficiently close to optimal.
    	
\begin{figure}[!t]
    		\centering
    		\includegraphics[width=\linewidth]{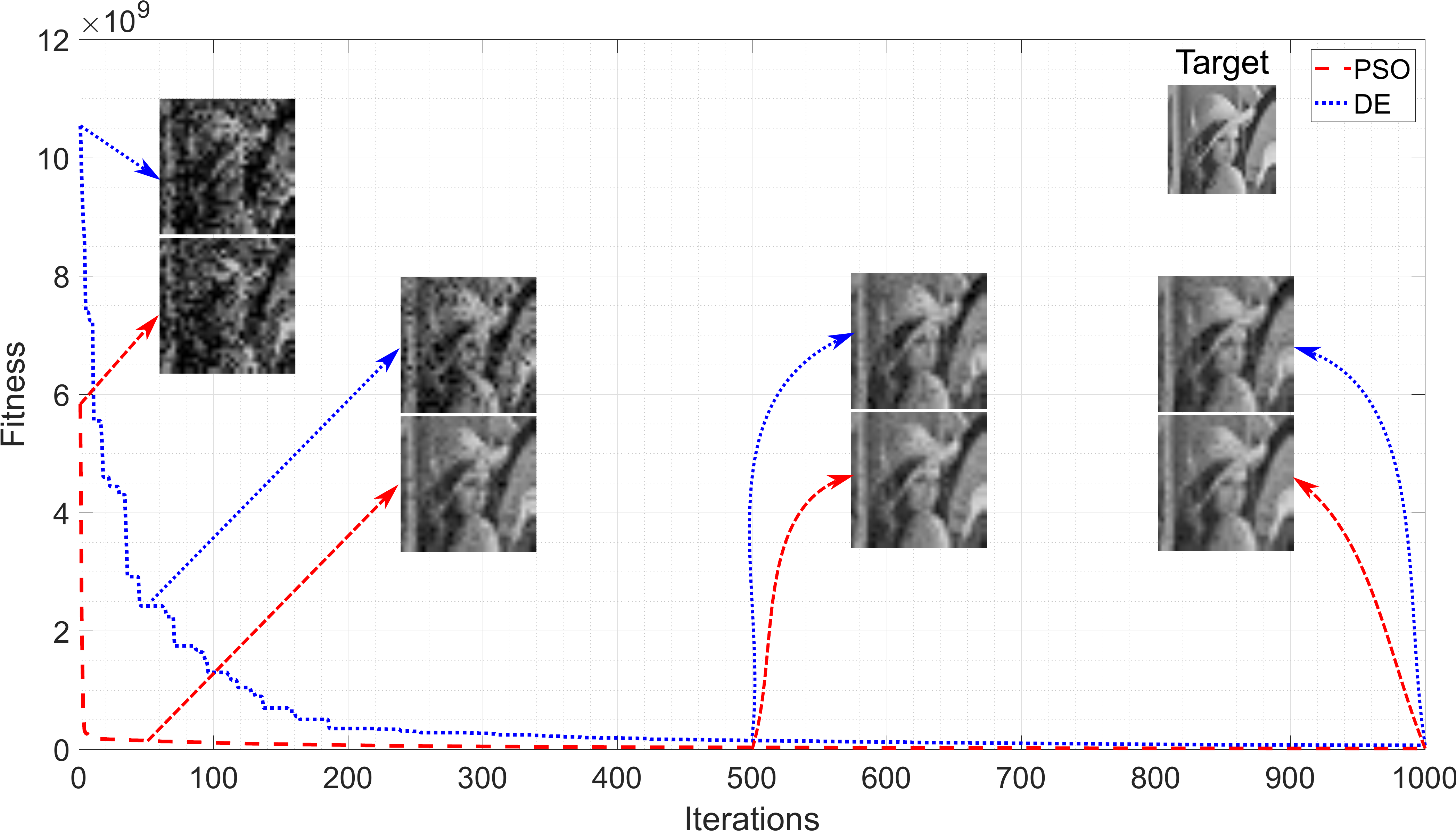}
    		\caption{PSO and DE on the 900D Rosenbrock function using 30x30 continuous Lena image. Images scaled by 500\%.}
    		\label{fig:known-pso-de-rosenbrock}
    	\end{figure}

    	Figure \ref{fig:known-pso-de-rosenbrock} presents a comparison of PSO and DE on the Rosenbrock function, where it was observed that PSO lead to a much more rapid improvement in fitness and resulting image quality. Similar to the Qing function, both optimizers arrived at similar objective fitnesses after 1000 iterations, but at noticeably different locations in the search space as evidenced by the differences in the resulting images.
    	
\begin{figure}[!t]
    		\centering
    		\includegraphics[width=\linewidth]{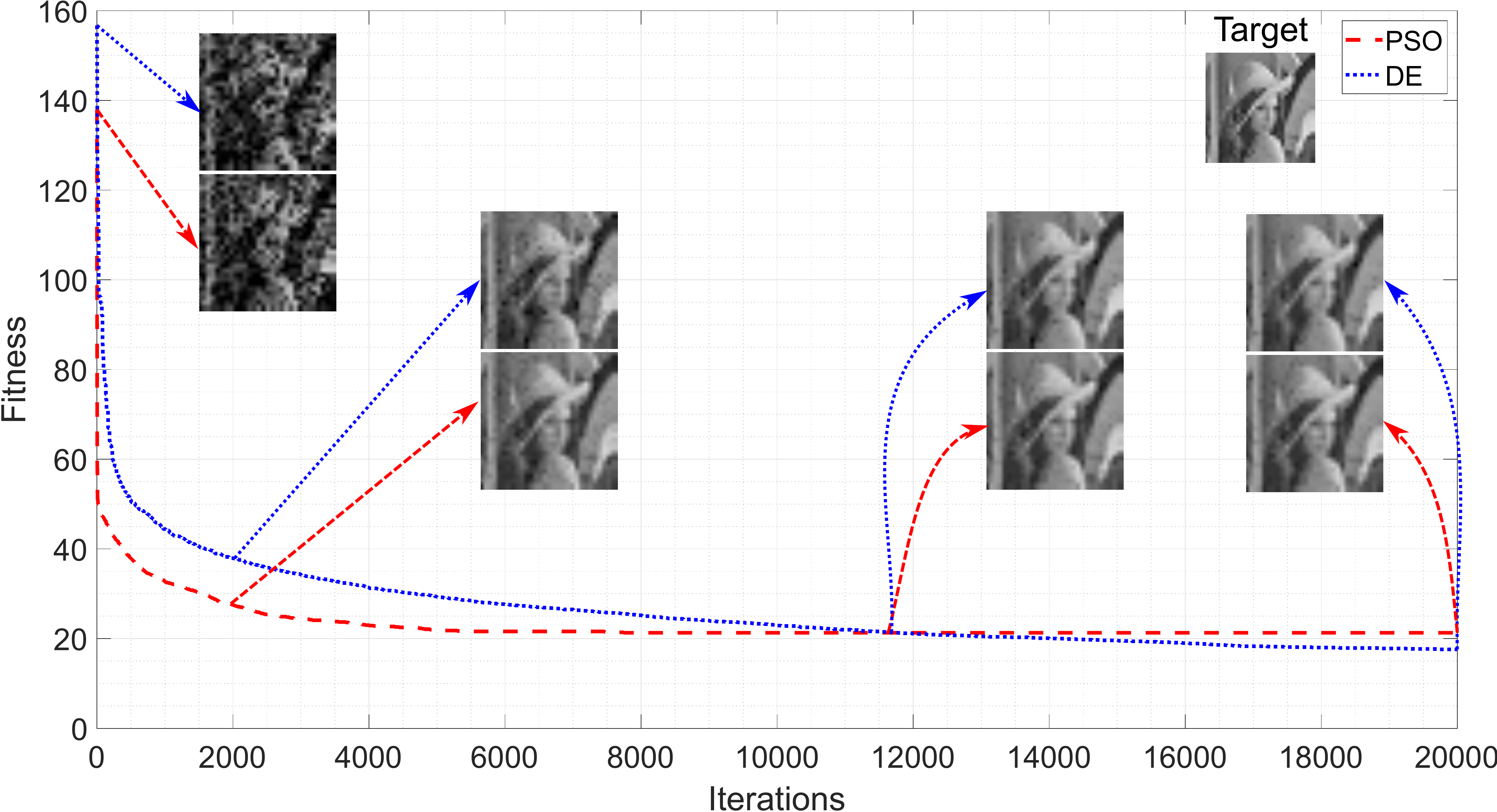}
    		\caption{PSO and DE on the 900D Salomon function using 30x30 continuous Lena image. Images scaled by 500\%.}
    		\label{fig:known-pso-de-salomon}
    	\end{figure}

    	Figure \ref{fig:known-pso-de-salomon} presents a comparison of PSO and DE on the Salomon function. Initially, PSO demonstrates superior performance and leads to a noticeably clearer image after 2000 iterations. However, after 2000 iterations, DE also produced a relatively high-quality image, thereby indicating that both optimizers found well-fit solutions at this point in the search. After 11747 iterations, the fitness of the optimizers is approximately equal, after which point DE outperforms PSO. Nonetheless, the high quality of the images produced by both optimizers demonstrates that both optimizers performed relatively well on the Salomon function. 
    	
\begin{figure}[!t]
    		\centering
    		\includegraphics[width=\linewidth]{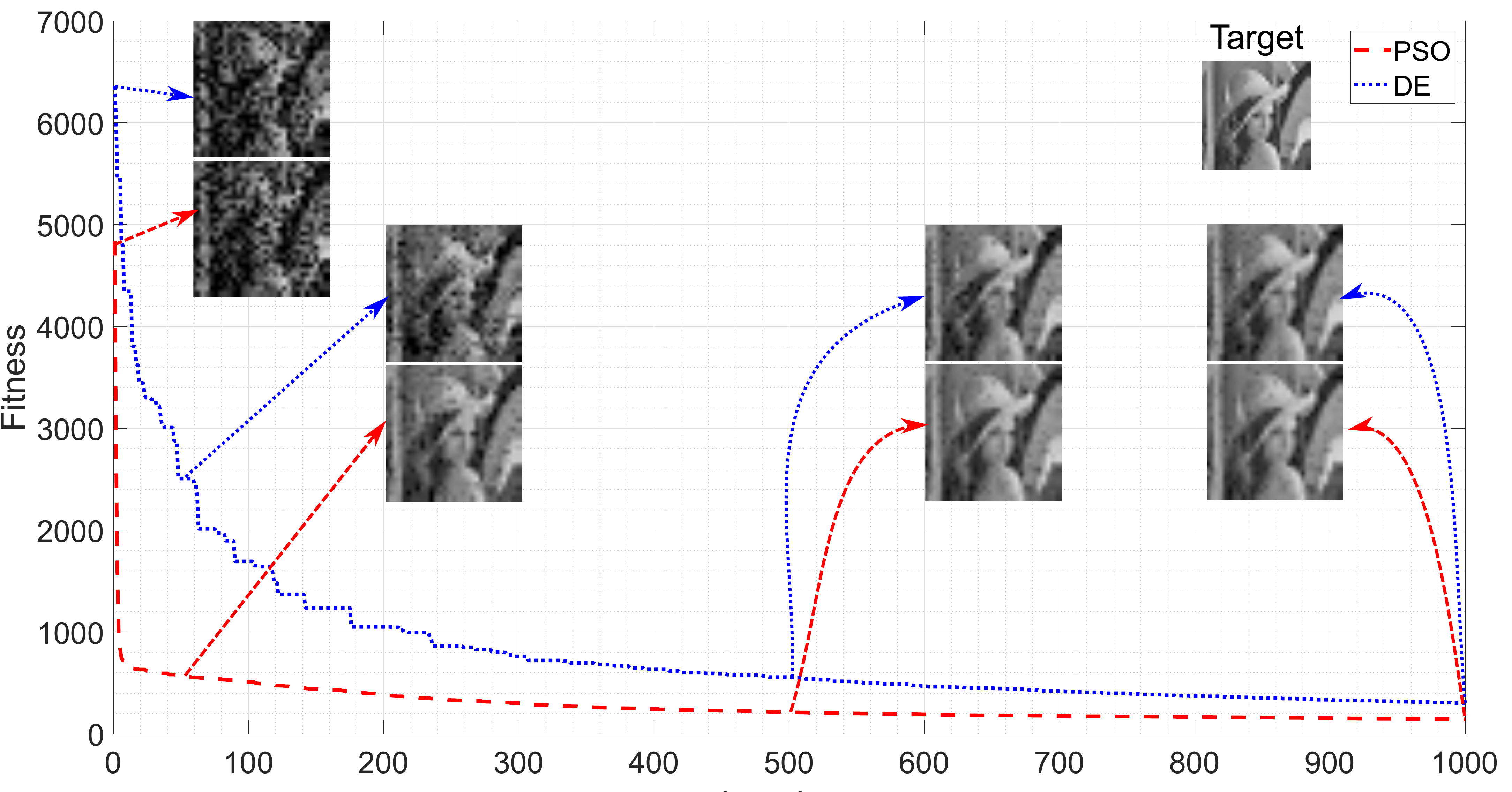}
    		\caption{PSO and DE on the 900D Spherical function using 30x30 continuous Lena image. Images scaled by 500\%.}
    		\label{fig:known-pso-de-spherical}
    	\end{figure}

    	Figure \ref{fig:known-pso-de-spherical} presents a comparison of PSO and DE on the Spherical function. Note that, the Spherical function is considered to be relatively simple, and thus after 1000 iterations, both PSO and DE found high-quality solutions, thereby resulting in high quality images. Despite the similar fitness and image quality, the image produced by PSO is noticeably better, correspoding to its superior performance.
    	
\begin{figure}[!t]
    		\centering
    		\includegraphics[width=\linewidth]{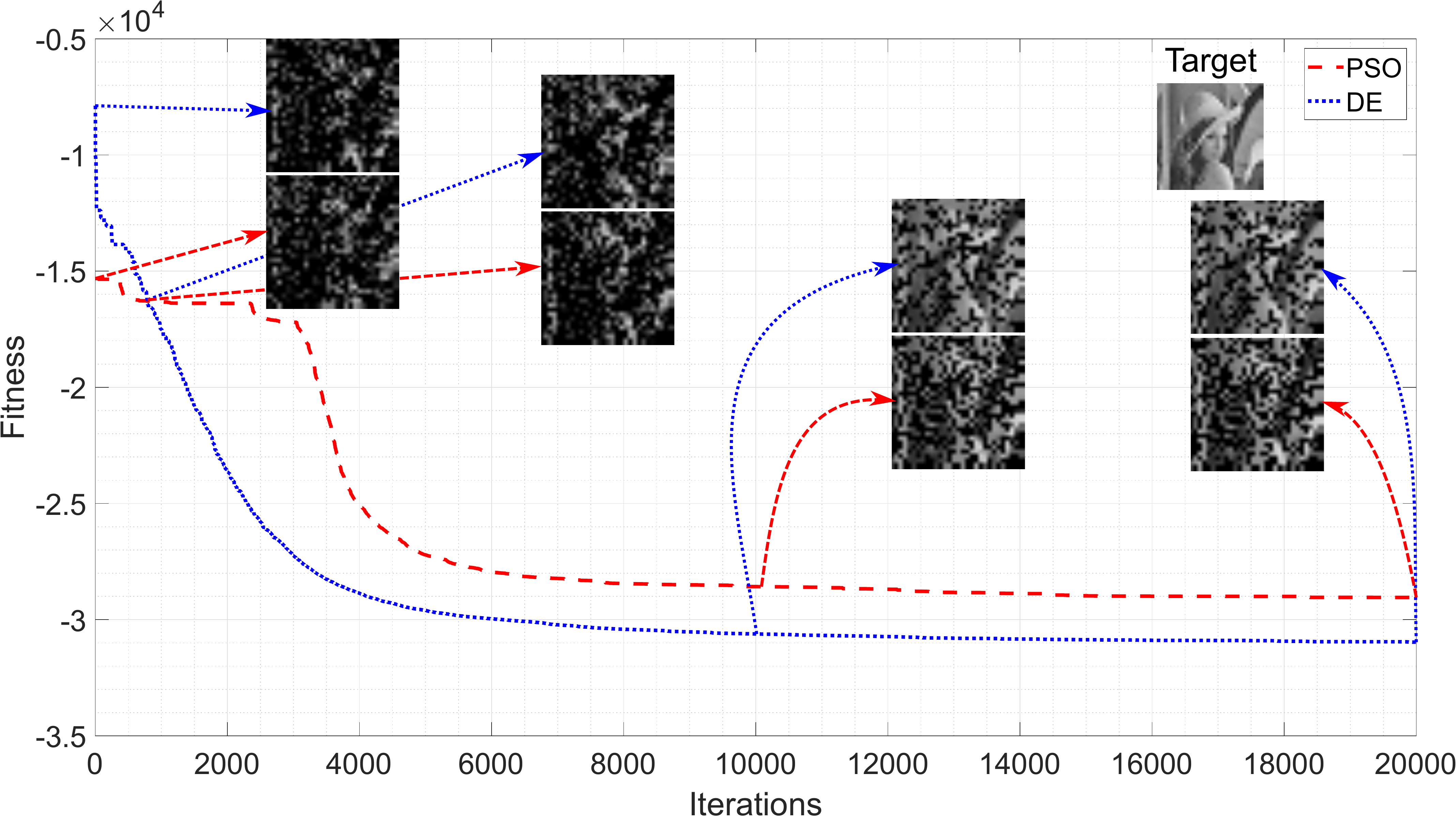}
    		\caption{PSO and DE on the 900D Styblinksi-Tang function using 30x30 continuous Lena image. Images scaled by 500\%.}
    		\label{fig:known-pso-de-styblinskitang}
    	\end{figure}

    	Figure \ref{fig:known-pso-de-styblinskitang} presents a comparison of PSO and DE on the Styblinski-Tang function. The poor quality of images, even after 20000 iterations, indicates that neither optimizer was able to achieve a reasonably-fit solution. Moreover, the relatively low quality images facilitates concluding that the Styblinski-Tang function was the most challenging problem for both optimizers among those considered. Furthermore, the black pixels in the resulting images indicates that both optimizers were deceived into searching near the lower end of the search domain in many dimensions. Based on this observation, it was hypothesized that a basin of attraction exists in the vicinity of the lower bound of the search space. Visual inspection of the 2D problem landscape confirmed this hypothesis and exemplified a striking benefit to the usage of image-based visualization -- namely, that an intrinsic property of the fitness landscape associated with benchmark problem was made clearly visible through the use of the image-based visualization framework.
    	
\begin{figure}[!t]
    		\centering
    		\includegraphics[width=\linewidth]{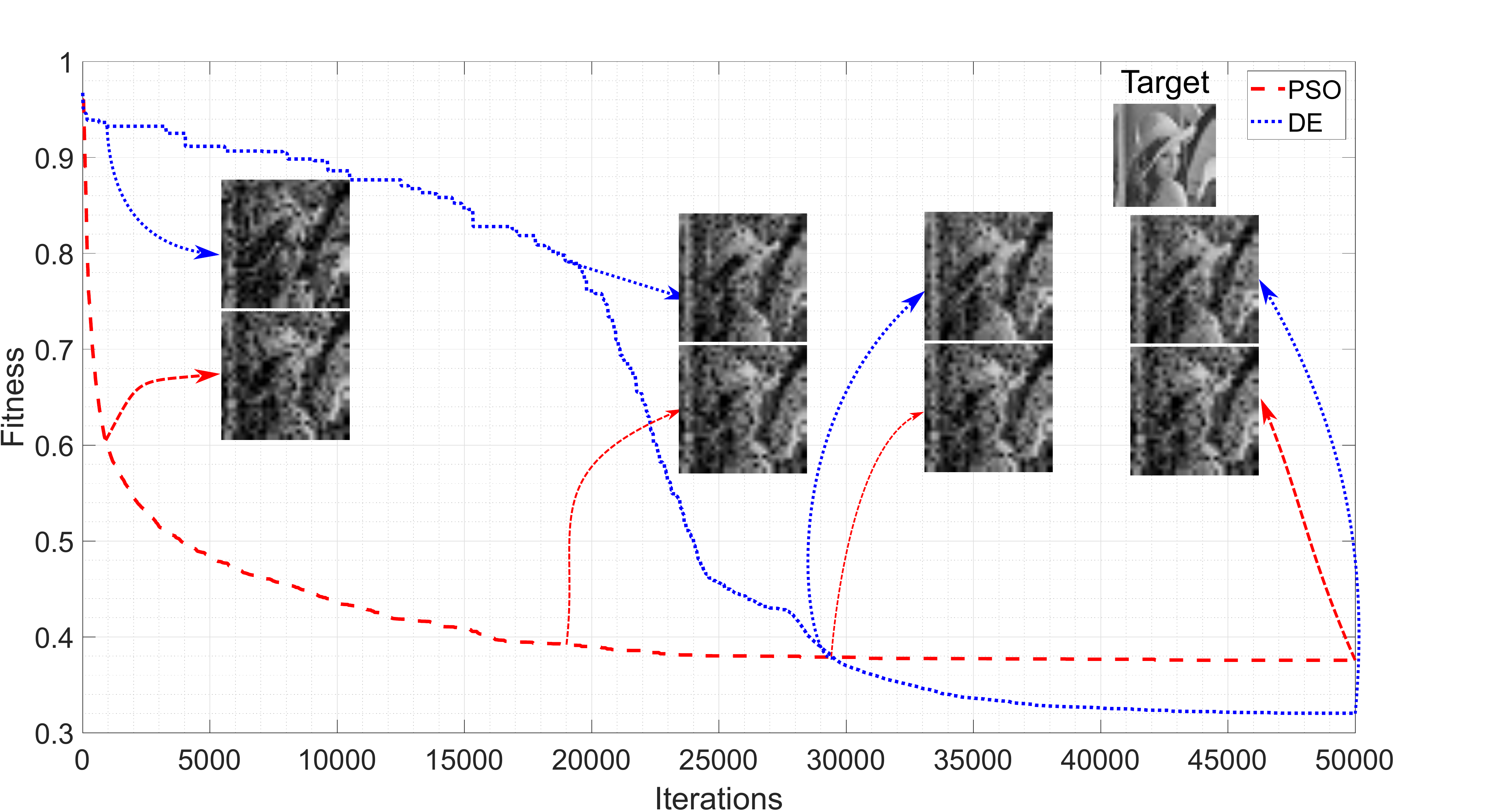}
    		\caption{PSO and DE on the 900D Wavy function using 30x30 continuous Lena image. Images scaled by 500\%.}
    		\label{fig:known-pso-de-wavy}
    	\end{figure}

    	Finally, Figure \ref{fig:known-pso-de-wavy} presents a comparison of PSO and DE on the Wavy function. Again, the low quality of solutions indicated that neither optimizer was able to effectively optimize this problem, even after 50000 iterations. As with most problems, PSO demonstrated superior initial performance. After 29481 iterations, the fitness of both optimizers was approximately equal, after which DE demosntrated superior performance. Regardless of its superior performance, DE did not arrive at a high-quality solution overall and thus produced an image that did not reasonably replicate the target.
    	
    	In summary, this section exemplified the usage of the image-based visualization framework on eight arbitrary benchmark problems with known optima. The results depict how the quality of the resulting images aligns with the observations made using traditional performance plots but additionally facilitates critical insights into both the optimizers' performance and the intrinsic characteristics of the benchmark problems themselves.
		
\section{Conclusions and Future Work}
	\label{sec:conclusions}

	The primary aim of this study was to propose a novel image-based visualization framework for large-scale global optimization. The proposed framework is the first instance of a visualization technique that explicitly visualizes both decision-space and objective-space without dimensionality reduction. The proposed framework offers a number of advantages over existing visualization techniques, such as dimensionality preservation and scalability, alignment with human perception, and the flexibility to visualize a wide variety of different problem types. However, it should be noted that the scalability is with regards to the number of dimensions, not necessarily the number of entities. In contrast, many of the existing techniques offer strong scalability in terms of the number of entities, while sacrificing the ability to maintain visualization of complex relationships between variables.
	
	The flexibility and robustness of the proposed framework were first demonstrated on a suite of different optimization types and mapping schemes using image replication as an example optimization problem. The experiments encompassed mapping schemes for continuous, discrete, binary, combinatorial, constrained, dynamic, and multi-objective optimization, thereby demonstrating the suitability of the approach for a wide variety of optimization environments. A number of different image quality metrics were also examined to ascertain their alignment with human perception and to induce different landscape characteristics. The critical insights facilitate by the image-based visualization framework were further exemplified by its application to eight arbitrary benchmark functions. Through these examples, the benefits of the proposed framework were elucidated, and the ability for the user to derive critical information about the underlying optimization process was highlighted.
	
	It should be noted that the proposed framework is easily extensible and can, for example, address mixed-type optimization by introducing a mapping function that explicitly accounts for differing variable types. Moreover, many of the examined mapping schemes can be composed, such that more complex types of optimization problems (such as dynamic, constrained, multi-objective optimization), can also be readily visualized. 
	
	This study constitutes only the first phase of a much larger study onm image-based visualization. Specifically, subsequent studies that examine more complex mapping functions are immediate future work. For example, the next phase involves further work on devising mapping functions that can visualize solutions to arbitrary optimization problem without a known optimal solution. Future work will also examine how dimensionality in terms of the number of entities can be addressed more effectively. Furthermore, the use of fitness-based mapping, where the overall quality of the resulting image is degraded based on a function of the objective fitness, will also be examined.
	
 \section*{Conflict of interest}

 The authors declare that they have no conflict of interest.	
	
\bibliographystyle{spbasic}
\bibliography{visualization}

\end{document}